\documentclass[letterpaper,12pt]{book}
\usepackage[letterpaper]{geometry}

\usepackage[utf8]{inputenc}
\usepackage[T1]{fontenc} 
\usepackage[cjk]{kotex}

\usepackage{amsmath}
\usepackage{graphicx}
\usepackage{url}
\usepackage{latexsym}

\usepackage{booktabs}
\usepackage[table]{xcolor}

\usepackage{hyperref}
\definecolor{darkblue}{rgb}{0, 0, 0.5}
\hypersetup{colorlinks=true,citecolor=darkblue, linkcolor=darkblue, urlcolor=darkblue}

\usepackage{tabularx}
\usepackage{datetime}
\usepackage{arydshln}

\usepackage{amsmath} 
\usepackage{amsfonts}
\usepackage{amsthm}
\usepackage{amssymb}
\usepackage{mathtools}
\usepackage{multicol}

\usepackage{setspace}
\usepackage{lineno}

\usepackage[authoryear,round]{natbib}
\usepackage{algorithmic,algorithm}
\renewcommand{\algorithmiccomment}[1]{\bgroup\hfill$\triangleright$~#1\egroup}

\usepackage[linguistics]{forest} 
\usepackage{synttree}

\usepackage{CJKutf8}

\usepackage{caption}
\usepackage{subcaption}
\usepackage{tikz-dependency}

\usepackage{titlesec}
\usepackage{gb4e}

\newtheorem{case}{Case}

\newcommand{\zh}[1]{\begin{CJK}{UTF8}{gbsn}#1\end{CJK}}
\newcommand{\np}[1]{\begin{CJK}{UTF8}{min}#1\end{CJK}}

\begin{document}


\begin{titlepage}
    \centering
    \vspace*{2cm}

    {\Huge\bfseries Foundations and Evaluations in NLP \par}
    {\Large New Methods for Annotating Linguistic Resources and Evaluating System Performance \par}
    
    \vspace{0.5cm}

    {\Huge\bfseries Fondements et évaluations en TAL \par}
    {\Large Nouvelles méthodes pour l’annotation des ressources linguistiques et l’évaluation des performances des systèmes \par}


    \vfill

    \includegraphics[width=.8\textwidth]{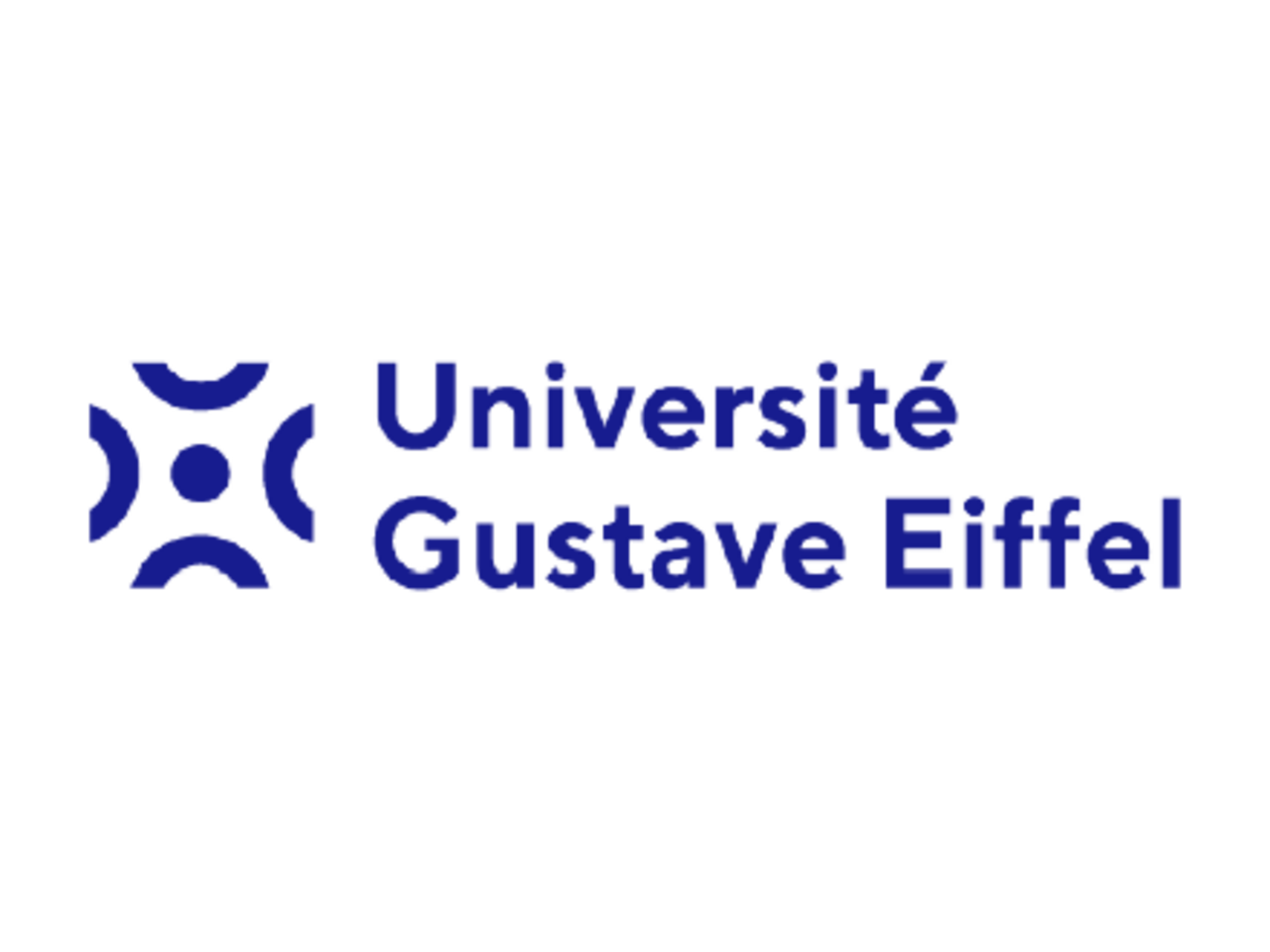} 

    \vfill
{\bf Jungyeul Park}\\
\vspace{.5cm}
Mémoire d'habilitation à diriger des recherches, 2025-2026

\end{titlepage}

\tableofcontents
\listoffigures
\listoftables

\doublespacing

\singlespacing
\chapter*{Acknowledgments}
\doublespacing

This \textit{Mémoire d’habilitation à diriger des recherches} is the result of a long intellectual journey that has unfolded over more than two decades, across institutions, countries, and research traditions. It is therefore a pleasure to acknowledge the many individuals who have accompanied, supported, and shaped this work.

I would first like to express my sincere gratitude to my supervisor, Professor Eric Laporte, for his guidance, trust, and thoughtful support throughout the preparation of this memoir. I still remember our first encounter in 1998, when I arrived in Paris as a student trained in computer science and eager to learn how linguistic questions could be pursued with the same rigor. From that moment onward, he has accompanied my work with a rare combination of intellectual generosity and demanding precision. He later served on my PhD jury in 2006, and I have valued ever since the continuity of his support across these major milestones. His steady encouragement and his commitment to scholarly rigor have been invaluable at every stage of this work, but so has his human presence: the quiet confidence with which he sustained long term projects, the patience with which he read and discussed them, and the care with which he helped me keep ambition and clarity in balance.

I am deeply grateful to the rapporteurs of this HDR: Professor Djamé Seddah, Professor Maxime Amblard, and Professor Jee-Sun Nam. I thank them warmly for the time and care they devoted to reading this manuscript, and for their insightful evaluations and constructive perspectives. I have had the privilege of knowing Professors Djamé Seddah and Maxime Amblard over many years as both colleagues and friends, and I am especially grateful for the constancy of their support and for the intellectual generosity that has marked our exchanges. I have also known Professor Jee-Sun Nam’s work and name since my arrival in France in 1998, through her remarkable achievements and long-standing contributions to the study of the Korean language. I am sincerely honored that she accepted the role of rapporteuse for this HDR, and I thank her for the seriousness, care, and rigor with which she engaged with this manuscript.

My sincere thanks also go to the members of the jury: Professor Chulwoo Park, Professor Alexis Nasr, and Professor Owen Rambow. I am especially grateful to Professor Chulwoo Park, who for more than twenty-five years has been not only a mentor in my academic work, but also a constant source of guidance, encouragement, and personal support throughout my life, extending far beyond any formal institutional role. His advice has shaped not only particular research decisions, but also the way I have learned to navigate an academic life with steadiness and purpose, and I remain deeply thankful for his enduring generosity.

I am particularly moved to acknowledge Professors Alexis Nasr and Owen Rambow, who served on my PhD jury in 2006 and now, nearly twenty years later, again as members of my HDR jury. Their continued presence represents, for me, a rare and meaningful continuity across an academic life. I am grateful for the seriousness and care with which they have followed my work over time, and I feel genuinely honored by their long-standing support and the confidence they have placed in me at these two defining moments.

This memoir builds upon collaborative research carried out with many colleagues and students over the years. I am grateful to all my co-authors and collaborators mentioned throughout this manuscript for their intellectual generosity, for the many discussions that sharpened my thinking, and for the joint efforts that shaped the ideas presented here. I would like to express my particular appreciation to KyungTae Lim, Mija Kim, Mengyang Qiu, Jayoung Song, Francis Tyers, Eunkyul Leah Jo, Yige Chen, Zihao Huang, Junrui Wang, and Yang Gu, whose contributions have been especially formative in the development of this work. Working with them has been both a privilege and a constant source of inspiration.

I would also like to express my sincere gratitude to several senior colleagues who have supported my academic path over the years with their advice, encouragement, and generosity: Professor Natasha Warner and Professor Mike Hammond (University of Arizona), Professor Jeff Good (The State University of New York at Buffalo), Professor Emily Bender (University of Washington), Professor Strang Burton (The University of British Columbia), and Professor Anne Abeillé (Université Paris Cité). Their support at important moments, and their continued confidence in my work, have meant a great deal to me.

I would also like to express my appreciation to the institutions and individuals who supported the dissemination of this memoir beyond the dissertation itself, in the form of two extended monographs. I thank Ji-sun Yoon at the Academy of Korean Studies Press for her support and patience in the preparation of Part~I of this work, which will appear as a standalone volume in March~2026 (\textit{Rethinking Linguistic Units in the Age of AI: A Morpheme-Based Approach to Korean Language Processing}). I am equally grateful to Susanne Filler at Springer Nature for her professionalism and support in bringing Part~II to publication as a volume in the \textit{Synthesis Lectures on Computer Science} series (\textit{Evaluation by Alignment – A Framework for Robust End-to-End NLP Assessment}). I would also like to thank Latifa Zeroual-Belbou at Paris-Est Sup for her careful administrative support and guidance throughout the HDR process.

Finally, my deepest gratitude goes to my family for their unconditional love and support—wherever I may be in France, the United States, Canada, or Korea, as has often been the case over the past twenty-five years. They have been the quiet strength behind every page of this mémoire.

\singlespacing
\chapter*{Abstract}
\doublespacing

This memoir explores two fundamental aspects of Natural Language Processing (NLP): the creation of linguistic resources and the evaluation of NLP system performance. Over the past decade, my work has focused on developing a morpheme-based annotation scheme for the Korean language that captures linguistic properties from morphology to semantics. This approach has achieved state-of-the-art results in various NLP tasks, including part-of-speech tagging, dependency parsing, and named entity recognition. Additionally, this work provides a comprehensive analysis of segmentation granularity and its critical impact on NLP system performance.
In parallel with linguistic resource development, I have proposed a novel evaluation framework, the jp-algorithm, which introduces an alignment-based method to address challenges in preprocessing tasks like tokenization and sentence boundary detection (SBD). Traditional evaluation methods assume identical tokenization and sentence lengths between gold standards and system outputs, limiting their applicability to real-world data. The jp-algorithm overcomes these limitations, enabling robust end-to-end evaluations across a variety of NLP tasks. It enhances accuracy and flexibility by incorporating linear-time alignment while preserving the complexity of traditional evaluation metrics.
This memoir provides key insights into the processing of morphologically rich languages, such as Korean, while offering a generalizable framework for evaluating diverse end-to-end NLP systems. My contributions lay the foundation for future developments, with broader implications for multilingual resource development and system evaluation.

\singlespacing
\chapter*{Résumé}
\doublespacing

Ce mémoire explore deux aspects fondamentaux du traitement automatique des langues (TAL) : la création de ressources linguistiques et l’évaluation des performances des systèmes de TAL. Au cours de la dernière décennie, mes travaux se sont concentrés sur le développement d’un schéma d’annotation basé sur les morphèmes pour la langue coréenne, qui capture les propriétés linguistiques allant de la morphologie à la sémantique. Cette approche a permis d’obtenir des résultats à la pointe de l’état de l’art dans diverses tâches de TAL, y compris l’étiquetage morphosyntaxique, l’analyse en dépendances et la reconnaissance d’entités nommées. De plus, ce travail propose une analyse approfondie de la granularité de segmentation et de son impact crucial sur les performances des systèmes de TAL.
En parallèle du développement de ressources linguistiques, j’ai proposé un nouveau cadre d’évaluation, le jp-algorithme, qui introduit une méthode basée sur l’alignement pour traiter les défis liés aux tâches de prétraitement telles que la tokenisation et la détection des frontières de phrases. Les méthodes d’évaluation traditionnelles supposent une tokenisation et une longueur de phrases identiques entre les standards de référence et les sorties du système, limitant ainsi leur applicabilité aux données réelles. Le jp-algorithme surmonte ces limitations en permettant des évaluations robustes de bout en bout à travers une variété de tâches de TAL. Il améliore la précision et la flexibilité en intégrant un alignement en temps linéaire tout en préservant la complexité des métriques d’évaluation traditionnelles.
Ce mémoire apporte des perspectives clés sur le traitement des langues morphologiquement riches, telles que le coréen, tout en offrant un cadre généralisable pour l’évaluation de divers systèmes de TAL de bout en bout. Mes contributions posent les bases de futurs développements, avec des implications plus larges pour le développement de ressources multilingues et l’évaluation des systèmes.

\singlespacing
\chapter{Introduction}
\doublespacing

In this memoir, I bring together two intertwined concerns in Natural Language Processing (NLP): designing linguistic resources that support robust modeling, and evaluating model outputs under realistic variation in representation. Over the past decade, I have developed a morpheme-based annotation scheme that captures core linguistic properties of Korean across levels ranging from morphology to semantics, and this scheme has produced state-of-the-art results in several tasks. The same diversity in representation that makes such resources effective, however, makes evaluation fragile: many commonly used metrics require identical tokenization, implicitly assuming equal-length sentences and word sequences between gold-standard and system outputs. I therefore introduce an evaluation approach that performs alignment-based preprocessing of sentences and words while leaving the original metrics unchanged. The result preserves the established evaluation practice but extends it with a linear-time alignment step, enabling end-to-end evaluation for numerous NLP tasks that were previously difficult to assess in a principled way.

\section{Morpheme-based Language Annotation}

My research over the past decade has primarily focused on the development of linguistic resources (LRs), particularly for Korean. From 2011 to 2013, I developed syntactic analysis systems for Korean, utilizing both phrase structure and dependency parsing, in collaboration with KAIST, South Korea, and Kyoto University, Japan. This work extends directly from my PhD dissertation, which was grounded in tree-adjoining grammars \citep{joshi-levy-takahashi-1975-tree, joshi-schabes-1991-tree}, a formalism that models sentence structure through derived trees (constituency structure) and derivation trees (dependency structure).
The syntactic analysis systems I developed during that period achieved state-of-the-art (SOTA) performance in Korean parsing in both phrase structure parsing \citep{choi-park-choi:2012:SP-SEM-MRL} and dependency parsing \citep{park-EtAl:2013:IWPT}. I have since continued refining these systems, and they have consistently maintained SOTA performance for Korean in constituency parsing \citep{kim-park:2022} and dependency parsing \citep{chen-etal-2022-yet}.

Working with two distinct approaches to syntactic parsing for Korean led me to recognize a crucial aspect of language processing: the granularity of word segmentation. In constituency parsing, I proposed a morpheme-based structure in which terminal nodes represent the morphemes within a Korean word. In contrast, in my dependency parsing work during the 2010s, prior to the emergence and widespread adoption of the Universal Dependencies framework, I adopted the eojeol (a space-separated unit in Korean) as the basic unit of analysis. This choice reflected the dominant practice in Korean language processing at the time: most widely used systems and corpora, including the Sejong Corpus, treat the eojeol as the primary unit of analysis. Consequently, Korean dependency parsing systems of that period largely followed eojeol-based approaches, which I also incorporated into my work. At the same time, because Korean is an agglutinative language, the productive combination of content and functional morphemes within words motivates a finer-grained analysis. Under a morpheme-based representation, each morpheme is assigned a distinct part-of-speech tag and serves as a terminal node in a constituency parse tree, aligning Korean resources with the terminal structure assumed in many constituency treebanks for other languages.

Research in Korean language processing has consistently emphasized the representation of words through various component units, ranging from eojeols, a basic word-like unit, to more detailed morphological decompositions. These segmentation strategies have typically been justified on linguistic or technical grounds, with minimal consideration of alternative approaches. However, the choice of segmentation granularity critically affects the performance of algorithms in a wide range of tasks, such as part-of-speech (POS) tagging, syntactic parsing, and machine translation.
In response to these challenges, I have conducted an extensive analysis of the granularity levels previously proposed and employed in Korean language processing. Beginning with the application of these various granularity levels in dependency parsing \citep{park:2017:Depling}, I introduced a comprehensive framework for segmentation granularity in Korean, organizing it into five distinct levels from both linguistic and computational perspectives. Furthermore, I have presented the outcomes of these five segmentation levels across various tasks, evaluating their relative effectiveness in Korean language processing \citep{park-kim-2024-word}.

Through an in-depth analysis of how the Korean language has been represented across various language processing systems, I developed a new annotation method for morphologically segmented words, analogous to the annotation of multiword tokens (MWTs) in the CoNLL-U format. Using this novel annotation scheme, I collaborated with colleagues and supervised students in investigating various tasks for Korean, including part-of-speech (POS) tagging \citep{park-tyers:2019:LAW}, named entity recognition (NER) \citep{chen-lim-park-2024-korean}, dependency parsing \citep{chen-etal-2022-yet} and FrameNet parsing \citep{chen-etal-2024-towards-standardized}.
For each of these tasks, I re-annotated linguistic resources using the proposed method, achieving state-of-the-art (SOTA) results, which continue to represent the leading benchmarks in Korean language processing.

\section{Evaluation by Alignment}

Accurate evaluation is fundamental in NLP to assess the effectiveness of systems. Traditional methods, which compare system outputs with gold-standard references, have been useful in component-based systems but fail to address the complexities posed by modern end-to-end systems. These systems, which process entire tasks without relying on modular components, highlight the limitations of conventional evaluation, particularly in handling real-world data where sentence boundaries and tokenization may not align between system outputs and gold standards.

In response to these challenges, I proposed a novel alignment-based evaluation algorithm, termed the jp-algorithm (jointly preprocessed alignment-based evaluation algorithm), specifically designed to address mismatches that arise during preprocessing tasks such as tokenization and sentence boundary detection (SBD).
I began by integrating this evaluation method for preprocessing tasks, including tokenization and SBD, into my undergraduate course, with the support of the Centre for Teaching, Learning and Technology (CTLT) at the University of British Columbia (UBC), to redesign LING 242, \textit{Computational Tools for Linguistic Analysis}.\footnote{This course was initially designed for graduate students at the University of Arizona during 2016-2017, and I taught the same material at The State University of New York at Buffalo during 2018-2019. At UBC, I undertook the redesign of the course to adapt the material specifically for undergraduate students through the \textit{Students as Partners} program: \url{https://sap.ubc.ca/funded-projects/redesigning-ling242-computational-tools-for-linguistic-analysis/}}
Building on alignment techniques from machine translation (MT), this approach improves evaluation robustness and accuracy by accounting for inconsistencies between system outputs and gold standards. The proposed algorithm improves precision and recall, particularly in tasks such as constituent parsing, grammatical error correction (GEC) as well as preprocessing tasks, where traditional evaluation metrics rely on consistent tokenization and sentence segmentation.

I made three key contributions based on the concept of evaluation by alignment: (1) introducing an alignment-based method that enhances the evaluation of end-to-end NLP systems, (2) extending the method’s applicability across a range of NLP tasks, including preprocessing, parsing, and grammatical error correction (GEC), and (3) addressing the complexities of real-world writing structures, ensuring reliable evaluations despite variations in sentence boundaries and tokenization.
The jp-algorithm provides a more accurate, flexible, and robust evaluation framework, offering significant improvements over traditional methods in assessing NLP systems across diverse text inputs.

\section{Summary}
Following this \textit{Introduction} chapter, the remainder of this memoir is divided into two parts: Part I, How to \textit{Create} LRs for NLP, and Part II, How to \textit{Evaluate} results of NLP.  
These two parts will each appear as separate published volumes: Part I in my forthcoming monograph with the Academy of Korean Studies Press, and Part II in my forthcoming monograph in the \textit{Synthesis Lectures on Computer Science} series (Springer Nature).\footnote{Part I will be published as {인공지능과 언어 단위, 한국어 처리의 새로운 모색 -- 형태소 기반 접근을 중심으로} (\textit{Rethinking Linguistic Units in the Age of AI: A Morpheme-Based Approach to Korean Language Processing}), a 186-page volume re-written in Korean and published by the Academy of Korean Studies Press in South Korea. Part II will be published as \textit{Evaluation by Alignment -- A Framework for Robust End-to-End NLP Assessment}, in the \textit{Synthesis Lectures on Computer Science} series (Springer Nature, Switzerland), a 145-page monograph. Both volumes are forthcoming in 2025.}
In Part I, Chapter~\ref{word-segmentation-granularity-in-korean} addresses word segmentation granularity in Korean, based on my publication in \textit{Korean Linguistics} \citep{park-kim-2024-word}. This research began during my time at the University of Arizona in 2016-2017 and was published in 2024. It provides the theoretical foundation for morpheme-based language annotation for Korean.
Chapter~\ref{pos-to-framenet} expands on my collaborative work with Yige Chen (The Chinese University of Hong Kong, Hong Kong) and KyungTae Lim (Korea Advanced Institute of Science \& Technology). This chapter originates from my initial work on POS tagging \citep{park-tyers:2019:LAW}, where I proposed a morpheme-based annotation scheme. Yige Chen extended this annotation approach to named entity recognition \citep{chen-lim-park-2024-korean}, dependency parsing \citep{chen-etal-2022-yet} and FrameNet parsing \citep{chen-etal-2024-towards-standardized}.
In Part II, Chapter~\ref{evaluation-by-alignment} introduces the jp-algorithm, detailing the alignment-based evaluation approach. 
This research was initially supported by the Centre for Teaching, Learning and Technology (CTLT) through collaborations with students at The University of British Columbia (UBC), including Eunkyul Leah Jo (Department of Computer Science) and Angela Yoonseo Park (Department of Linguistics). It was also supported by the Academy of Korean Studies (AKS), which provided funding to hire Junrui Wang as my research assistant during his master’s studies at UBC.
In Chapter~\ref{applications-of-jp-algorithms}, I developed applications of the jp-algorithm with the assistance of Eunkyul Leah Jo and Angela Yoonseo Park for jp-preprocessing \citep{jo-etal-2024-untold-story,park-etal-2024-jp} and jp-evalb \citep{jo-etal-2024-novel}, and with Junrui Wang for jp-errant \citep{wang-etal-2025-refined}.
Finally, I will present my future research plans in the \textit{Conclusion} chapter.

\part{How to \textit{Create} LRs for NLP}

\singlespacing
\chapter{Word Segmentation Granularity in Korean} \label{word-segmentation-granularity-in-korean}
\chaptermark{Segmentation Granularity}
\doublespacing

\section{Introduction} 

Morphological analysis for Korean has been based on an eojeol, which has been considered as a basic segmentation unit in Korean delimited by white blank spaces in a sentence. Almost all of the language processing systems and language data sets previously developed for Korean have utilized this eojeol as a fundamental unit of analysis. Given that Korean is an agglutinative language, joining content and functional morphemes of words is very productive and the number of their combinations is exponential. We can treat a given noun or verb as a stem (also content) followed by several functional morphemes in Korean. Some of these morphemes can, sometimes, be assigned its syntactic category. Let us consider the sentence in~\eqref{ungaro}.

The corresponding morphological analysis is also provided in Figure~\ref{ungaro-pos}. 
\textit{Unggaro} (`Ungaro') is a content morpheme (a proper noun) and a postposition \textit{-ga} (nominative) is a functional morpheme. They form together a single eojeol (or word) \textit{unggaro-ga} (`Ungaro+\textsc{nom}'). 
For the sake of convenience, we add a figure dash (\textit{-}) at the beginning of functional morphemes, such as \textit{-ga} (\textsc{nom}) to distinguish between content and functional morphemes. The nominative case markers \textit{-ga} or \textit{-i} may vary depending on the previous letter --- vowel or consonant. A predicate \textit{naseo-eoss-da} also consists of the content morpheme \textit{naseo} (`become') and its functional morphemes, \textit{-eoss} (`\textsc{past}') and \textit{-da} (`\textsc{decl}'), respectively.

\begin{exe}
 \ex \label{ungaro}
 \glll
 {프랑스의}    {세계적인}    의상    디자이너    엠마누엘    {웅가로가}    실내    장식용    {직물}    디자이너로    {나섰다.} \\ 
 \textit{peurangseu-ui} {\textit{segye-jeok-i-n}} \textit{uisang} \textit{dijaineo} \textit{emmanuel} {\textit{unggaro-ga}} \textit{silnae} \textit{jangsik-yong   } \textit{jikmul   } \textit{dijaineo-ro} \textit{{naseo-eoss-da}.}\\ 
 France-\textsc{gen} world~class-\textsc{rel} fashion designer Emanuel Ungaro-\textsc{nom} interior    decoration textile designer-\textsc{ajt} become-\textsc{past}-\textsc{decl}\\
 \trans 'The world-class French fashion designer Emanuel Ungaro became an interior textile designer.' 
\end{exe}

\begin{figure}[!ht]
\begin{center}
\footnotesize{
\begin{tabular} {rlcl l}
프랑스의 & \textit{peurangseu-ui} & &\textit{peurangseu}/NNP+\textit{ui}/JKG & France-\textsc{gen}\\
세계적인 &\textit{segye-jeok-i-n} && \textit{segye}/NNG+\textit{jeok}/XSN+\textit{i}/VCP+\textit{n}/ETM & world~class-\textsc{rel}\\
의상 &\textit{uisang} &&\textit{uisang}/NNG & fashion\\
디자이너 &\textit{dijaineo} && \textit{dijaineo}/NNG & designer\\
엠마누엘 &\textit{emmanuel} && \textit{emmanuel}/NNP & Emanuel\\
웅가로가 &\textit{unggaro-ga} && \textit{unggaro}/NNP+\textit{ga}/JKS& Ungaro-\textsc{nom}\\
실내 &\textit{silnae} &&\textit{silnae}/NNG& interior\\
장식용 &\textit{jangsikyong} &&\textit{jangsikyong}/NNG& decoration\\
직물 &\textit{jikmul} &&\textit{jikmul}/NNG& textile\\
디자이너로 &\textit{dijaineo-ro} && \textit{dijaineo}/NNG+\textit{ro}/JKB& designer-\textsc{ajt}\\
나섰다.& \textit{naseo-eoss-da.} & &\textit{naseo}/VV+\textit{eoss}/EP+\textit{da}/EF+./SF& become-\textsc{past}-\textsc{decl} \\
\end{tabular}}
\end{center}
\caption{
Morphological analysis and part of speech (POS) tagging example in the Sejong corpus: NN* are nouns, JK* are case makers and postpositions, V* are verbs, and E* are verbal endings.\label{ungaro-pos}}
\end{figure}

Every approach for Korean language processing has decided how to separate \textit{sequences} of morphemes into component parts, ranging from eojeols, a basic word-like unit, all the way down to a complete morphological parse. These decisions have been, for the most part, argued as either linguistically or technically motivated, with little or no interest in exploring an alternative. The choice does have some impact on the performance of algorithms in various tasks, such as part of speech (POS) tagging, syntactic parsing and machine translation. In the study, we analyze different granularity levels previously proposed and utilized for Korean language processing. In accordance with these analyzing works, we present the results of language processing applications using different segmentation granularity levels for future reference. To the best of the authors' knowledge, this is the first time that different granularity levels in Korean have been compared and evaluated against each other. This would contribute to fully understanding the current status of various granularity levels that have been developed for Korean language processing. Specifically the main goal of this section is to diagnose the current state of natural language processing in Korean by tracing its development procedures and classifying them into five steps. Additionally, this section aims to clearly explicate and evaluate the challenges unique to Korean language processing, with the objective of contributing to the improvement of various methodologies in this field. 
To this end, after presenting previous work in Section~\ref{previous-work}, the study introduces the segmentation granularity in Korean by classifying them into five different levels with a linguistic perspective as well as a natural language processing perspective in Section~\ref{definition}, and presents several application for Korean language processing using the five segmentation granularity levels by comparing them each other in Section~\ref{applications}. Finally, Section~\ref{chapter2-conclusion} concludes the discussion.

\section{Previous Work} \label{previous-work}

Different granularity levels have been proposed mainly due to varying different syntactic analyses in several previously proposed Korean treebank datasets: KAIST \citep{choi-EtAl:1994}, Penn \citep{han-EtAl:2002}, and Sejong. While segmentation granularity, which we deal with, is based on morphological analysis, the syntactic theories are implicitly presented in the corpus for Korean words. 
Figure~\ref{syntactic-granularity} summarizes the syntactic trees which can be represented in Korean treebanks for different segmentation granularity levels. Korean TAG grammars \citep{park:2006} and CCG grammars \citep{kang:2011} described Korean by separating case markers. Most work on language processing applications such as phrase structure parsing and machine translation for Korean which uses the sentence by separating all morphemes \citep{choi-park-choi:2012:SP-SEM-MRL,park-hong-cha:2016:PACLIC,kim-park:2022}. The Penn Korean treebank introduced a tokenization scheme for Korean, while the KAIST treebank separates functional morphemes such as postpositions and verbal endings. Note that there are no functional tags (\textit{i.e.}, \texttt{-sbj} or \texttt{-ajt}) in the KAIST treebank.

\begin{figure}[!ht]
\centering

\begin{subfigure}[b]{0.39\textwidth}
\resizebox{\textwidth}{!}{
\synttree 
[S
	[NP-SBJ
	[NP
	[VNP-MOD [\textit{segye-jeok-i-n}] ]
[NP [.x $\cdots$] ]]
		[NP-SBJ [\textit{unggaro-ga}]] ]
	[VP
		[NP-AJT[.x $\cdots$ ]]
		[VP [ \textit{naseo-eoss-da$.$} ]]]]
}
\caption{Sejong{-style} treebank} 
\label{sejong-treebank}
\end{subfigure}
\begin{subfigure}[b]{0.51\textwidth}
\resizebox{\textwidth}{!}{
\synttree 
[S[NP-SBJ[NP [S [WHNP-1 [$*$op$*$]] [S [NP-SBJ [$*$T$*$-1]] [VP [NP [\textit{segye-jeok-i-n}]]]]] [NP [.x $\cdots$ ]] ][NP-SBJ [\textit{unggaro-ga}]]] [VP[NP-COMP[.x $\cdots$ ]] [VP [\textit{naseo-eoss-da}]]] [ $\cdot$ ]]
}
\caption{Penn{-style} Korean treebank} 
\label{penn-treebank}
\end{subfigure}

\begin{subfigure}[b]{0.44\textwidth}
\resizebox{\textwidth}{!}{
\synttree 
[S [S [NP-SBJ [NP-SBJ[NP[VNP-MOD [\textit{segye-jeok-i-n}]] [NP [.x $\cdots$ ]] ][NP-SBJ [\textit{unggaro}]] ] [\textit{-ga}]] [VP[NP-AJT[.x $\cdots$ ]] [VP [\textit{naseo-eoss-da}]]] ] [ $\cdot$ ]]
}
\caption{Korean tree-adjoining grammar}
\label{korean-ltag}
\end{subfigure}
\begin{subfigure}[b]{0.49\textwidth}
\resizebox{\textwidth}{!}{
\synttree 
[S[VP [NP[NP [VP [\textit{segye-jeok -i}] ] [\textit{-n}] [NP [.x $\cdots$ ]] ][NP [\textit{unggaro}] ]] [\textit{-ga}] [VP[NP[.x $\cdots$ ]] [VP [\textit{naseo}] ] ] [AUXP [\textit{-eoss}] ]] [ \textit{-da} ][ $\cdot$ ]]
}
\caption{KAIST{-style} treebank}
\label{kaist-treebank}
\end{subfigure}

\begin{subfigure}[b]{0.65\textwidth}
\resizebox{\textwidth}{!}{
\synttree 
[S[NP-SBJ[NP[VNP-MOD [\textit{segye}] [\textit{-jeok}] [\textit{-i}] [\textit{-n}]] [NP [.x $\cdots$ ] ] ][NP-SBJ [\textit{unggaro}] [-ga]]] [VP[NP-AJT[.x $\cdots$ ] ] [VP [\textit{naseo}] [\textit{-eoss}] [\textit{-da}] [ $\cdot$]]]]
}
\caption{Phrase structure parsing {structure}}
\label{phrase-structure-parsing-treebank}
\end{subfigure}

\caption{{Different} syntactic analyses using different segmentation granularities. POS labels are omitted. } 
\label{syntactic-granularity}
\end{figure}

Syllable-based granularity (\textit{e.g.}, \textit{se $\sqcup$ gye $\sqcup$ jeok $\sqcup$ i}, `world-class') \citep{yu-EtAl:2017:SCLeM,choi-EtAl:2017:SCLeM} and even character-based granularity using the Korean alphabet (\textit{s $\sqcup$ e $\sqcup$ g $\sqcup$ ye $\sqcup$ j $\sqcup$ eo $\sqcup$ k $\sqcup$ i}) \citep{stratos:2017:EMNLP,song-park:2020:TALLIP} have also been proposed  where $\sqcup$ indicates a blank space. They incorporate sub-word information to alleviate data sparsity especially in neural models. Dealing with sub-word level granularity using syllables and characters does not consider our linguistic intuition. We describe granularity based on a linguistically motivated approach in this section, in which each segmentation is a meaningful morphological unit.

\section{Definition of Segmentation Granularity} \label{definition}

The annotation guidelines for Universal Dependencies stipulate each syntactic word, which is an atom of syntactic analysis, as a basic unit of dependency annotation \citep{nivre-etal-2020-universal}. This stipulation presupposes that there must be a separate autonomous level of syntax and morphology. One of the features of agglutinative languages is that there is a one-to-one mapping between suffixes and syntactic categories, indicating that each suffix must have an appropriate syntactic category to which it belongs. More specifically, nouns can have individual markers indicating case, number, possessive, etc., whose orders are fixed. Thus, we can regard any given noun or verb as a stem followed by several inflectional or derivational morphemes. The number of slots for a given part of a category may be pretty high. In addition, an agglutinating language adds information such as negation, passive voice, past tense, honorific degree to the verb form. That is, in an agglutinating language, verbal morphemes are added to the root of the verb to convey various grammatical features, such as negation, passive voice, past tense, and honorific degree. One of the characteristics in Korean is to use a system of honorifics to express the hierarchical status and familiarity of participants in conversation with respect to the subject, object or the interlocutor. This system plays a great role in Koreamar. When a speaker uses honorific expression, we can figure out the social relationship between the speaker, the interlocutor, and the object in the subject position at the same time. This honorific system is reflected in the honorific markers attached to the nouns, and verbal endings to the verb.

Such a complex and rich morphological system in agglutinative languages poses numerous challenges for natural language processing. The key obstacle lies in a voluminous number of word forms that can be derived from a single stem. Word form analysis involves breaking a given surface word form into a sequence of morphemes in an order that is admissible in Korean. However, several difficulties may arise in dividing these sequences of morphemes into appropriate units. This section describes the segmentation granularity procedures that could influence the performance of algorithms in various tasks and the various analyses that have been adopted in Korean. First of all, we define five different levels of segmentation granularity for Korean, which have been independently proposed in previous work as different segmentation units. While Levels 1 (eojeols as they are), 2 (tokenization – separating words and symbols process), and 5 (separating all morphemes process) are due to technical reasons, Levels 3 (separating case markers process) and 4 (separating verbal endings process) are based on linguistic intuition.

\subsection{Level 1: eojeols} \label{l1}

As described previously, most Korean language processing systems and corpora have used the eojeol as a fundamental unit of analysis. For example, the Sejong corpus, the most widely-used corpus for Korean, uses the eojeol as the basic unit of analysis.\footnote{The Ministry of Culture and Tourism in Korea launched the 21st Century Sejong Project in 1998 to promote Korean language information processing.
The project has its name from Sejong the Great who conceived and led the invention of \textit{hangul}, the writing system of the Korean language.} The Sejong corpus was first released in 2003 and was continually updated until 2011. The project produced the largest corpus for the Korean language. It includes several types of corpora: historical, contemporary, and parallel text. Contents of the Sejong corpus represent a variety of sources: newswire data and magazine articles on various subjects and topics, several book excerpts, and scraped texts from the Internet. The Sejong corpus consists of the morphologically (part of speech tagged), the syntactically (treebank), and the lexical-semantically annotated text as well as a list of Korean words as dictionaries based on part of speech categories. Figure~\ref{sejong-corpus} shows an example of the Sejong corpus for the sentence in \eqref{ungaro}.

\begin{figure}[!ht]
\begin{subfigure}[b]{\textwidth}
\centering
\scriptsize{
\begin{tabular}{lll}
BTAA0001-00000012 &프랑스의& 프랑스/NNP+의/JKG \\
BTAA0001-00000013 &세계적인& 세계/NNG+적/XSN+이/VCP+ㄴ/ETM \\
BTAA0001-00000014 &의상& 의상/NNG \\
BTAA0001-00000015 &디자이너& 디자이너/NNG \\
BTAA0001-00000016 &엠마누엘& 엠마누엘/NNP \\
BTAA0001-00000017 &웅가로가& 웅가로/NNP+가/JKS \\
BTAA0001-00000018 &실내& 실내/NNG \\
BTAA0001-00000019 &장식용& 장식/NNG+용/XSN \\
BTAA0001-00000020 &직물& 직물/NNG \\
BTAA0001-00000021 &디자이너로& 디자이너/NNG+로/JKB \\
BTAA0001-00000022 &나섰다.& 나서/VV+었/EP+다/EF+./SF \\
\end{tabular}
}
\caption{Morphologically (part of speech tagged) analyzed corpus where the word is analyzed as in the surface form. Therefore, even the punctuation mark is a part of the word.} 
\label{sejong-morph-corpus}
\end{subfigure}

\begin{subfigure}[b]{\textwidth}
\centering
\scriptsize{
\begin{tabular}{lll llll}
\multicolumn{7}{l}{}\\
(S &(NP-SBJ &(NP &\multicolumn{4}{l}{(NP-MOD 프랑스/NNP+의/JKG)}\\
&&&(NP &\multicolumn{3}{l}{(VNP-MOD 세계/NNG+적/XSN+이/VCP+ㄴ/ETM)}\\
&&&&(NP &\multicolumn{2}{l}{ (NP 의상/NNG) }\\
&&&&& (NP 디자이너/NNG)))) \\
&&(NP-SBJ &\multicolumn{4}{l}{(NP 엠마누엘/NNP)} \\
&&&\multicolumn{3}{l}{(NP-SBJ 웅가로/NNP+가/JKS)))} \\
& (VP& (NP-AJT& (NP& (NP&\multicolumn{2}{l}{(NP 실내/NNG)} \\
& &&&& \multicolumn{2}{l}{(NP 장식/NNG+용/XSN))} \\
& &&& \multicolumn{3}{l}{(NP 직물/NNG)) }\\
& && \multicolumn{4}{l}{(NP-AJT 디자이너/NNG+로/JKB))}\\
& & \multicolumn{5}{l}{(VP 나서/VV+었/EP+다/EF+./SF)))}\\
\end{tabular}
}
\caption{Syntactically analyzed corpus (treebank) where it inherits the annotation from the morphologically (part of speech tagged) analyzed corpus and added bracketing syntactic tree structure.} 
\label{sejong-treebank-figure1}
\end{subfigure}

\begin{subfigure}[b]{\textwidth}
\centering
\scriptsize{
\begin{tabular}{lll}
\multicolumn{3}{l}{}\\
BSAA0001-00000012 &프랑스의& 프랑스/NNP+의/JKG \\
BSAA0001-00000013 &세계적인& 세계\_\_02/NNG+적/XSN+이/VCP+ㄴ/ETM \\
BSAA0001-00000014 &의상& 의상\_\_01/NNG \\
BSAA0001-00000015 &디자이너& 디자이너/NNG \\
BSAA0001-00000016 &엠마누엘& 엠마누엘/NNP \\
BSAA0001-00000017 &웅가로가& 웅가로/NNP+가/JKS \\
BSAA0001-00000018 &실내& 실내/NNG \\
BSAA0001-00000019 &장식용&  장식\_\_05/NNG+용/XSN \\
BSAA0001-00000020 &직물& 직물/NNG \\
BSAA0001-00000021 &디자이너로& 디자이너/NNG+로/JKB \\
BSAA0001-00000022 &나섰다.& 나서/VV+었/EP+다/EF+./SF \\
\end{tabular}
}
\caption{Semantically analyzed corpus where it includes the lexical semantic annotation to disambiguate the sense of the word by the Sejong dictionary.} 
\label{sejong-semantics}
\end{subfigure}

\caption{Example of the Sejong corpus for the sentence in \eqref{ungaro}}
\label{sejong-corpus}
\end{figure}

We define eojeols, as in the Sejong corpus, as granularity Level 1. Rationale of this segmentation granularity in Korean language processing is simply to use the word as it is in the surface form, in which the word is separated by a blank space in the sentence (that is, in a manner of what you see is what you get). Most morphological analysis systems have been developed based on eojeols (Level 1) as input and can yield morphologically analyzed results, in which a single eojeol can contain several morphemes. The dependency parsing systems described in \citet{oh-cha:2013} and \citet{park-EtAl:2013:IWPT} used eojeols as an input token to represent dependency relationships between eojeols. 
\citet{oh-EtAl:2011} {presented a system which predict phrase-level syntactic label for eojeols based on the sequence of morphemes in the eojeol.}
What is the most interesting is that \citet{petrov-das-mcdonald:2012:LREC} proposed Universal POS tags for Korean based on the eojeol and \citet{stratos-collins-hsu:2016:TACL} worked on POS tagging accordingly. Taking these basic trends into consideration, the study defines eojoeols as Level 1. Recently released KLUE (Korean Language Understanding Evaluation) also used the eojeol as a fundamental unit of analysis \citep{park-etal-2021-klue}.\footnote{\url{https://klue-benchmark.com}}

\subsection{Level 2: separating words and symbols} \label{l2}

The process of tokenization in the Korean language has often been overlooked, primarily because eojeols has traditionally been used as the basic unit of analysis. However, it has come to our attention that certain corpora have started adopting an English-like tokenization approach, which results in preprocessed words within these corpora. For example, the Penn Korean treebank \citep{han-EtAl:2002}, which punctuation marks are separated from words.\footnote{While the Penn Korean treebank separates all punctuation marks, quotation marks are the only symbols that are separated from words in the Sejong treebank {to distinguish between the quoted clause and the main sentence in the tree structure.}
We also note that among the existing corpora for Korean, only the Sejong treebank separates quotation marks from the word. 
Other Sejong corpora including the morphologically analyzed corpus do not separate the quotation marks, and still use the eojeol as a basic analysis unit.} 
This segmentation granularity especially in the Penn-treebank style corpus focuses on multilingual processing where Penn treebanks include English \citep{marcus-etal-1993-building,taylor-marcus-santorini:2003}, Chinese \citep{xue-etal-2005-ctb}, Arabic \citep{maamouri-bies:2004} and Korean \citep{han-EtAl:2002}. The Penn Korean treebank follows the tokenization scheme that has been used in the other language of the Penn treebanks, as shown in Figure~\ref{penn-figure2}. The most distinctive feature in Level 2 lies in that the punctuation mark is all separated from the original word (tokenized).

\begin{figure}[!ht] 
\centering
\footnotesize{
\begin{tabular}{llll llll}
(S& (NP-SBJ& \multicolumn{6}{l}{그/NPN+은/PAU)} \\
  & (VP& (S-COMP& (NP-SBJ& \multicolumn{3}{l}{르노/NPR+이/PCA)} \\
&&&(VP& (VP& (NP-ADV& \multicolumn{2}{l}{3/NNU}\\
&&&&&& \multicolumn{2}{l}{월/NNX+말/NNX+까지/PAU)}\\
&&&&& (VP& (NP-OBJ& \multicolumn{1}{l}{인수/NNC+제의/NNC}\\
&&&&&    &        & 시한/NNC+을/PCA)\\
&& & & & & \multicolumn{2}{l}{갖/VV+고/ECS))} \\
&&&&\multicolumn{4}{l}{있/VX+다/EFN+고/PAD))} \\
&&\multicolumn{6}{l}{덧붙이/VV+었/EPF+다/EFN)}\\
&\multicolumn{6}{l}{./SFN)}\\
\end{tabular}
}
\caption{Example of the Penn Korean treebank where the punctuation mark is separated from the word (tokenized): N* are nouns PA* are case markers and postpositions, V* are verbs, and E* are verbal endings.}
\label{penn-figure2}
\end{figure}

We define the tokenization by separating words and symbols as a granularity Level 2. 
\citet{chung-gildea:2009:EMNLP} used a granularity Level 2 for a baseline tokenization system for a machine translation system from Korean into English where they proposed an unsupervised tokenization method to improve the machine translation result. Figure~\ref{penn-figure2} illustrates that the punctuation marker has been separated from the verb \textit{deosbut-i-eoss-da} (`added') and assigned its own category with the marker being designated as `texttt{sfn}' in Penn Treebank. In addition, the tokenization schema of the sentence follows the method similar to the English language. That is, syntactic unit \textit{3-wol-mal-kka-ji} (`until the end of March') {is traditionally treated} as one eojeol, but in Level 2, this unit is tokenized as three different units such as \textit{3-wol} (`March'), \textit{mal} (end) and \textit{kka-ji} (`until'), which is tokenized identically to that of English such as until the end of March. As mentioned in Level 1, most Korean language processing systems have used an eojeol as their basic unit of analysis, resulting in a single eojeol involved with several different morphemes, which is a prominent feature in Level 1. According to this principle, we can easily identify that the noun phrase in a subject position \textit{geu-eun} forms one eojeol consisting of a stem \textit{geu} and a {topic} marker \textit{eun}. In the same way, the verb phrase \textit{deosbut-i-eoss-da} (`added') creates one eojeol with a root \textit{deosbut}, a {passive suffix} \textit{-i}, a past tense marker \textit{-eoss} and verb ending marker \textit{-da}.

\citet{park-EtAl:2014:ISWCPOSTER} also used this granularity to develop Korean FrameNet lexicon units by using the cross-lingual projection method from the Korean translation of the English Propbank \citep{palmer-gildea-kingsbury:2005:CL}. Universal Dependencies \citep{nivre-EtAl:2016:LREC,nivre-etal-2020-universal} contains two Korean dependency treebanks, namely the GSD treebank \citep{mcdonald-etal-2013-universal} and the KAIST treebank \citep{choi-EtAl:1994,chun-EtAl:2018:LREC}, which also use the tokenization scheme by separating words and punctuation marks.

{Recently, \citet{park-kim-2023-role} insisted that the functional morphemes in Korean should be treated as part of a word in Korean categorial grammars, with the result that their categories for detailed morphemes do not require to be assigned individually in a syntactic level, and also that it would be more efficient to assign the syntactic categories on the fully inflected lexical word derived by the lexical rule of the morphological processes in the lexicon.}

\subsection{Level 3: separating case markers} \label{l3}

From a purely linguistic perspective, postpositions as functional morphemes in Korean convey grammatical cases (\textit{e.g.}, nominative or accusative), adverbial relations (spatial, temporal or directional), semantic roles and conjunctives by combining with the lexical words. We may separately indicate them as case marker, adverbial postposition, auxiliary postposition, and conjunctive postposition, respectively, though we generally term them as postpositions or case markers, depending on the authors. In linguistics, a marker also refers to a free or bound morpheme indicating the grammatical function of the word, phrase or sentence. For the sake of convenience, this section uses case markers as a term for covering them. Case markers are immediately attached following a noun or pronoun. They are used to indicate the grammatical roles of a noun in a sentence such as subject, object, complement or topic.

First of all, \textit{-i} and \textit{-ga} are nominative case markers whose form depends on whether the stem ends with a vowel or consonant. When the honorific subject is used, this nominative case marker will be replaced by the honorific marker \textit{-kkeseo}, instead of \textit{-i} or \textit{-ga}. An honorific is marked to encode the relative social status of the interlocutors. A major feature of this honorific system is typically to convey the same message in both honorific and familiar forms. Korean honorifics are added to nouns, verbs, and adjectives. Similarly to this nominative case marker, the honorific dative case marker \textit{-kke} will be used instead of the familiar dative case marker \textit{-ege}. The rest of the markers are used to express the adverbial relations such as directional, temporal, spatial including source and destination, and accompaniment.
All of these markers attached to the noun stem cannot be duplicated, showing {complementary distribution}. As shown in the example \eqref{holangi-ga}, the nominative case marker \textit{-ga} cannot be together with the instrumental case marker \textit{-ro} in \eqref{holangi-ga-ro}, and cannot collocate also with the dative case marker \textit{-ege} in \eqref{holangi-ga-ege}.

\begin{exe} 
\ex \label{holangi-ga}
    \begin{xlist}
        \ex \gll {\textit{holangi-ga}} {\textit{sanab-da}}. \\
            {tiger-\textsc{nom}}  {fierce-\textsc{decl}}  \\
            \trans 'A tiger is fierce.'        
        \ex \label{holangi-ga-ro}  
            \gll *\textit{holangi-ga-ro} \textit{sanab-da}.\\
            {tiger-\textsc{nom}-\textsc{\textit{`to'}}}  {fierce-\textsc{decl}}  \\
            \trans 'A tiger is fierce.'
        \ex \label{holangi-ga-ege} 
            \gll *\textit{holangi-ga-ege} \textit{sanab-da}.\\
            {tiger-\textsc{nom}-\textsc{dat}}  {fierce-\textsc{decl}}  \\
            \trans 'A tiger is fierce.'      
    \end{xlist}
\end{exe}

Under a perspective of natural language processing, the Sejong corpus has been criticized for the scope of the case marker, in which only a final noun (usually the lexical anchor) in the noun phrase is a modifier of the case marker. For example, \textit{Emmanuel Ungaro-ga} in the Sejong corpus is annotated as (NP (NP \textit{Emmanuel}) (NP \textit{Ungaro-ga})), in which only \textit{Ungaro} is a modifier of \textit{-ga} (`\textsc{nom}'). 
For example as described in \citet{ko:2010}, while there are several debates on whether a noun or a case marker is a modifier in Korean, this is beyond the scope of the section. The Penn Korean treebank does not explicitly represent this phenomenon. 
It just groups a noun phrase together: \textit{e.g.}, (NP \textit{Emmanuel Ungaro-ga}), which seems to be treated superficially as a simple compound noun phrase. Collins' preprocessing for parsing the Penn treebank adds intermediate NP terminals for the noun phrase \citep{collins:1997:ACL,bikel:2004:CL}, and so NPs in the Penn Korean treebank will have a similar NP structure to the Sejong corpus \citep{chung-post-gildea:2010:SPMRL}. 
To fix the problem in the previous treebank annotation scheme, there are other annotation schemes in the corpus and lexicalized grammars. They are introduced to correctly represent the scope of the case marker. \citet{park:2006} considered case markers as independent elements within the formalism of Tree adjoining grammars \citep{joshi-levy-takahashi-1975-tree}. Therefore, he defined case markers as an auxiliary tree to be adjoined to a noun phrase. In contrast to case markers, verbal endings in the inflected forms of predicates are still in the part of the eojoel and they are represented as initial trees for Korean TAG grammars. The lemma of the predicate and its verbal endings are dealt with as inflected forms instead of separating functional morphemes \citep{park:2006}. 
This idea is going back to Maurice Gross's lexicomars in 1970s \citep{gross:1975} and his students who worked on a descriptive analysis of Korean in which the number of predicates in Korean could be fixed by generating all possible inflection forms: \textit{e.g.}, \citet{pak:1987,nho:1992,nam:1994,shin:1994,park:1996,chung:1998,han:2000}.

\subsection{Level 4: separating verbal endings} \label{l4}

With a purely linguistic perspective, Korean verbs are formed in terms of the agglutinating process by adding various endings to the stem. Korean is widely known to have a great many verbal endings between this stem and final verbal endings. More specifically, the verbal endings in Korean are well known to be complex in their syntactic structures in the sense that the verbal endings carry much of functional load in the grammatical aspects such as sentence mood, tense, voice, aspect, honorific, conjunction, etc.: 
for example, 
\textit{inter alia}, 
{tense \citep{byungsun:2003}, 
grammatical voice \citep{chulwoo:2007}, 
interaction of tense–aspect–mood marking with modality \citep{jaemog:1998}, 
evidentiality \citep{donghoon:2008}, and 
interrogativity \citep{donghoon:2011}.}
More additional endings can be used to denote various semantic connotations. That is, a huge number of grammatical functions are achieved by adding various verbal endings to verbs. 
The number can also vary depending on the theoretical analyses, naturally differing in their functions and meanings. These endings, of course, do not change the argument structures of a predicate. A finite verb in Korean can have up to seven suffixes as its endings, whose order is fixed. As mentioned in the previous section, the Korean honorific system can also be reflected in verbs with honorific forms. When a speaker expresses his respect toward the entities in a subject or indirect object position, the honorific marker \textit{-(eu)si} is attached to the stem verb, thereby resulting in the verb form \textit{sanchaegha} (`take a walk'). The suffixes denoting tense, aspect, modal, formal, mood are followed by the honorific.

Unlike the markers attached to nouns, {Korean verbal endings are added to the verb stem in a specific order, depending on the tense, mood, and politeness level of the sentence,} as illustrated in \eqref{halabeoji}. The verb stem \textit{sanchaegha} (`to take a walk') can be followed by the honorific \textit{-si} in \eqref{halabeoji-sanchaegha-si-n-da}. 
The two suffixes indicating an honorific and past tense can be attached to the verb stem in \eqref{halabeoji-sanchaegha-sy-eoss-da}. 
One more additional suffix of retrospective aspect is added in the example in \eqref{halabeoji-sanchaegha-sy-eoss-deon-jeog-i-iss-da}. 
If the order of a past suffix and honorific suffix is changed in the verbal endings, the sentence would be ungrammatical, as in \eqref{halabeoji-kkeseo-sanchaegha-eoss-sy-deon-jeog-i-iss-da}.

\begin{exe}
\ex \label{halabeoji}
\begin{xlist}
\ex \label{halabeoji-sanchaegha-si-n-da}
\gll \textit{halabeoji-kkeseo}  \textit{jamsi} \textit{sanchaegha-si-n-da}.\\
{Grandfather-\textsc{nom-hon}} {for~a~while}  {take~a~walk-\textsc{hon}-\textsc{pres}-\textsc{decl}}   \\
\trans 'Grandfather takes a walk for a moment.'

\ex \label{halabeoji-sanchaegha-sy-eoss-da}  
\gll \textit{halabeoji-kkeseo} \textit{jamsi} \textit{sanchaegha-sy-eoss-da}.\\
{Grandfather-\textsc{hon}} {for a while} {take~a~walk-\textsc{hon}-\textsc{past}-\textsc{decl}}\\
\trans 'Grandfather took a walk for a moment.'

\ex \label{halabeoji-sanchaegha-sy-eoss-deon-jeog-i-iss-da}
\gll \textit{halabeoji-kkeseo} \textit{jamsi} \textit{sanchaegha-sy-eoss-deon} \textit{jeog-i} \textit{iss-da}.\\
Grandfather-\textsc{hon} {once} {go~for~a~walk-\textsc{hon}-\textsc{past}-\textsc{asp}} {experience-\textsc{cop}} {be-\textsc{decl}} \\
\trans 'Grandfather once went for a walk for a moment.'

\ex  \label{halabeoji-kkeseo-sanchaegha-eoss-sy-deon-jeog-i-iss-da}
\gll *\textit{halabeoji-kkeseo} \textit{jamsi} \textit{sanchaegha-eoss-sy-deon} \textit{jeog-i} \textit{iss-da}.\\
Grandfather-\textsc{hon} {once} {go~for~a~walk-\textsc{past}-\textsc{hon}-\textsc{asp}}  {experience-\textsc{cop}} {be-\textsc{decl}}   \\
\trans 'Grandfather once went for a walk for a moment.'
\end{xlist}
\end{exe}

Government and Binding (GB) theory \citep{chomsky:1981,chomsky:1982} for Korean syntactic {analyses}, in which the entire sentence depends on verbal endings as described in Figure \ref{kaist-ex} for \textit{naseo-eoss-da} (`became'). This means that the functional morpheme \textit{-eoss} is assigned its own syntactic category T(ense) and the verbal ending \textit{-da} C(omplimentizer) attached in the final position determines the whole syntactic category CP in Korean.

\begin{figure}[!ht]
    \centering
\footnotesize{    
\begin{forest}
[CP [C\texttt{'} [IP   [NP [$\cdots$] ] [I\texttt{'} [VP [$\cdots$] 
[V  [\textit{naseo}] ]] [I [T [\textit{-eoss}] ]]]] [C [\textit{-da}] ]]]
\end{forest}  
}
    \caption{{GB theory for Korean syntactic analyses, in which the entire sentence depends on verbal endings}}
    \label{kaist-ex}
\end{figure}

From the Natural Language Processing perspective, the KAIST treebank \citep{choi-EtAl:1994}, an earliest Korean treebank, introduced this type of analysis, which is Level 4. It is the granularity Level 4 that we adapt the KAIST treebank representation. While the KAIST treebank separates case markers and verbal endings with their lexical morphemes, punctuation marks are not separated and they are still a part of preceding morphemes as represented in the Sejong treebank. Therefore, strictly speaking, one could judge that the KAIST treebank is not granularity Level 4 by our definition because we separate punctuation marks. In addition, while it also represents derivational morphology in the treebank annotation (\textit{i.e.}, for a copula \textit{segye-jeok $\sqcup$ -i $\sqcup$ -n} (`world-class') in the KAIST treebank), we separate only verbal endings (\textit{i.e.}, \textit{segye-jeok-i $\sqcup$ -n}).

\subsection{Level 5: separating all morphemes} \label{l5}

Many downstream applications for Korean language processing are based on the granularity Level 5, in which all morphemes are separated: 
POS tagging \citep{jung-lee-hwang:2018:TALLIP,park-tyers:2019:LAW}, 
phrase-structure parsing \citep{choi-park-choi:2012:SP-SEM-MRL,park-hong-cha:2016:PACLIC,kim-park:2022} and 
statistical machine translation (SMT) \citep{park-hong-cha:2016:PACLIC,park-EtAl:2017:Cupral}, etc. where the applications take all the morphemes separated sequence instead of the surface sentence segmented by a blank, as input for language processing. 
{A morpheme-based annotation scheme proposed in \citet{park-tyers:2019:LAW} for POS tagging has been extended to dependency parsing \citep{chen-etal-2022-yet} and named-entity recognition \citep{chen-lim-park-2024-korean} and it attained the most advanced evaluation outcomes.}
Figure \ref{level5-all-morphemes} shows examples of the downstream application process: constituent parsing using the Sejong treebank and machine translation from Korean into English. The sentence often in these applications is converted into the sequence of morphemes to be parsed or translated. They mostly implement granularity level 5 to avoid the problems of data sparsity and unknown words because the number of possible types combined in longer segmentation granularities, such as eojeol, can increase exponentially. Such morpheme-based analysis for the word can be generated by a morphological analysis system. Therefore, most POS tagging systems can produce segmentation granularity Level 5. Separating these morphemes is straightforward from such morphological analysis results. 
{For instance, 
as shown in Figure~\ref{level5-all-morphemes}, in Level 5, 
The phrase \textit{segyejeokin} (`world-class'), which also includes derivational morphemes, is treated as a separated four morphemes sequence \textit{segye-jeok-i-n} instead of one surface segment as input for language processing. 
Specifically, this phrase is assigned four different categories: a NNG (common noun for \textit{segye}), XSN (nominal derivational affix for \textit{jeok}), VCP (copular for \textit{i}) and ETM (adnominal affix for \textit{n}), respectively. These categories consist of the word stem, two derivational morphemes, and an inflectional morpheme, resulting in a new category verb functioning as a modifier in this sentence.}

\begin{figure}[!ht] 
\begin{subfigure}[b]{\textwidth}
\resizebox{\textwidth}{!}{
\synttree 
[SENT [S [NP-SBJ [NP [NP-MOD 
[NNP [프랑스] ] 
[JKG [의]]]
[NP [VNP-MOD 
[NNG [세계] ] 
[XSN [적] ] 
[VCP [이] ] 
[ETM [ㄴ]
]][NP [NP 
[NNG [의상] ]
][NP 
[NNG [디자이너] ] 
]]]][NP-SBJ [NP 
[NNP [엠마누엘] ]
][NP-SBJ 
[NNP [웅가로] ] 
[JKS [가] ] 
]]][VP [NP-AJT [NP [NP [NP 
[NNG [실내] ]
][NP 
[NNG [장식] ] 
[XSN [용] ] 
]][NP 
[NNG [직물] ]
]][NP-AJT 
[NNG [디자이너] ] 
[JKB [로] ] 
]][VP 
[VV [나서] ] 
[EP [었] ] 
[EF [다] ] 
[SF [$\cdot$] ]
]]] ]
}
\caption{Morpheme-based treebank for constituent parsing \citep{choi-park-choi:2012:SP-SEM-MRL}} \label{sjtree-constituent-converted}
\end{subfigure}

\begin{subfigure}{\textwidth}
\footnotesize{
\begin{tabular}{c}
\\
{프랑스의~세계적인~의상~디자이너~엠마누엘~웅가로가~실내~장식용~직물~디자이너로~나섰다. } \\
{} \\
$\Downarrow$ {\footnotesize (separating all morphemes)} \\
{} \\
{프랑스~의~세계~적~이~ㄴ~의상~디자이너~엠마누엘~웅가로~가~실내~장식~용~직물~디자이너~로~나서~었~다~. } \\
{} \\$\Downarrow$ {\footnotesize (machine translation)}\\
{} \\
\textit{The world-class French fashion designer Emanuel Ungaro became an interior textile designer.} \\
\end{tabular}
}
\caption{Example of a machine translation process \citep{park-EtAl:2017:Cupral}} \label{mt-converted}
\end{subfigure}

\caption{Example of downstream application processes} \label{level5-all-morphemes}
\end{figure}

\subsection{Discussion} \label{granularity-discussion}

Figure~\ref{granularity} summarizes an example of each segmentation granularity level. For our explanatory purpose, we use the following sentence in \eqref{ungaro}: \textit{segye-jeok-i-n ... unggaro-ga ... naseo-eoss-da.} (`The world-class ... Ungaro became ...'). The advantage of Level 1 is that it has many linguistics resources to work with. The main weakness of Level 1 is that it requires segmentation including the tokenization process which has been a main problem in language processing in Korean.
While Level 2 has appeared more frequently especially in recent Universal Dependencies (UD)-related resources, and Levels 3 and 4 propose an analysis more linguistically pertinent, they do not mitigate the segmentation problem. Level 5 has the practical merits of a processing aspect. However, the eventual problem for the reunion of segmentation morphemes, for example the generation task in machine translation, still remains, and it has not been discussed much yet.

\begin{figure}[!ht]
\centering
\begin{tabular} {|c|} \hline
\begin{footnotesize}
\begin{tabular} {lllll} 
$_1$ & & $_2$ && $_3$ \\
\textit{segye-jeok-i-n} &...  & \textit{unggaro-ga} & ... & \textit{naseo-eoss-da} \\
NNG+XSN+VCP+ETM & & NNP+JKS && VV+EP+EF+SF \\
`world-class' & & `Ungaro'+\textsc{nom} && `become'+\textsc{past}+\textsc{decl}+\textsc{punct} \\
\end{tabular}\end{footnotesize} \\~\\ 
$\Big\downarrow$ separating  words and symbols \\~\\

\begin{footnotesize}
\begin{tabular} {llll ll }
$_1$ & & $_2$ & & $_3$ & $_4$\\
\textit{segye-jeok-i-n} &... &\textit{unggaro-ga} &... & \textit{naseo-eoss-da} & . \\
NNG+XSN+VCP+ETM & & NNP+JKS && VV+EP+EF & SF \\
\end{tabular} \end{footnotesize} \\~\\ $\Big\downarrow$ separating case markers \\~\\

\begin{footnotesize}
\begin{tabular} {llll lll}
$_1$ & &$_2$ & $_3$ && $_4$ & $_5$\\
\textit{segye-jeok-i-n} &...& \textit{unggaro} & \textit{-ga} &... & \textit{naseo-eoss-da} & . \\
NNG+XSN+VCP+ETM & & NNP & JKS && VV+EP+EF & SF \\
\end{tabular} \end{footnotesize} \\~\\ $\Big\downarrow$ separating verbal endings \\~\\

\begin{footnotesize}
\begin{tabular} {lllll lllll}
$_1$ & $_2$ & & $_3$ & $_4$ && $_5$ & $_6$ &$_7$&$_8$\\
\textit{segye-jeok-i} & \textit{-n} & ...&\textit{unggaro} & \textit{-ga} &... & \textit{naseo} & \textit{-eoss} & \textit{-da} & .  \\
NNG+XSN+VCP & ETM & & NNP & JKS && VV & EP & EF & SF \\
\end{tabular} \end{footnotesize} \\~\\ $\Big\downarrow$ separating all morphemes \\~\\

\begin{footnotesize}
\begin{tabular} {lllll lll llll}
$_1$ & $_2$ & $_3$ & $_4$ & & $_5$ & $_6$ & &$_7$&$_8$& $_9$& $_{10}$ \\
\textit{segye} &\textit{-jeok} &\textit{-i} & \textit{-n} &...& \textit{unggaro} & \textit{-ga} & {...} & \textit{naseo} & \textit{-eoss} & \textit{-da} & .  \\
NNG & XSN & VCP & ETM & & NNP & JKS && VV & EP & EF & SF \\
\\
\end{tabular}\end{footnotesize} 
\\\hline
\end{tabular}
\caption{Five levels of segmentation granularity in Korean and their POS annotation.} \label{granularity}
\end{figure}

\section{Diagnostic Analysis} \label{applications} 
In this section, we present several applications for Korean language processing using proposed segmentation granularity levels to compare them to each other. We use the default options that the system provides for experiments. For experiments, we convert all data sets into each segmentation granularity. We utilize a 90-10 split for the Sejong treebank for the training and evaluation for POS tagging and syntactic parsing. We utilize training and evaluation data sets for Korean-English machine translation provided by \citet{park-hong-cha:2016:PACLIC}.

Firstly, Table~\ref{stat} shows the number of tokens, the ratio of morphologically complex words (MCW) which are made up of two or more morphemes, and the number of immediate non-terminal (NT) nodes (the number of monomorphemic and complex word patterns) in the entire Sejong treebank. Therefore, the immediate NT nodes signify the POS labels, and can be eojeols, morphemes and symbols according to different segmentation granularity.

\begin{table}[!ht]
\centering
\footnotesize{
\begin{tabular} {r  ccccc l} 
\hline
& Level 1 &Level 2 & Level 3 & Level 4 & Level 5 & \\ 
\hline
Token & 370,729 & 436,448 & 577,153 & 752,654 & 829,506 & \\
MCW & 0.7881 & 0.6451 & 0.2939 & 0.0934 & 0 & \\
Immediate NT &4,318&2,378&1,228&526&45 & \\ \hline 
\end{tabular}
}
\caption{
The number of tokens, the ratio of the morphologically complex words (MCW) and the number of immediate non-terminal (NT) in the corpus} \label{stat} 
\end{table}

\subsection{Language processing tasks} \label{lang-tasks}

\paragraph{Word segmentation, morphological analysis and POS tagging}
Word segmentation, morphological analysis and {POS tagging} for Korean requires detection of morpheme boundaries. We use UDPipe, a trainable pipeline \citep{straka-hajic-strakova:2016:LREC} to perform tokenizing and POS tagging tasks. The current experimental setting achieved the {state} of the art word segmentation and POS tagging result for Korean \citep{park-tyers:2019:LAW}. 
Each trained POS tagging model assigns POS labels for its tokens of granularity. For example, a model should generate \textit{segye+jeok+i+n} for morpheme boundary and \texttt{nng+xsn+vcp+etm} as a single POS label in Level 1 for \textit{segyejeokin} (`world-class'), or \texttt{nng} in Level 5 for \textit{segye} (`world').  
{We present the f1 score ($2 \cdot \frac{\text{precision} \cdot \text{recall}}{\text{precision} + \text{recall}}$) for word segmentation evaluation using precision and recall described in \eqref{word-segmentation-metric}, and} the accuracy score for POS tagging evaluation as in \eqref{pos-metric}.

{
\begin{align}
\begin{split}\label{word-segmentation-metric}
\text{precision} &= \frac{\text{\# of relevant word segments} \cap \text{\# of retrieved word segments}}{\text{\# of retrieved word segments}}\\
\text{recall} &   = \frac{\text{\# of relevant word segments} \cap \text{\# of retrieved word segments}}{\text{\# of relevant word segments}}
\end{split}
\end{align}
}

{
\begin{align}
\begin{split}\label{pos-metric}
\text{accuracy} &= \frac{\text{correct \# of POS tagging labels}}{\text{total \# of POS tagging labels}}
\end{split}
\end{align}
}

\paragraph{Syntactic parsing}
Using the granularity Level 5 has been the de facto standard for Korean phrase structure parsing \citep{choi-park-choi:2012:SP-SEM-MRL,park-hong-cha:2016:PACLIC,kim-park:2022}. 
We train and evaluate the Berkeley parser \citep{petrov-EtAl:2006:COLACL,petrov-klein:2007:main} with the different granularity levels. 
The Berkeley parser uses the probabilistic CFG with latent annotations previously proposed in \citet{matsuzaki-miyao-tsujii:2005:ACL}, and performs a series of split and merge cycles of non-terminal nodes to maximize the likelihood of a treebank. It still shows relatively good parsing results. We keep the structure of the Sejong treebank, and terminal nodes and their immediate NTs are varied depending on the granularity level. We provide gold POS labels as input instead of predicting them during parsing to the original word boundary in the word. This allows us to evaluate parsing results with the same number of terminals for all granularity levels. We present the f1 score by precision and recall of bracketing using EVALB \citep{black-etal-1991-procedure} for parsing evaluation {which uses the f1 score based on precision and recall presented in \eqref{parsing-metric}}.

{
\begin{align}
\begin{split}\label{parsing-metric}
\text{precision} &= \frac{\text{\# of relevant constituents} \cap \text{\# of retrieved constituents}}{\text{\# of retrieved constituents}}\\
\text{recall} &   = \frac{\text{\# of relevant constituents} \cap \text{\# of retrieved constituents}}{\text{\# of relevant constituents}}
\end{split}
\end{align}
}

\paragraph{Machine translation}

Using the granularity Level 5 has been the de facto standard for machine translation for Korean \citep{park-hong-cha:2016:PACLIC,park-EtAl:2017:Cupral}. We use the Moses statistical machine translation system \citep{koehn-EtAl:2007:PosterDemo} with the different granularity levels for Korean to train the phrase-based translation model and minimum error rate training \citep{och:2003:ACL} during validation. We present the BLEU {(BiLingual Evaluation Understudy)} score \citep{papineni-etal-2002-bleu} for evaluation.

\subsection{Results and discussion}

\begin{table}[!ht]
\centering
\footnotesize{
\begin{tabular} {r  ccccc l} 
\hline
& Level 1 &Level 2 & Level 3 & Level 4 & Level 5 & \\ 
\hline
Segmentation & \textbf{100.00} & 95.43 & 94.31 & 93.05 & 90.15 & (\textsc{f$_1$}) \\
POS tagging  &83.18&86.28&89.21&92.82& \textbf{96.01}& (\textsc{acc})\\
Syntactic parsing  &76.69&77.50&81.54&\textbf{84.64}&82.23& (\textsc{f$_1$}) \\
Machine translation &5.86&6.87&7.64&7.85& \textbf{7.98}& (\textsc{bleu})\\ \hline
\end{tabular}
}
\caption{
Experiment results on POS tagging, syntactic parsing and machine translation based on different segmentation granularity levels. For comparison purposes, we convert POS tagging results into Level 1 and syntactic parsing results into Level 5. Translation direction is Korean into English.
} \label{results} 
\end{table}

The direct interpretation of task results between the different granularity levels would be difficult because the levels of representation are different (\textit{e.g.}, the number of lexical tokens is different in Table~\ref{stat}). For comparison purposes of experiment results, 
(1) we report segmentation results where Level 1 does not require any segmentation. 
(2) We convert all POS tagging results into Level 1 based on eojeol after training and predicting results for each segmentation granularity level. Therefore, the presented POS tagging accuracy is based on Level 1 eojeols as in previous work on POS tagging \citep{cha-EtAl:1998,hong:2009,na:2015:TALLIP}. 
(3) We convert syntactic parsing results into morpheme-based Level 5 as in previous work on phrase structure parsing \citep{choi-park-choi:2012:SP-SEM-MRL,park-hong-cha:2016:PACLIC,kim-park:2022}. Although the Berkeley parser can predict the POS label during parsing, we provide gold POS labels, {which is correct POS labels from the test dataset} as input for the parsing system to keep original morpheme boundaries. After parsing sentences for each segmentation granularity level, we convert parsing results into Level 5. 
(4) For machine translation, we translate Korean sentences in different segmentation granularity into English where there is no different segmentation granularity. {We {use} \texttt{multi-bleu.perl} provided by Moses \citep{koehn-EtAl:2007:PosterDemo} to evaluate the translation result.\footnote{\url{https://github.com/moses-smt/mosesdecoder}}}

All results based on different segmentation granularity levels are reported in Table~\ref{results}. 
The interpretation of results of segmentation is straightforward where no tokenization is required in Level 1 and more tokenization is required in Level 5. From POS tagging to MT, we provide a gold-segmented sequence to evaluate each task. Results of POS tagging indicate that such morpheme-based analysis outperform other granularity, which conforms to the previous results on morphological analysis and POS tagging for Korean \citep{park-tyers:2019:LAW}. 
As we described, fine-grained granularity by separating all morphemes (Level 5) has been utilized for downstream applications such as machine translation for Korean and it shows the best performance in the BLEU score \citep{papineni-etal-2002-bleu}. 
Whereas phrase structure parsing also uses by separating all morphemes (Level 5) as input for the previous parsing system \citep{choi-park-choi:2012:SP-SEM-MRL,park-hong-cha:2016:PACLIC,kim-park:2022}, granularity by separating only functional morphemes including case markers and verbal endings and keeping other affixes for morphological derivation (Level 4) outperform Level 5. 
The modern statistical parsers have used markovization annotation for non-terminal nodes to elaborate context-free grammar rules for parsing either using the manual heuristics \citep{johnson:1998:CL,klein-manning:2003:ACL} or machine learning techniques \citep{petrov-EtAl:2006:COLACL,petrov-klein:2007:main}. Parsing performance in the statistical parsers is directly related with the size and the quality of CFG rules generated by these annotation schemes of non-terminal nodes. The other explanation for Level 4's parsing performance involves its linguistically soundness of its segmentation of the word, in which its immediate non-terminal nodes represent actual part-of-speech information of the word with its adjoined functional morphemes. Linguistic information of this kind might help to improve the representation of the treebank grammar that is implied by the parsing system.

\section{Conclusion} \label{chapter2-conclusion}

The study addresses word segmentation granularity  for the segmentation in Korean language processing. There have been multiple possible word segmentation granularity levels from a word to morphemes in Korean, and for specific language processing and annotation tasks, several different granularity levels have been proposed and developed. It is that the agglutinative languages including Korean can have a one-to-one mapping between functional morpheme and syntactic category, even though the annotation guidelines for Universal Dependencies typically regard a basic unit of dependency annotation as a syntactic word. We have presented five different levels of segmentation granularity in Korean. We have analyzed and compared these levels of granularity by using Korean language applications as well. Previous work for Korean language processing has not explicitly mentioned which level of segmentation granularity is used, and this makes it difficult to properly compare results between systems. As described, these different levels of segmentation granularity could exist mainly because various Korean treebanks represent their syntactic structure differently. These treebanks also use the different segmentation of words depending on their linguistic and computational requirements. While a certain segmentation granularity may be well suited for some linguistic phenomena or applications, we need to find a correct segmentation granularity level to adapt to our requirements and expectations for Korean language processing.\footnote{This chapter is based on "Word segmentation granularity in Korean" by Jungyeul Park and Mija Kim, published in \textit{Korean Linguistics}, John Benjamins Publishing Company. 20(1):82-112 \citep{park-kim-2024-word}.}

\singlespacing
\chapter{From Morphology to Semantics} \label{pos-to-framenet}
\doublespacing

This chapter develops a single representational thread from morpheme level analysis to sentence level and frame semantic annotation for Korean. The central claim is that the natural segmentation of Korean, while convenient for surface processing, conflates lexical material with function marking in a way that obscures the units required by downstream models. A morpheme based representation, if designed to preserve the original surface segmentation as recoverable metadata, offers a stable interface across tasks that range from local tagging to semantic parsing.

The chapter therefore adopts a CoNLL-U style format that treats eojeols as multiword tokens and exposes their internal morpheme sequence as the primary locus of annotation. This choice is not merely a formatting preference. It makes it possible to define conversion procedures that are reversible, to compare models under controlled tokenization conditions, and to interpret improvements as consequences of representational adequacy rather than of accidental preprocessing differences. The subsequent sections instantiate this strategy in four settings, starting with end to end morphological analysis and POS tagging, then extending the same representation to named entity recognition, dependency parsing, and FrameNet parsing, with explicit conversion scripts and evaluation protocols linking each task back to a shared morpheme level backbone.

\section{POS Tagging}

This section examines the Sejong POS-tagged corpus to propose a new annotation method for end-to-end morphological analysis and POS tagging.
Many upstream applications in Korean language processing are based on a segmentation scheme where all morphemes are separated. For instance, previous work on phrase-structure parsing \citep{choi-park-choi:2012:SP-SEM-MRL,park-hong-cha:2016:PACLIC} and statistical machine translation (SMT) \citep{park-hong-cha:2016:PACLIC,park-EtAl:2017:Cupral} follows this approach to mitigate data sparsity, as longer segmentation granularity can combine words exponentially. To address this, we propose a novel annotation method using morphologically separated words, based on the approach for annotating multiword tokens (MWT) in the CoNLL-U format.\footnote{\url{http://universaldependencies.org/format.html}}

\subsection{CoNLL-U Format for Korean}

We use CoNLL-U style Universal Dependency (UD) annotation for Korean morphology.
We first review the current approaches to annotating Korean in UD and their potential limitations. 
The CoNLL-U format is a revised version of the previous CoNLL-X format, which  contains ten fields from word index to dependency relation to the head. 
This paper concerns only the morphological annotation: word form, lemma, universal POS tag and language-specific POS tag (Sejong POS tag). 
The other fields will be annotated either by an underscore which represents not being available or dummy information so that it is well-formed for input into applications that process the CoNLL-U format such as UDPipe \citep{straka-strakova:2017:CoNLL}.

\subsubsection{Universal POS tags and their mapping} \label{universal-pos}

To facilitate future research and to standardize best practices, \cite{petrov-das-mcdonald:2012:LREC} proposed a tagset of Universal POS categories. 
The current Universal POS tag mapping for Sejong POS tags is based on a handful of POS patterns of eojeols.
However, combinations of words in Korean are very productive and exponential.
Therefore, the number of POS patterns of the word does not converge even though the number of words increases.
For example, the Sejong treebank contains about 450K words and almost 5K POS patterns. 
We also test with the Sejong morphologically analysed corpus which contains 9.2M eojeols. 
The number of POS patterns does not converge and it increases up to over 50K.
The wide range of POS patterns is mainly due to the fine-grained morphological analysis, which shows all possible segmentations divided into lexical and functional morphemes. 
These various POS patterns might indicate useful morpho-syntactic information for Korean. 
To benefit from the detailed annotation scheme in the Sejong treebank, \cite{oh-EtAl:2011} predicted function labels (phrase-level tags) using POS patterns that improve dependency parsing results. 
Table~\ref{universal} shows the summary of the Sejong POS tagset and its detailed mapping to the Universal POS tags.
Note that we convert the \texttt{XR} (non-autonomous lexical root) into the \texttt{NOUN} because they are mostly considered nouns or a part of a noun:\textit{e.g.}, \textit{minju}/XR (`democracy').

\begin{table}[!ht]
\centering
\resizebox{\textwidth}{!}{
\footnotesize{
\begin{tabular}{ rcl }  \hline 
Sejong POS (S) & description & Universal POS (U)\\  
\noalign{\smallskip}\hline\noalign{\smallskip} 
NNG, NNP, NNB, NR, XR & noun related & NOUN\\ 
NNP & proper noun & PROPN \\
NP & pronoun & PRON\\
MAG & adverb & ADV\\ 
MAJ & conjunctive adverb & CONJ \\
MM & determiner & DET\\ 
VV, VX, VCN, VCP & verb related & VERB\\ %
VA & adjective & ADJ\\ 
EP, EF, EC, ETN, ETM & verbal endings & PART\\ 
JKS, JKC, JKG, JKO, JKB, JKV, JKQ, JX, JC & postpositions (case markers) & ADP \\ 
XPN, XSN, XSA, XSV & suffixes & PART \\ 
IC & interjection & INTJ \\
SF, SP, SE, SO, SS & punctuation marks & PUNCT\\ 
SW & special characters & X\\ 
SH, SL & foreign characters & X \\ 
SN & number & NUM\\
NA, NF, NV & unknown words & X \\ \hline 
\end{tabular}
}}
\caption{POS tags in the Sejong corpus and their 1-to-1 mapping to Universal POS tags} \label{universal}
\end{table}

\subsubsection{MWTs in UD} \label{korean-mwe}

Multiword token (MWT) annotation has been accommodated in the CoNLL-U format, in which MWTs are indexed with ranges from the first token in the word to the last token in the word, e.g. 1-2. These have a value in the word form field, but have an underscore in all the remaining fields. 
This multiword token is then followed by a sequence of words (or morphemes). 
For example, a Spanish MWT \textit{vámonos} (`let’s go') from the sentence \textit{vámonos al mar} (`let’s go to the sea') is represented in the CoNLL-U format as in Figure~\ref{vamonos}.\footnote{The example copied from \url{http://universaldependencies.org/format.html}}
\textit{Vámonos} which is the first-person plural present imperative of \textit{ir} (`go') consists of \textit{vamos} and \textit{nos} in MWT-style annotation. 
In this way, we annotate the Korean eojoel as MWTs. 
Figure~\ref{naseossda} shows that \textit{naseossda} (`became') in Korean can also be represented as MWTs, and all morphemes including a verb stem and inflectional-modal suffixes are separated.
\citet{sag-EtAl-2002-multiword} defined the various kinds of MWTs, and \citet{salehi-cook-baldwin:2016:COLING} presented an approach to determine MWT types even with no explicit prior knowledge of MWT patterns in a given language. 
\cite{coltekin:2016} describes a set of heuristics for determining when to annotate individual morphemes as features or separate syntactic words in Turkish. The two main criteria are (1) does the word enter into a labelled syntactic relation with another word in the sentence (e.g. obviating the need for a special relation for derivation); and (2) does the addition of the morpheme entail possible feature class (e.g. two different values for the \texttt{Number} feature in the same syntactic word).

\begin{figure} [!ht]
\centering
\footnotesize{
\subfloat[\textit{vámonos} (`let’s go')\label{vamonos}]{
\begin{tabular} {|lll|}  \hline
1-2&    vámonos  & \_ \\
1   &   vamos   &  ir (`go')\\  
2    &  nos    &   nosotros (`us') \\  
... &&  \\ \hline
\end{tabular}}\\
\subfloat[\textit{naseossda} (`became')\label{naseossda}]{
\begin{tabular} {|lll|}  \hline
... &      &  \\
18-20   &   {naseossda}   &  \_ \\  
18    &  naseo    &   {naseo} (`become') \\  
19   & eoss    &    {eoss} (`\textsc{past}')\\  
20      &da    &     {da}  (`\textsc{ind}') \\ \hline
\end{tabular}} 
}
\caption{Examples of MWTs in UD} \label{mwe-ud}
\end{figure}

\subsection{A new annotation scheme}

This section describes a new annotation scheme for Korean. 
We propose a conversion method for the existing UD-style annotation of the Sejong POS tagged corpus to the new scheme.	

\subsubsection{Conversion scheme}

The conversion is straightforward. 
For one-morpheme words, we convert them into word index, word form, lemma, universal POS tag and Sejong POS tag.
For multiple-morpheme words, we convert them as described in $\S$\ref{korean-mwe}: word index ranges and word form followed by lines of morpheme form, lemma, universal POS tag and Sejong POS tag.
For the lemma of suffixes, we use the Penn Korean treebank-style \citep{han-EtAl:2002} suffix normalisation as described in Table~\ref{suffix-conversion}. 
Figure~\ref{conllu-sejong} shows an example of the proposed CoNLL-U format for the Sejong POS tagged corpus. 
As previously proposed for Korean Universal Dependencies, we separate punctuation marks from the word in order to tokenize them, which is the only difference from the original Sejong corpus which is exclusively based on the {eojeol} (that is, punctuation is attached to the word that precedes it). 
One of the main problems in the Sejong POS tagged corpus is ambiguous annotation of symbols usually tagged with SF, SP, SE, SO, SS, SW. 
For example, the full stop in \textit{naseo}/VV + \textit{eoss}/EP + \textit{da}/EF + ./SF (`became') and the decimal point in 3/SN + ./SF + 14/SN (`3.14') are not distinguished from each other. 
We identify symbols whether they are punctuation marks using heuristic rules, and tokenize them.

\begin{table}
\centering
\footnotesize{
\begin{tabular} {r ccc} \hline 
& word form & lemma & \\
\noalign{\smallskip}\hline\noalign{\smallskip} 
verbal ending & ㄴ & 은 & \\
 & ㄹ지 & 을지 & \\
case marker & 가 & 이 & (`\textsc{nom}') \\
 & 를 & 을 & (`\textsc{acc}') \\
   & 는 & 은 & (`\textsc{aux}') \\\hline 
\end{tabular}
}
\caption{Suffix normalisation examples} \label{suffix-conversion}
\end{table}

\begin{figure} [!ht]
{\scriptsize 
\begin{tabularx} \textwidth {|X|} \hline
{\# sent\_id = BTAA0001-00000012}\\
{\# text = 프랑스의 세계적인 의상 디자이너 엠마누엘 웅가로가 실내 장식용 직물 디자이너로 나섰다.} \\
\begin{tabular} {lll lll l}
1-2&프랑스의&\_&\_&\_&\_  & \textit{peurangseu-ui} (`France-\textsc{gen}')\\
1&프랑스&프랑스&PROPN&NNP&\_ &\textit{peurangseu} (`France')\\
2&의&의&ADP&JKG&\_ & \textit{-ui} (`-\textsc{gen}')\\
3-6&세계적인&\_&\_&\_&\_ & \textit{segye-jeok-i-n} (`world~class-\textsc{rel}')\\
3&세계&세계&NOUN&NNG&\_& \textit{segye} (`world')\\
4&적&적&PART&XSN&\_&\textit{-jeok} (`-\textsc{suf}')\\
5&이&이&VERB&VCP&\_&\textit{-i} (`-\textsc{cop}')\\
6&ㄴ&은&PART&ETM&\_&\textit{-n} (`-\textsc{rel}')\\
7&의상&의상&NOUN&NNG&\_ & \textit{uisang} (`fashion')\\
8&디자이너&디자이너&NOUN&NNG&\_ & \textit{dijaineo} (`designer')\\
9&엠마누엘&엠마누엘&PROPN&NNP&\_ & \textit{emmanuel} (`Emanuel')\\
10-11&웅가로가&\_&\_&\_&\_ &\textit{unggaro-ga} (`Ungaro-\textsc{nom}')\\
10&웅가로&웅가로&PROPN&NNP&\_ & \textit{unggaro} (`Ungaro')\\
11&가&가&ADP&JKS&\_ & \textit{-ga} (`-\textsc{nom}')\\
12&실내&실내&NOUN&NNG&\_ & \textit{silnae} (`interior')\\
13-14&장식용&\_&\_&\_&\_ & \textit{jangsikyong} (`decoration')\\
13&장식&장식&NOUN&NNG&\_& \textit{jangsik} (`decoration')\\
14&용&용&PART&XSN&\_& \textit{-yong} (`usage')\\
15&직물&직물&NOUN&NNG&\_ & \textit{jikmul} (`textile')\\
16-17&디자이너로&\_&\_&\_&\_ & \textit{dijaineo-ro} (`designer-\textsc{ajt}')\\
16&디자이너&디자이너&NOUN&NNG&\_ &\textit{dijaineo} (`designer')\\
17&로&로&ADP&JKB&\_ & \textit{-ro} (`-\textsc{ajt}')\\
18-20&나섰다&\_&\_&\_&SpaceAfter=No & \textit{naseo-eoss-da}  (`become-\textsc{past}-\textsc{ind})\\
18&나서&나서&VERB&VV&\_ & \textit{naseo} (`become')\\
19&었&었&PART&EP&\_ & \textit{-eoss}  (`\textsc{past}')\\
20&다&다&PART&EF&\_ & \textit{-da}  (`-\textsc{ind})\\
21&.&.&PUNCT&SF&\_\\
\end{tabular} \\ \hline

\end{tabularx}}
\caption{The {proposed} CoNLL-U style annotation with multi-word tokens (MWT) for morphological analysis and POS tagging
} 
\label{conllu-sejong}
\label{fig:conllu-proposed}
\end{figure}

\subsubsection{Experiments and results} 
For our experiments, we automatically convert the Sejong POS-tagged corpus into CoNLL-U style annotation with MWE annotation for eojeols. 
We evaluate tokenisation, morphological analysis, and POS tagging results using UDPipe \citep{straka-strakova:2017:CoNLL}. 
We obtain 99.88\% f$_1$ score for segmentation and 94.75\% accuracy for POS tagging for language specific POS tags (Sejong tag sets).
Previously, \citet{na:2015:TALLIP} obtained 97.90\% and 94.57\% for segmentation and POS tagging respectively using the same Sejong corpus.
While we outperform the previous results including \citet{na:2015:TALLIP}, it would not be the fair to make a direct comparison because the previous results used a different size of the Sejong corpus and a different division of  the corpus.\footnote{Previous work often used cross validation or a corpus split without specific corpus-splitting guidelines. This makes it difficult to correctly compare the POS tagging results. 
}
\cite{jung-lee-hwang:2018:TALLIP} showed 97.08\% f$_1$ score for their results (instead of accuracy). 
They are measured by the entire sequence of morphemes because of their seq2seq model. 
Our accuracy is based on a word level measurement.

\subsubsection{Comparison with the current UD annotation}

There are currently two Korean treebanks available in UD v2.2: the Google Korean Universal Dependency Treebank \citep{mcdonald-etal-2013-universal} and the KAIST Korean Universal Dependency Treebank \citep{chun-EtAl:2018:LREC}. 
For the lemma and language-specific POS tag fields, they use  annotation concatenation using the plus sign as shown in Figure~\ref{conllu-sejong-current}. 
We note that Sejong and KAIST tag sets are used as language-specific POS tags, respectively. 
However, while the current CoNLL-U style UD annotation for Korean can simulate and yield POS tagging annotation of the Sejong corpus, they cannot deal with NER or SRL tasks. 
For example, a word like \textit{peurangseuui} (`of France') is segmented and analysed into \textit{peurangseu}/\textsc{proper noun} and \textit{ui}/\textsc{gen}. 
The current UD annotation for Korean makes the lemma \textit{peurangseu+ui}  and makes  NNP+JKG  language-specific POS tag, from which we can produce Sejong style POS tagging annotation: \textit{peurangseu}/NNP+\textit{ui}/JKG. 
While a named entity \textit{peurangseu} (`France') should be recognised independently, UD annotation for Korean does not have any way to identify entities by themselves without case markers.
In addition, as we described in $\S$\ref{universal-pos} the number of POS patterns of the word which is used in the language-specific POS tag field does not converge. 
Recall that the language-specific POS tag is the sequence of concatenated POS tags such as NNP+JKG or NNG+XSN+VCP+ETM. 
The number of these POS patterns is exponential because of the agglutinative nature of words in Korean. 
However, this poses a significant challenge for system implementation if we aim to process the entire Sejong corpus, which contains over 50,000 tags and tag combinations. This increases the search space and may result in memory overload issues.\footnote{This section is based on "A New Annotation Scheme for the Sejong Part-of-speech Tagged Corpus" by Jungyeul Park and Francis Tyers, published in \textit{Proceedings of the 13th Linguistic Annotation Workshop}, pages 195–202 \citep{park-tyers:2019:LAW}.}

\begin{figure} [!ht] 
\centering
\scriptsize{
\begin{tabularx} \textwidth {|X|} \hline
{\# sent\_id = BTAA0001-00000012}\\
{\# text = 프랑스의 세계적인 의상 디자이너 엠마누엘 웅가로가 실내 장식용 직물 디자이너로 나섰다.} \\
\begin{tabular} {lll lll } 
1&   프랑스의&  프랑스+의&  PROPN &  NNP+JKG & \_  \\   
2&   세계적인&  세계+적+이+ㄴ&  NOUN & NNG+XSN+VCP+ETM& \_  \\
3&   의상  &의상  &NOUN&  NNG   & \_  \\
4&   디자이너&  디자이너&  NOUN  &NNG   & \_ \\
5&   엠마누엘&  엠마누엘&  PROPN  & NNP   & \_ \\
6&   웅가로가&  웅가로+가&  PROPN  & NNP+JKS & \_ \\
7&   실내  &실내  &NOUN  &NNG &  \_ \\
8&   장식용&  장식+용&  NOUN  &NNG+XSN & \_ \\
9&   직물  &직물  &NOUN  &NNG   & \_  \\
10&  디자이너로&  디자이너+로&  NOUN&  NNG+JKB & \_ \\
11&  나섰다  &나서+었+다  &VERB&  VV+EP+EF&  SpaceAfter=No \\
12&  .   &.&   PUNCT&   SF  & \\ 
\end{tabular} \\ \hline
\end{tabularx}
}
\caption{The {current} CoNLL-U style UD annotation for Korean. 
It is based on other agglutinative languages such as Finnish and Hungarian in Universal Dependencies. 
It separates punctuation marks for tokenisation.} \label{conllu-sejong-current}
\end{figure}

\section{Named Entity Recognition}

Due to the linguistic features of the name entities (NE) in Korean, conventional \textit{eojeol}-based segmentation, which makes use of whitespaces to separate phrases, does not produce ideal results in named entity recognition (NER) tasks. 
Most language processing systems and corpora developed for Korean {use} \textit{eojeol} delimited by whitespaces in a sentence as the fundamental unit of {text} analysis in Korean. This is partially because the Sejong corpus, the most widely-used corpus for Korean, employs \textit{eojeol} as the basic unit. The rationale of \textit{eojeol}-based processing is simply {treating} the words as they are in the surface form.
It is necessary for a better format and a better annotation scheme for the Korean language to be adapted. In particular, a word should be split into its morphemes.
To capture the language-specific features in Korean and utilize them to boost the performances of NER models, we propose a new morpheme-based scheme for Korean NER corpora that handles NE tags on the morpheme level based on the CoNLL-U format designed for Korean, as in \citet{park-tyers:2019:LAW}. We also present an algorithm that converts the conventional Korean NER corpora into the {morpheme-based} CoNLL-U format, which includes not only NEs but also the morpheme-level information based on the morphological segmentation.
The contributions of this study for Korean NER are as follows:
(1) An algorithm is implemented {in this study} to convert the Korean NER corpora to the proposed morpheme-based CoNLL-U format. We have investigated the best method to represent NE tags in the sentence along with their linguistic properties {and therefore developed the conversion algorithm with sufficient rationales}. 
(2) The proposed Korean NER models {in the paper} are distinct from other systems since our neural system is simultaneously trained for part-of-speech (POS) tagging and NER with a unified, continuous representation. This approach is beneficial {as it captures} complex syntactic information between NE and POS. {This is only possible with a scheme that contains additional linguistic information such as POS.} 
(3) The proposed morpheme-based scheme for NE tags provides a satisfying performance based on the automatically predicted linguistic features. Furthermore, we thoroughly investigate various POS types, {including} the language-specific XPOS and the universal UPOS \citep{petrov-das-mcdonald:2012:LREC}, and determine the type that has the most effect. We {demonstrate and account for the fact that} the proposed BERT-{based} system with linguistic features yields better results over those in which such linguistic features are not used.

\subsection{Representation of NEs for the Korean language} \label{conllu}

The current Universal Dependencies \citep{nivre-EtAl:2016:LREC,nivre-etal-2020-universal} for Korean {uses} the tokenized word-based CoNLL-U format.
Addressing the NER problem using the annotation scheme of the Sejong corpus or other Korean language corpora is difficult because of the agglutinative characteristics of words in Korean.\footnote{Although the tokenized words in Korean as in the Penn Korean treebank \citep{han-EtAl:2002} and the \textit{eojeols} as {the word units} in the Sejong corpus are different, they basically use the punctuation mark tokenized \textit{eojeol} and the surface form \textit{eojeol}, respectively. Therefore, we distinguish them as word (or \textit{eojeol})-based {corpora} from the proposed morpheme-based annotation.}
They {adopt the \textit{eojeol}-based annotation} scheme which cannot handle {sequence-level morpheme boundaries of NEs because of the characteristics of agglutinative languages}. For example, an NE \textit{emmanuel unggaro} (\textsc{person}) without a nominative case marker instead of \textit{emmanuel unggaro-ga} (`Emanuel Ungaro-\textsc{nom}') should be extracted {for the purpose of NER}. 
However, this is not the case in previous work on NER for Korean.

We propose a novel approach for NEs in Korean by using morphologically separated words based on the morpheme-based CoNLL-U format of the Sejong POS tagged corpus proposed in \citet{park-tyers:2019:LAW}, which has successfully obtained better results in POS tagging {compared to using} the word-based Sejong corpus. While \citet{park-tyers:2019:LAW} have proposed the morpheme-based annotation scheme for POS tags and conceived the idea of using their scheme on NER, they have not proposed {any practical} method to adopt the annotation scheme to NER tasks. As a result, existing works have not explored {these} aspects such as how to fit the NE tags to the morpheme-based CoNLL-U scheme and how the NEs are represented in this format. Our proposed format for NER corpora considers the linguistic characteristics of Korean NE, and it also allows the automatic conversion between the word-based format and the morpheme-based format.\footnote{Details of the automatic conversion algorithm between the word-based format and the morpheme-based format {are addressed} in $\S$\ref{experiments}.}
Using the proposed annotation scheme in our work as demonstrated in Figure~\ref{sjtree-ner}, we can directly handle the word boundary problem between content and functional morphemes {while} using any sequence labeling algorithms.
{For example, only \textit{peurangseu} (`France') is annotated as a named entity (\texttt{B-LOC}) instead of \textit{peurangseu-ui} (`France-\textsc{gen}'), and 
\textit{emmanuel unggaro} (`Emanuel Ungaro-\') instead of \textit{emmanuel unggaro-ga} (`Emanuel Ungaro-\textsc{nom}') as \texttt{B-PER} and \texttt{I-PER} in Figure~\ref{sjtree-ner}. 
}

In the proposed annotation scheme, NEs are, therefore, no longer marked on the \textit{eojeol}-based natural segmentation. Instead, each NE annotation corresponds to a specific set of morphemes that belong to the NE. Morphemes that do not belong to the NE are excluded from the set of the NE annotations and thus are not marked as part of the NE. As mentioned, this is achieved by adapting the CoNLL-U format as it provides morpheme-level segmentation of the Korean language. 
While those morphemes that do not belong to the NE are usually postpositions, determiners, and particles, this does not mean that all the postpositions, determiners, and particles are not able to be parts of the NE. An organization's name or the name of an artifact may include postpositions or the name of an artifact may include postpositions or particles in the middle, in which case they will not be excluded. The following NE illustrates the aforementioned case:\label{new-ne-example}

\begin{center}
\footnotesize{
\begin{tabular}{ l lll l  }
1& . . . & \multicolumn{3}{c}{} \\
2&`&\_& SpaceAfter=No&\\
3-4&프로페셔널의&\_&\_& \textit{peuropesyeoneol-ui}\\
3&프로페셔널&B-AFW & \_&\textit{peuropesyeoneol} (`professional')\\
4&의&I-AFW&\_&\textit{-ui} (`-\textsc{gen}')\\
5&원칙&I-AFW&SpaceAfter=No&\textit{wonchik} (`principle')\\
6&'&\_ & SpaceAfter=No&\\
7&은&\_&\_&\textit{-eun} (`-\textsc{top}')\\
8& . . . & \multicolumn{3}{c}{ } \\
\end{tabular}
}
\end{center} \label{professionla}

The NE 프로페셔널의 원칙 \textit{peuropesyeoneol-ui wonchik} (`the principle of professional') is the {title} of a book belonging to \texttt{AFW} (\textsc{artifacts/works}). Inside this NE, the genitive case marker \textit{-ui}, which is a particle, {remains a part of the NE.}
Because these exceptions can also be captured by sequence labeling, such an annotation scheme of Korean NER can provide a more detailed approach to NER tasks in which the NE annotation on \textit{eojeol}-based segments can now be decomposed to morpheme-based segments, and purposeless information can be excluded from NEs to improve the performance of the machine learning process.

\begin{figure} [!ht]
\centering
\footnotesize{
\begin{tabular} {|p{0.9\textwidth}|}    \hline
{\# sent\_id = BTAA0001-00000012}\\
{\# text = 프랑스의 세계적인 의상 디자이너 엠마누엘 웅가로가 실내 장식용 직물 디자이너로 나섰다.} \\
\resizebox{.9\textwidth}{!}
{
\begin{tabular} {lll lll ll}
1-2&프랑스의&\_&\_&\_&\_  &\_  & \\
1&프랑스&프랑스&PROPN&NNP& B-LOC &\_  &\textit{peurangseu} (`France')\\
2&의&의&ADP&JKG&\_ &\_  & \textit{-ui} (`-\textsc{gen}')\\
3-6&세계적인&\_&\_&\_&\_ &\_  & \\
3&세계&세계&NOUN&NNG&\_&\_  & \textit{segye} (`world')\\
4&적&적&PART&XSN&\_&\_  &\textit{-jeok} (`-\textsc{suf}')\\
5&이&이&VERB&VCP&\_&\_  &\textit{-i} (`-\textsc{cop}')\\
6&ㄴ&은&PART&ETM&\_&\_  &\textit{-n} (`-\textsc{rel}')\\
7&의상&의상&NOUN&NNG&\_ &\_  & \textit{uisang} (`fashion')\\
8&디자이너&디자이너&NOUN&NNG&\_ &\_  & \textit{dijaineo} (`designer')\\
9&엠마누엘&엠마누엘&PROPN&NNP& B-PER &\_  & \textit{emmanuel} (`Emanuel')\\
10-11&웅가로가&\_&\_&\_&\_ &\_  &\\
10&웅가로&웅가로&PROPN&NNP& I-PER &\_  & \textit{unggaro} (`Ungaro')\\
11&가&가&ADP&JKS&\_ &\_  & \textit{-ga} (`-\textsc{nom}')\\
12&실내&실내&NOUN&NNG&\_ &\_  & \textit{silnae} (`interior')\\
13-14&장식용&\_&\_&\_&\_ &\_  &\\
13&장식&장식&NOUN&NNG&\_&\_  & \textit{jangsik} (`decoration')\\
14&용&용&PART&XSN&\_&\_  & \textit{-yong} (`usage')\\
15&직물&직물&NOUN&NNG&\_ &\_  & \textit{jikmul} (`textile')\\
16-17&디자이너로&\_&\_&\_&\_ &\_  &\\
16&디자이너&디자이너&NOUN&NNG&\_ &\_  &\textit{dijaineo} (`designer')\\
17&로&로&ADP&JKB&\_ &\_  & \textit{-ro} (`-\textsc{ajt}')\\
18-20&나섰다&\_&\_&\_&\_  &SpaceAfter=No &\\
18&나서&나서&VERB&VV&\_ &\_  & \textit{naseo} (`become')\\
19&었&었&PART&EP&\_ & \_  &\textit{-eoss}  (`\textsc{past}')\\
20&다&다&PART&EF&\_ &\_  & \textit{-da}  (`-\textsc{decl}')\\
21&.&.&PUNCT&SF&\_& \_  &\\
\end{tabular}} \\ \hline
\end{tabular}
}
\caption{CoNLL-U style annotation with multiword tokens for morphological analysis and POS tagging. It can include BIO-based NER annotation where B-LOC is for a beginning word {of} \textsc{location} and I-PER for an inside word of \textsc{person}.} \label{sjtree-ner}
\end{figure}

\subsection{Experiments} \label{experiments}

\subsubsection{Data} \label{experiments-data}
The Korean NER data we introduce in this study is from NAVER, which was originally prepared for a Korean NER competition in 2018.
NAVER’s data includes 90,000 sentences, and each sentence consists of indices, words/phrases, and NER annotation in the BIO-like format from the first column to the {following columns}, respectively. However, sentences in NAVER’s data were segmented based on \textit{eojeol}, which, as mentioned, is not the best way to represent Korean NEs.

We resegment all the sentences into the morpheme-based representation as described in \citet{park-tyers:2019:LAW}, such that the newly segmented sentences follow the CoNLL-U format. An {\texttt{eoj2morph}} script is implemented to map the NER annotation from {NAVER’s \textit{eojeol}-based data to the morpheme-based} data in the CoNLL-U format by pairing each character in NAVER’s data with the corresponding character in the CoNLL-U data, and removing particles and postpositions from the NEs mapped to the CoNLL-U data when necessary. {The following presents an example of such conversions:}

\begin{center}
\footnotesize{
\begin{tabular}{lll c lll l}
\multicolumn{3}{c}{\textsc{naver}} & & \multicolumn{3}{c}{(proposed) \textsc{conllu}} & \\
 &&    & & 1-2& 프랑스의& \_ & \textit{peurangseu-ui}\\
1 &프랑스의     &B-LOC& $\Rightarrow$ & 1& 프랑스& B-LOC &\textit{peurangseu} (`France')\\
&&&& 2& 의 & \_ & \textit{-ui} (`-\textsc{gen}')\\
\end{tabular}
}
\end{center}

In particular, the script includes several heuristics that determine {whether} the morphemes in an \textit{eojeol} belong to the named entity {the \textit{eojeol} refers to}. When both NAVER’s data that have \textit{eojeol}-based segmentation and the corresponding NE annotation, and the data {in the CoNLL-U format} that only {contain} morphemes, UPOS (universal POS labels, \citet{petrov-das-mcdonald:2012:LREC}) and XPOS (Sejong POS labels as language-specific POS labels) features, are provided, the script first {aligns} each \textit{eojeol} from NAVER’s data to its morphemes in the CoNLL-U data, and then determines the morphemes {in} this \textit{eojeol} that {should} carry the NE {tag(s)} of this \textit{eojeol}, if any.
The {criteria} {for} deciding {whether} the morphemes {are supposed to} carry the NE {tags} is that these morphemes should not be adpositions (prepositions and postpositions), punctuation marks, particles, determiners, or verbs. 
However, cases exist in which some \textit{eojeols} that carry {NEs} only contain morphemes of the aforementioned types. In {these cases}, when the script does not find any morpheme that can carry the NE {tag}, the size of the excluding POS set above will be reduced {for} the given \textit{eojeol}. The script will first attempt to find a verb in the \textit{eojeol} to bear the NE annotation, and subsequently, will attempt to find a particle or a determiner. The excluding set keeps shrinking until a morpheme is found to carry the NE annotation of the \textit{eojeol}.
Finally, the script marks the morphemes that are in between two NE {tags} 
{representing the same named entity}
as part of that {named entity} (e.g., a morpheme that is between B-LOC and I-LOC is marked as I-LOC), and assigns all other morphemes an ``O'' notation, where B, I, and O denote beginning, inside and outside, respectively. 
Because the official evaluation set is not publicly available, the {converted data in the CoNLL-U format from NAVER's training set} are then divided into three {subsets}, {namely} the training, holdout, and evaluation sets, with portions of 80\%, 10\%, and 10\%, respectively  {The corpus is randomly split with seed number 42 for the baseline. In addition, during evaluation of neural models, we use {the seed values 41-45, and report the average and their standard deviation.}} \label{seed42}


{We implement a \texttt{syl2morph} script that maps the NER annotation from syllable-based data (e.g., KLUE and MODU) to the data in the morpheme-based CoNLL-U format to further test our proposed scheme. 
While the \texttt{eoj2morph} script described above utilizes UPOS and XPOS tags to decide which morphemes in the \textit{eojeol} should carry the NE tags, they are not used in \texttt{syl2morph} anymore, as the NE tags annotated on the syllable level already excludes the syllables that belong to the functional morphemes. 
Additionally, NEs are tokenized as separate \textit{eojeols} at the first stage before being fed to the POS tagging model proposed by \citet{park-tyers:2019:LAW}. This is because the canonical forms of Korean morphemes do not always have the same surface representations as the syllables do. 
Because the \texttt{syl2morph} script basically follows the similar principles of \texttt{eoj2morph}, except for the two properties (or the lack thereof) mentioned above, we simply present an example of the conversion described above:} 

\begin{center}
\footnotesize{
\begin{tabular}{lll c lll l}
\multicolumn{3}{c}{\textsc{klue}} & & \multicolumn{3}{c}{(proposed) \textsc{conllu}} & \\
1&프&B-LOC& & 1-2& 프랑스의& \_ & \textit{peurangseu-ui}\\
2&랑&I-LOC& $\Rightarrow$ & 1& 프랑스& B-LOC &\textit{peurangseu} (`France')\\
3&스&I-LOC& & 2& 의 & \_ & \textit{-ui} (`-\textsc{gen}')\\
4&의&\_     & & &  &  & \\
\end{tabular}
}
\end{center}

{We further provide \texttt{morph2eoj} and \texttt{morph2syl} methods which allow back-conversion from the proposed morpheme-based format to either NAVER's \textit{eojeol}-based format or the syllable-based format, respectively. The alignment algorithm for back-conversion is simpler and more straightforward given that our proposed format preserves the original \textit{eojeol} segment at the top of the morphemes for each decomposed \textit{eojeol}. As a result, it is not necessary to align morphemes with \textit{eojeols} or syllables. Instead, only \textit{eojeol}-to-\textit{eojeol} or \textit{eojeol}-to-syllable matching is required.
The \texttt{morph2eoj} method assigns NE tags to the whole \textit{eojeol} that contains the morphemes these NE tags belong to, given that in the original \textit{eojeol}-based format, \textit{eojeols} are the minimal units to bear NE tags.} 
{The \texttt{morph2syl} method first locates the \textit{eojeol} that carries the NE tags in the same way as described above for \texttt{morph2eoj}. Based on the fact that NEs are tokenized as separate \textit{eojeols} by \texttt{syl2morph}, the script assigns the NE tags to each of the syllables in the \textit{eojeol}. 
}

{Back-conversions from the converted datasets in the proposed format to their original formats are performed. Both \textit{eojeol}-based and syllable-based datasets turned out to be identical to the original ones after going through conversions and back-conversions using the aforementioned script, which shows the effectiveness of the proposed algorithms. \label{result-identical}
Manual inspection is conducted on parts of the converted data, and no error is found. While the converted dataset may contain some errors that manual inspection fails to discover, back-conversion as a stable inspection method recovers the datasets with no discrepancy. Therefore, we consider our conversion and back-conversion algorithms to be reliable. On the other hand, no further evaluation of the conversion script is conducted, mainly because the algorithms are tailored in a way that unlike machine learning approaches, linguistic features and rules of the Korean language, which are regular and stable, are employed during the conversion process.}

\subsubsection{Experimental setup}
Our feature set for baseline CRFs is described in Figure~\ref{crf-features}. 
We use {\texttt{crf++}\footnote{\url{https://taku910.github.io/crfpp}} {as} the implementation of CRFs, where \texttt{crf++}} automatically generates a set of feature functions using the template.  

\begin{figure}[!th]
\centering
\footnotesize{
\begin{tabular}{|l l  ll |} \hline
    \texttt{\# Unigram}&  && \\
$w_{-2}$ & & $p_{-2}$ &\\
$w_{-1}$ & & $p_{-1}$ &\\
$w_{0}$ & (current word)& $p_{0}$ & (current pos)\\
$w_{1}$ & & $p_{1}$ &\\
$w_{2}$ & & $p_{2}$ &\\
~& && \\
\texttt{\# Bigram}& & &\\
$w_{-2}/w_{-1}$ & &$p_{-2}/p_{-1}$&\\
$w_{-1}/w_{0}$ & &$p_{-1}/p_{0}$&\\
$w_{0}/w_{1}$ & &$p_{0}/p_{1}$&\\
$w_{1}/w_{2}$ & &$p_{1}/p_{2}$&\\\hline
\end{tabular}
}
\centerline{\vbox to 1pc{\hbox to \textwidth{}}}
\caption{CRF feature template example for {\textit{word} and \textit{pos}}}
\label{crf-features}
\end{figure}

We apply the same hyperparameter settings as in \citet{lim-etal-2018-sex} for BiLSTM dimensions, MLP, optimizer including $\beta$, and learning rate to compare our results with those in a similar environment. We set 300 dimensions for the parameters {including} $LSTM$, $Q$, and $MLP$. 
In the training phase, we train the {models} over the entire training dataset as an epoch with a batch size of 16. For each epoch, we evaluate the performance on the development set and save the best-performing model within 100 epochs{, with early stopping applied}. \label{early-stopping}

The standard $F_{1}$ metric ($= 2 \cdot \frac{P \cdot R}{P + R}$) is used to evaluate NER systems, where precision ($P$) and recall ($R$) are as follows: 
\begin{align*}
P = \frac{\textrm{retrieved named entities}~\cap~\textrm{relevant named entities}}{\textrm{retrieved named entities}} \\
R  = \frac{\textrm{retrieved named entities}~\cap~\textrm{relevant named entities}}{\textrm{relevant named entities}}
\end{align*}
We evaluate our NER outputs on the official evaluation metric script provided by the organizer.\footnote{\url{https://github.com/naver/nlp-challenge}}

\subsection{Results} 

We {focus on the} following aspects of our results: (1) whether {the conversion of Korean NE corpora into the proposed morpheme-based CoNLL-U format} is more beneficial {compared to} the previously proposed \textit{eojeol}-based and syllable-based styles,
(2) the effect of multilingual {transformer-based} models, and (3) the impact of the additional POS features on Korean NER. 
{The outputs of the models trained using the proposed morpheme-based data {are converted back to their original format, either \textit{eojeol}-based or syllable-based}, before evaluation.}\label{results-converting-back}
{Subsequently, {all reported results are calculated in their original format for fair comparisons, given the fact that the numbers of tokens in different formats vary for the same sentence.} Nevertheless, it is worth noting that all experiments using the morpheme-based CoNLL-U data are both trained and predicted in this proposed format before conducting back-conversion.} \label{fair-comparison}

\subsubsection{Intrinsic results on various types of models} \label{intrinsic}

We compare the intrinsic results generated by different types of models. By saying ``intrinsic'', it implies that the results in this subsection differ owing to the ways and approaches of learning from the training data. We compare the performances of the baseline CRFs and our proposed neural models, and we also investigate the variations when our models use additional features in the data.

Table~\ref{bio+word+ne} summarizes the evaluation results on the test data based on the proposed machine learning models. 
Comparing the {transformer-based} models with LSTM+\textsc{crf}, {we found that} both mutlingual BERT-based models (\textsc{bert-multi}, \textsc{xlm-roberta}) outperformed LSTM+\textsc{crf}. {The} comparison reveals a clear trend: the word representation method is the most {effective} for the NER {tasks adopting our proposed scheme}. 
For the LSTM+\textsc{crf} model, we initialized its word representation, $v_{i}^{w}$, with a {\texttt{fastText}} word embedding \citep{joulin-EtAl:2016}. 
{However, once we initialized word representation using BERT, we observed performance improvements up to 3.4 points with the identical {neural network} structure {as shown in Table~\ref{bio+word+ne}}. Meanwhile, there are two notable observations in our experiment. The first observation is that the CRF classifier exhibits slightly better performance than the MLP {classifier}, with {an improvement of} 0.23, for \textsc{xlm-roberta}.}\label{R3Q16}
However, the improvement through the use of the CRF classifier is relatively {marginal} compared with the reported results of English NER \citep{ghaddar-langlais:2018:COLING}. 
{Moreover, when comparing the multilingual models (\textsc{bert-multilingual} and \textsc{xlm-roberta}) to the monolingual model (\textsc{klue-roberta}), we found that \textsc{klue-roberta} outperforms both \textsc{bert-multilingual} and \textsc{xlm-roberta}. This is because \textsc{klue-roberta} is trained solely on Korean texts and utilizes better tokenizers for the Korean language \citep{park-etal-2021-klue}.
}

\begin{table} [!th]
\centering
\resizebox{\textwidth}{!}{
\footnotesize{
\begin{tabular}{ c | c | cc | cc | cc} \hline
baseline & LSTM & \multicolumn{2}{c|}{\textsc{bert-multi}} & \multicolumn{2}{c|}{\textsc{xlm-roberta}} & \multicolumn{2}{c}{\textsc{klue-roberta}} \\ 
 \textsc{crf} & \textsc{+crf}  & \textsc{+mlp} & \textsc{+crf} & \textsc{+mlp} & \textsc{+crf} & \textsc{+mlp} & \textsc{+crf}  \\ \hline
{$71.50$} & {$84.76_{\pm0.29}$} & {$87.04_{\pm0.41}$} & {$87.28_{\pm0.37}$} & {$87.93_{\pm0.30}$} & {$88.16_{\pm0.33}$} & {$88.77_{\pm0.39}$} & {$88.84_{\pm0.43}$} \\ \hline
\end{tabular}
}}
\caption{CRF/Neural results using different models {using NAVER's data converted into the proposed format}: {fastText} \citep{joulin-EtAl:2016} for \textsc{lstm+crf} word embeddings.} \label{bio+word+ne}
\end{table}

Table~\ref{bio+word+upos+xpos+ne} details results using various sets of features. We use incremental words, UPOS, and XPOS, for the input sequence $x$ and their unigram and bigram features for CRFs. 
Both LSTM and XLM-RoBERTa achieve their best F$_{1}$ score when only the +UPOS feature is attached.

\begin{table} [!th]
\centering
\footnotesize{
\begin{tabular}{ c | cc | ccc } \hline
\multicolumn{1}{c|}{baseline \textsc{crf}} & \multicolumn{2}{c|}{\textsc{lstm+crf}} & \multicolumn{3}{c}{\textsc{xlm-roberta+crf}} \\ 
\textsc{word}  & \textsc{word} & \textsc{+upos} & \textsc{word} & \textsc{+upos} & \textsc{+xpos} \\ \hline
{$71.50$} & {$84.76_{\pm0.29}$} & {$84.94_{\pm0.34}$} & {$88.16_{\pm0.33}$} & \textbf{{$88.41_{\pm0.27}$}} & {$88.37_{\pm0.22}$}\\ \hline
\end{tabular}
}
\caption{CRF/Neural results using the various sets of features {using NAVER's data converted into the proposed format}.
} \label{bio+word+upos+xpos+ne}
\end{table}

\subsubsection{Extrinsic results on different types of data}\label{result-extrinsic}

This subsection examines the extrinsic results given different types of data, whereas the previous subsection focuses on the differences in various models given only the {morpheme-based} CoNLL-U data. In this subsection, the performances of our models {trained on the datasets either in the proposed format or in their original formats,} and in either the BIO tagging format or the BIOES tagging format, are investigated.

As described in Table \ref{conllu+naver}, both the baseline CRF-based model and the BERT-based model achieve higher F$_{1}$ scores when the proposed CoNLL-U data are used, in contrast with NAVER’s \textit{eojeol}-based data. The testing data are organized {in a way} that the total number of tokens for evaluations remains the same, implying that the  F$_{1}$ scores generated by \texttt{conlleval} are fair for both groups. 
{This is realized by converting the model output of the morpheme-based CoNLL-U data back into NAVER's original \textit{eojeol}-based format such that they have the same number of tokens.} The {morpheme-based} CoNLL-U format outperforms the \textit{eojeol}-based format under the CRFs, whereas the CoNLL-U format still outperformed the \textit{eojeol}-based format by over 1\% {using} the BERT-based model.

Table~\ref{bio+bioes} presents the comparative results on two types of tagging formats where the models do not {use additional POS features}. {Previous studies show that} the BIOES annotation scheme {yields} superior results for several datasets {if the size of datasets is enough to disambiguate more number of labels} \citep{ratinov-roth:2009:CoNLL}.
Both the baseline CRF and neural models achieve higher F$_{1}$ scores than that of the BIOES tagging format when the BIO tagging format is used, which has two more types of labels -- E as endings, and S as single entity elements. {Our result reveals that adding more target labels} to the training data degrades model prediction {(14 labels $\times$ 2 for E and S)}. 
{Accordingly, we observe} that {in} this specific case, {adding} the two additional labels mentioned {increases} the difficulty of predictions.\label{R3Q35}

{Table \ref{modu+klue+klps} compares the results of the proposed morpheme-based CoNLL-U data and the syllable-based data. We use the \textsc{xlm-roberta+crf} model and only use word features for the experiments. Similar to the previous experiments, we back-convert the model output of the morpheme-based data back to the syllable-based format for fair comparisons. 
The results are consistent that for all four datasets, we observe performance improvement ranging from 3.12 to 4.31 points when the CoNLL-U format is adopted.}

The performance difference between syllable-based NER results and morpheme-based NER results is mainly due to the fact that the \textsc{xlm-roberta+crf} model we used employs a subword-based tokenizer with larger units, rather than a syllable-based tokenizer. Therefore, one needs to use a BERT model with a syllable-based tokenizer for fair comparisons, if the goal is to only compare the performance between the morpheme-based format and the syllable-based format. However, this makes it difficult to conduct a fair performance evaluation in our study, because the evaluation we intended is based on BERT models trained in different environments when morpheme-based, \textit{eojeol}-based, or syllable-based data is given. Since subword-based tokenizers are widely employed in a lot of pre-trained models, our proposed format would benefit the syllable-based Korean NER corpora in a way that not only language models using syllable-based tokenizers can take them as the input, but those using BPE tokenizers or other types of subword-based tokenizers can also be trained on these syllable-based corpora once converted.\footnote{This section is based on "Korean named entity recognition based on language-specific features" by Yige Chen, KyungTae Lim and Jungyeul Park, published in \textit{Natural Language Engineering}, Cambridge University Press, 30(3), 625–649 \citep{chen-lim-park-2024-korean}.}

\begin{table} [!th]
\centering
\footnotesize{
\begin{tabular}{ cc | cc } \hline
\multicolumn{2}{c|}{baseline \textsc{crf}} &  \multicolumn{2}{c}{\textsc{xlm-roberta+crf}} \\ 
\textsc{conllu} & \textsc{naver} & \textsc{conllu} & \textsc{naver}  \\ \hline
{$71.50$} & {$49.15$} & {$88.16_{\pm0.33}$} & {$86.72_{\pm0.49}$}\\ \hline
\end{tabular}
}
\caption{CRF/Neural result comparison between {the proposed} CoNLL-U {format} versus NAVER's {\textit{eojeol}-based format using NAVER's data} {where POS features are not applied}.} \label{conllu+naver}
\end{table}

\begin{table} [!th]
\centering
\footnotesize{
\begin{tabular}{ cc | cc } \hline
\multicolumn{2}{c|}{baseline \textsc{crf}} &  \multicolumn{2}{c}{\textsc{xlm-roberta+crf}} \\ 
\textsc{bio} & \textsc{bioes} & \textsc{bio} & \textsc{bioes}  \\ \hline
{$71.50$} & {$70.67$} & {$88.16_{\pm0.33}$} & {$85.70_{\pm0.35}$}\\ \hline
\end{tabular}
}
\caption{CRF/Neural result comparison between BIO versus BIOES annotations {using NAVER's data converted into the proposed format} {where POS features are not applied}.}  \label{bio+bioes}
\end{table}

\begin{table}[!ht]
\centering
\resizebox{\textwidth}{!}{
\footnotesize{
\begin{tabular}{ cc | cc | cc | cc} \hline
\multicolumn{2}{c|}{MODU 19} &  \multicolumn{2}{c|}{MODU 21} & \multicolumn{2}{c|}{KLUE} & \multicolumn{2}{c}{ETRI}\\ 
\textsc{conllu} & \textsc{syllable} & \textsc{conllu} & \textsc{syllable} & \textsc{conllu} & \textsc{syllable} & \textsc{conllu} & \textsc{syllable} \\ \hline
{$88.03_{\pm0.20}$} & {$84.91_{\pm0.35}$} & {$81.72_{\pm0.31}$} & {$78.10_{\pm0.45}$} & {$91.72_{\pm0.29}$} & {$88.15_{\pm0.42}$} & {$97.59_{\pm0.12}$} & {$93.28_{\pm0.37}$}\\ \hline
\end{tabular}
}}
\caption{Result comparison between {the proposed} CoNLL-U {format} and {the syllable-based format using MODU (19 \& 21), KLUE, and ETRI datasets} {where POS features are not applied} (Model:\textsc{xlm-roberta+crf}).}  \label{modu+klue+klps}
\end{table}

\section{Dependency Parsing}

In this study, we propose a morpheme-based scheme for Korean dependency parsing that is developed based on \citet{park-tyers:2019:LAW}, and adopt the proposed scheme to Universal Dependencies \citep{nivre-EtAl:2016:LREC,nivre-etal-2020-universal}, which contains two Korean dependency parsing treebanks, namely the GSD treebank \citep{mcdonald-etal-2013-universal} and the Kaist treebank \citep{choi-EtAl:1994,chun-EtAl:2018:LREC}. While the two Korean treebanks meet the standards of Universal Dependencies and have been studied for dependency parsing tasks extensively, the treebanks are formatted in a way that the natural segmentations of Korean texts are preserved, and even with some morpheme-level information, only the language-specific part-of-speech tags on the morpheme level are included in the treebanks, and {both treebanks do} not have any morpheme-level parsing tags. Different from the traditional scheme based on natural segmentation, this scheme utilizes the inherent morphological and typological features of the Korean language, and the morpheme-level parsing tags can therefore be derived using a set of linguistically motivated rules, which are further used to produce the morpheme-level dependency parsing results and automatic conversions between the morpheme-based format and the traditional format.
The proposed morpheme-based representation is examined using several dependency parsing models, including UDPipe \citep{straka-hajic-strakova:2016:LREC,straka-strakova:2017:CoNLL} and Stanza \citep{qi-etal-2020-stanza}. Compared to the baseline models trained using the two treebanks without modification, our proposed format makes statistically significant improvements in the performances of the parsing models for the Korean language as reported in the error analysis.

\subsection{Representation of {\textsc{morphUD}}} \label{section-represent-morphud-korean}

\begin{figure} [!ht]
\centering
\resizebox{\textwidth}{!}{
\footnotesize{
\begin{dependency}
\begin{deptext}
프랑스 \&의 \& 세계 \&적 \&이 \&ㄴ  \& 의상  \&디자이너 \&엠마누엘  \&웅가로 \&가 \& 실내 \& 장식 \&용 \& 직물  \& 디자이너 \&로  \& 나서 \&었 \&다  \&. \\
 \textit{peurangseu} \& \textit{-ui} \&  \textit{segye} \& \textit{-jeok} \& \textit{-i} \& \textit{-n} \&
 \textit{uisang} \& \textit{dijaineo} \& \textit{emmanuel} \& \textit{unggaro} \& \textit{-ga} \& 
 \textit{silnae} \&  \textit{jangsik} \& \textit{-yong} \& 
 \textit{jikmul} \& \textit{dijaineo} \& \textit{-ro} \& 
 \textit{naseo} \& \textit{-eoss} \& \textit{-da} \& \textit{.}\\ 
 France \& -\textsc{gen} \& world \& -\textsc{suf} \& -\textsc{cop}\& -\textsc{rel} \& 
 fashion \& designer \& Emanuel \& Ungaro \& -\textsc{nom} \& 
 interior \& decoration \& usage \& textile \& designer \& -\textsc{ajt} \& 
 become \& -\textsc{past} \& -\textsc{ind} \& {.}\\
1 \& 2 \& 3 \& 4 \& 5 \& 6 \& 7 \& 8 \& 9 \& 10 \& 11 \& 12 \& 13 \& 14 \& 15 \& 16 \& 17 \& 18 \& 19 \& 20 \& 21\\
\end{deptext}
   \depedge[edge unit distance=1.7ex]{1}{8}{nmod} 
   \depedge[edge below]{2}{1}{case}
   \depedge[edge unit distance=1.5ex]{3}{8}{acl} 
   \depedge[edge below]{4}{3}{aux}
   \depedge[edge below]{5}{3}{aux}
   \depedge[edge below]{6}{3}{aux}
   \depedge{7}{8}{compound}
   \depedge{8}{10}{compound}
   \depedge{9}{10}{compound}
   \depedge[edge unit distance=0.96ex]{10}{18}{nsubj} 
   \depedge[edge below]{11}{10}{case}
   \depedge{12}{13}{compound} 
   \depedge[edge unit distance=2.6ex]{13}{15}{compound} 
   \depedge[edge below]{14}{13}{aux}
   \depedge{15}{16}{compound} 
   \depedge[edge unit distance=1.9ex]{16}{18}{advcl} 
   \depedge[edge below]{17}{16}{case}
   \deproot{18}{root} 
   \depedge[edge below]{19}{18}{aux}
   \depedge[edge below]{20}{18}{aux}
   \depedge[edge unit distance=1.3ex]{21}{18}{punct} 
\end{dependency}
}}
\caption{Example of morpheme-based universal dependencies for Korean: while dependencies in top-side are the original dependencies between words, dependencies in bottom-side are newly added dependencies for between morphemes.}
\label{morphUD-example}
\end{figure}

In this study, we adopt a morpheme-based format that captures the linguistic properties of the Korean language proposed by \citet{park-tyers:2019:LAW}. The natural segmentation of Korean is based on eojeol, which does not necessarily reflect the actual word or morpheme boundaries of the language. For example, an eojeol of Korean may contain both a noun and its postposition, or both a verb and its particles marking tense, aspect, honorifics, etc. While this is typical for Korean as an agglutinative language, it creates difficulties and challenges for NLP tasks regarding the Korean language, including dependency parsing. It is not ideal that the tokens dependency relations are annotated on are sometimes words, and sometimes phrases as an eojeol may consist of more than a word. Furthermore, Korean as an agglutinative language has very regular conjugations, which makes it easy and natural to split those words and phrases into morphemes when analyzing the language since nearly every piece of an eojeol can be identified to be of a certain meaning or function.

The morpheme-based format aims at decomposing the Korean sentences further into morphemes, which means that dependency relations are no longer marked on the eojeol level. Instead, they are marked on morphemes such that within each eojeol that is not monomorphemic, a head of that eojeol will be found and all other morphemes will be attached directly to the head. As a result, the head of a non-monomorphemic eojeol carries the dependency relation this eojeol originally has, and all other morphemes will be attached to it. In order to find the head, we develop a script and apply some heuristics which include that the head of an eojeol is usually a noun, a proper noun, or a verb, and while there is no noun or verb in an eojeol, the script we implemented continues to find other morphemes such as pronouns, adjectives, adverbs, numerals, etc. The script also excludes the use of adpositions, conjunctions, and particles as heads in most cases, unless these are the only part-of-speeches in an eojeol except for punctuations. While there are multiple morphemes that can be heads in an eojeol, the script will decide which one to take based on the part-of-speeches of the morphemes. For instance, when there are multiple nouns, the last one will carry the dependency relation of the eojeol, whereas when there are multiple verbs, the first verb will carry the dependency relation as Korean is a head-final language. Once the head of a non-monomorphemic eojeol is found, the other morphemes will be dependent on the head and be assigned with other dependency relations such as compound, case, auxiliary depending on their UPOS and XPOS.

\subsection{Experiments and results} \label{section-experiments-results}

\subsubsection{Data and systems}
In this study, we deploy two parsers to evaluate our proposed format, namely UDPipe \citep{straka-hajic-strakova:2016:LREC} as a baseline system and Stanza \citep{qi-etal-2020-stanza} as one of the {state-of-the-art} dependency parsers. 
UDPipe is a pipeline designed for processing CoNLL-U formatted files, which performs tokenization using Bi-LSTM, morphological analysis, part-of-speech tagging, lemmatization using MorphoDiTa \citep{strakova-etal-2014-open}, and dependency parsing using slightly modified Parsito \citep{straka-etal-parsing-2015}. Since the whole pipeline needs no language-specific knowledge, which means that it can be trained using corpora in a different scheme, we choose UDPipe as our baseline. 
Stanza is another natural language processing toolkit that includes Dozat's biaffine attention dependency parser \citep{dozat-manning:2017:ICLR}. Dozat's dependency parser uses the minimum spanning tree algorithm that can deal with non-projectivity dependency relations, and more importantly it excelled all of dependency parsers during CoNLL 2017 and 2018 Shared Task \citep{zeman-etal-2017-conll,zeman-etal-2018-conll}
In this study, the two dependency parsing pipelines take both the original word-based form\footnote{While words and eojeols are not the same in Korean based on their definitions, in this study, the terms ``word-based'' and ``eojeol-based'' are interchangeable. } (the current scheme adopted by Universal Dependencies), which we denote as \textsc{wordUD}, and the morpheme-based form, which we denote as \textsc{morphUD}, of the GSD and KAIST treebanks as the input.

We develop the script to convert between the \textsc{wordUD} format and our proposed \textsc{morphUD} format. The script consists of two major components, which are \textsc{wordUD} to \textsc{morphUD} (\texttt{Word2Morph}) and \textsc{morphUD} to \textsc{wordUD} (\texttt{Morph2Word}). The Word2Morph component splits the word tokens in the CoNLL-U treebank of Korean into morphemes using the lemmas already provided, and assigns dependency relations on the resegmented tokens based on the original dependency relations annotated on the word tokens. 
The Morph2Word component, on the other hand, firstly pairs the tokens in the \textsc{wordUD} dataset and the \textsc{morphUD} dataset, and then assigns the dependency relations from morpheme tokens in \textsc{morphUD} to word tokens in \textsc{wordUD}. 
Within both components, a root detector for the word is implemented in order to find the root (or stem) of a word when the word is multimorphemic (i.e., needs to be split into morphemes and attach dependency relations on it correspondingly). { Evaluations of the conversion scripts are not conducted in this study, since the morphemes are inherited from the lemmas in the treebanks, and the part-of-speech tags, roots, and dependency relations are predicted and assigned to the morphemes based on the linguistic features and the grammar of Korean that are regular, as presented in Section \ref{section-represent-morphud-korean}.}

\subsubsection{Results}
We report the labeled attachment score (LAS), which is a standard evaluation metric in dependency parsing, using the evaluation script (2018 version) provided by CoNLL 2018 Shared Task.\footnote{\url{https://universaldependencies.org/conll18/conll18_ud_eval.py}}
Table~\ref{results-main} shows results of udpipe as a baseline system and stanza as one of the state-of-the-art systems. 
All results are reported in the \textsc{wordUD} format. 
That is, all experiments are trained and predicted in the proposed \textsc{morphUD} format, and then the result is converted back to the \textsc{wordUD} format for comparison purposes. 
We train udpipe once because it can produce the same parsing model if we train it on the same machine.
For stanza, we provide average LAS and its standard deviation after five training and evaluation. 
Both systems use the finely crafted 300d embedding file by fastText \citep{bojanowski-EtAl:2017:TACL}: \textsc{wordUD} and \textsc{morphUD} use words and morphemes as their embedding entries, respectively to make sure that their input representation would be correctly matched.
For embeddings, there are 9.6M sentences and 157M words (tokenized)  based on \textsc{wordUD}. The set of documents for embeddings includes all articles published in \textit{The Hankyoreh} during 2016 (1.2M sentences), Sejong morphologically analyzed corpus (3M), and Korean Wikipedia articles (20201101) (5.3M).
As expected, all results of \textsc{morphUD} outperform \textsc{wordUD} in Table~\ref{results-main}.

\begin{table*}
\centering
\footnotesize{
\begin{tabular}{|r | cc | cc|} \hline
& \multicolumn{2}{c|}{\texttt{ko\_gsd}} &  \multicolumn{2}{c|}{\texttt{ko\_kaist}} \\
& \textsc{wordUD} 
& \textsc{+morphUD}
& \textsc{wordUD} 
& \textsc{+morphUD}\\ \hline
udpipe    & 70.90 
& 77.01 
& 77.01  
& 81.80 \\
stanza   & 84.63 ($\pm0.18$) 
& 84.98 ($\pm0.20$) 
& 86.67 ($\pm0.17$) 
& 88.46 ($\pm0.14$)\\ \hline
\end{tabular}
}
\caption{Dependency parsing results: for the comparison purpose all \textsc{morphUD} results are converted back to \textsc{wordUD} after training and predicting with the format of \textsc{morphUD}}
\label{results-main}
\end{table*}

\subsubsection{Error analysis and discussion}

Figure~\ref{confusion-matrix-direction} shows the confusion matrix between \textsc{wordUD} and \textsc{morphUD}, in which the column and the row represent the arc direction of \texttt{gold} and \texttt{system}, respectively. 
\textsc{morphUD} outperforms \textsc{wordUD} in predicting all directions except for right (\texttt{gold}) / left (\texttt{system}).
The system predicts the left arc instead of the correct right arc (212 arc direction errors in \textsc{wordUD} vs. 240 in \textsc{morphUD}). This is because we spuriously added left arcs for functional morphemes in \textsc{morphUD} where the system learned more left arc instances during training. 

\begin{figure}[!ht]
\centering
\subfloat[b][\textsc{wordUD}]{
{
\begin{tabular}{c ccc}
& \textsc{l}& \textsc{r} & \textsc{o} \\
\textsc{l} & 
\cellcolor{blue!0} & 
\cellcolor{blue!100.0} & 
\cellcolor{blue!28.482972136222912}\\
\textsc{r} &  
\cellcolor{blue!65.63467492260062} & 
\cellcolor{blue!0} & 
\cellcolor{blue!54.79876160990712} \\
\textsc{o} &  
\cellcolor{blue!57.27554179566563} & 
\cellcolor{blue!26.006191950464398} & 
\cellcolor{blue!0}\\
\end{tabular}
}\label{confusion-wordud-dir}}
\subfloat[b][\textsc{morphUD}]{
{
\begin{tabular}{c ccc}
& \textsc{l}& \textsc{r} & \textsc{o} \\
\textsc{l} & 
\cellcolor{blue!0} & 
\cellcolor{blue!73.37461300309597} & 
\cellcolor{blue!27.55417956656347} \\
\textsc{r} &  
\cellcolor{blue!74.30340557275542} & 
\cellcolor{blue!0} & 
\cellcolor{blue!52.94117647058824} \\

\textsc{o} &  
\cellcolor{blue!54.79876160990712} & 
\cellcolor{blue!25.696594427244584} & 
\cellcolor{blue!0} \\
\end{tabular}
}\label{confusion-morphud-dir}}

\caption{Confusion matrix for the direction of arcs where the column represents \texttt{gold}, and the row \texttt{system}: \textsc{l}eft, \textsc{r}ight, and \textsc{o} for \textsc{to root}.}
\label{confusion-matrix-direction}
\end{figure}

\begin{figure}[!ht]
\centering
\subfloat[b][\textsc{wordUD}]{
\resizebox{0.45\textwidth}{!}{
\begin{tabular}{c ccccc ccccc c}
  & 0 & 1 & 2 & 3 & 4 & 5 & 6 & 7 & 8 & 9 & 10 \\
0 & 
\cellcolor{red!0} & \cellcolor{red!23.52941176470588} &
\cellcolor{red!5.216426193118757} & \cellcolor{red!0.6659267480577136} &
\cellcolor{red!0.4439511653718091} &  \cellcolor{red!0} & 
\cellcolor{red!0} & \cellcolor{red!0} & 
\cellcolor{red!0} & \cellcolor{red!0} & \cellcolor{red!0} \\

1 & \cellcolor{red!24.30632630410655} & \cellcolor{red!0} & 
\cellcolor{red!77.80244173140954} & \cellcolor{red!19.644839067702552} & 
\cellcolor{red!5.105438401775805} & \cellcolor{red!2.2197558268590454} & 
\cellcolor{red!0.9988901220865706} & \cellcolor{red!0.3329633740288568} & 
\cellcolor{red!0} & \cellcolor{red!0} & \cellcolor{red!0} \\

2 & \cellcolor{red!4.439511653718091} &
\cellcolor{red!100.0} &
\cellcolor{red!0.0} &
\cellcolor{red!65.92674805771365} &
\cellcolor{red!19.20088790233074} &
\cellcolor{red!5.216426193118757} &
\cellcolor{red!2.774694783573807} &
\cellcolor{red!0.22197558268590456} &
\cellcolor{red!0.22197558268590456} &
\cellcolor{red!0.0} \\

3 & \cellcolor{red!0.4439511653718091} &
\cellcolor{red!25.527192008879023} &
\cellcolor{red!96.11542730299666} &
\cellcolor{red!0.0} &
\cellcolor{red!45.39400665926748} &
\cellcolor{red!16.75915649278579} &
\cellcolor{red!3.662597114317425} &
\cellcolor{red!2.108768035516093} &
\cellcolor{red!0.11098779134295228} &
\cellcolor{red!0.22197558268590456} &
\cellcolor{red!0.0} \\

4 & \cellcolor{red!0.3329633740288568} &
\cellcolor{red!8.102108768035517} &
\cellcolor{red!23.307436182019977} &
\cellcolor{red!69.47835738068812} &
\cellcolor{red!0.0} &
\cellcolor{red!26.304106548279687} &
\cellcolor{red!10.321864594894562} &
\cellcolor{red!2.2197558268590454} &
\cellcolor{red!2.2197558268590454} &
\cellcolor{red!0.0} &
\cellcolor{red!0.22197558268590456} \\

5 & \cellcolor{red!0.22197558268590456} &
\cellcolor{red!2.885682574916759} &
\cellcolor{red!6.8812430632630415} &
\cellcolor{red!16.537180910099888} &
\cellcolor{red!34.517203107658155} &
\cellcolor{red!0.0} &
\cellcolor{red!13.540510543840178} &
\cellcolor{red!5.105438401775805} &
\cellcolor{red!2.108768035516093} &
\cellcolor{red!1.2208657047724751} &
\cellcolor{red!0.0} \\

6 & \cellcolor{red!0.11098779134295228} &
\cellcolor{red!0.776914539400666} &
\cellcolor{red!1.6648168701442843} &
\cellcolor{red!4.328523862375139} &
\cellcolor{red!9.100998890122087} &
\cellcolor{red!13.540510543840178} &
\cellcolor{red!0.0} &
\cellcolor{red!4.661487236403995} &
\cellcolor{red!0.8879023307436182} &
\cellcolor{red!0.9988901220865706} &
\cellcolor{red!0.0} \\

7 & \cellcolor{red!0.0} &
\cellcolor{red!0.3329633740288568} &
\cellcolor{red!0.3329633740288568} &
\cellcolor{red!0.9988901220865706} &
\cellcolor{red!1.4428412874583796} &
\cellcolor{red!3.107658157602664} &
\cellcolor{red!5.660377358490567} &
\cellcolor{red!0.0} &
\cellcolor{red!0.9988901220865706} &
\cellcolor{red!0.0} &
\cellcolor{red!0.3329633740288568} \\

8 & \cellcolor{red!0.0} &
\cellcolor{red!0.0} &
\cellcolor{red!0.0} &
\cellcolor{red!0.0} &
\cellcolor{red!0.6659267480577136} &
\cellcolor{red!0.3329633740288568} &
\cellcolor{red!0.5549389567147613} &
\cellcolor{red!1.8867924528301887} &
\cellcolor{red!0.0} &
\cellcolor{red!0.0} &
\cellcolor{red!0.0} \\

9 & \cellcolor{red!0.0} &
\cellcolor{red!0.0} &
\cellcolor{red!0.0} &
\cellcolor{red!0.11098779134295228} &
\cellcolor{red!0.0} &
\cellcolor{red!0.0} &
\cellcolor{red!0.0} &
\cellcolor{red!0.11098779134295228} &
\cellcolor{red!0.11098779134295228} &
\cellcolor{red!0.0} &
\cellcolor{red!0.0} \\

10 & \cellcolor{red!0.0} &
\cellcolor{red!0.0} &
\cellcolor{red!0.0} &
\cellcolor{red!0.0} &
\cellcolor{red!0.22197558268590456} &
\cellcolor{red!0.0} &
\cellcolor{red!0.0} &
\cellcolor{red!0.0} &
\cellcolor{red!0.0} &
\cellcolor{red!0.0} &
\cellcolor{red!0.0} \\
\end{tabular}
}
\label{confusion-wordud}}
\subfloat[b][\textsc{morphUD}]{
\resizebox{0.45\textwidth}{!}{
\begin{tabular}{c ccccc ccccc c}
  & 0 & 1 & 2 & 3 & 4 & 5 & 6 & 7 & 8 & 9 & 10 \\
0 & 
\cellcolor{red!0.0} &
\cellcolor{red!22.97447280799112} &
\cellcolor{red!4.550499445061043} &
\cellcolor{red!1.2208657047724751} &
\cellcolor{red!0.11098779134295228} &
\cellcolor{red!0.0} &
\cellcolor{red!0.0} &
\cellcolor{red!0.0} &
\cellcolor{red!0.0} &
\cellcolor{red!0.0} &
\cellcolor{red!0.0} \\

1 &
\cellcolor{red!23.52941176470588} &
\cellcolor{red!0.0} &
\cellcolor{red!80.6881243063263} &
\cellcolor{red!18.756936736958934} &
\cellcolor{red!5.660377358490567} &
\cellcolor{red!1.2208657047724751} &
\cellcolor{red!0.5549389567147613} &
\cellcolor{red!0.3329633740288568} &
\cellcolor{red!0.0} &
\cellcolor{red!0.0} &
\cellcolor{red!0.0} \\

2 & 
\cellcolor{red!3.9955604883462823} &
\cellcolor{red!94.89456159822419} &
\cellcolor{red!0.0} &
\cellcolor{red!66.48168701442842} &
\cellcolor{red!18.201997780244174} &
\cellcolor{red!5.549389567147614} &
\cellcolor{red!0.8879023307436182} &
\cellcolor{red!0.22197558268590456} &
\cellcolor{red!0.22197558268590456} &
\cellcolor{red!0.0} &
\cellcolor{red!0.0} \\

3 & 
\cellcolor{red!0.8879023307436182} &
\cellcolor{red!24.195338512763595} &
\cellcolor{red!91.00998890122086} &
\cellcolor{red!0.0} &
\cellcolor{red!45.94894561598224} &
\cellcolor{red!15.760266370699222} &
\cellcolor{red!3.4406215316315207} &
\cellcolor{red!1.2208657047724751} &
\cellcolor{red!0.3329633740288568} &
\cellcolor{red!0.11098779134295228} &
\cellcolor{red!0.0} \\

4 & 
\cellcolor{red!0.22197558268590456} &
\cellcolor{red!8.76803551609323} &
\cellcolor{red!27.081021087680355} &
\cellcolor{red!64.37291897891232} &
\cellcolor{red!0.0} &
\cellcolor{red!23.52941176470588} &
\cellcolor{red!10.099889012208656} &
\cellcolor{red!2.5527192008879025} &
\cellcolor{red!1.3318534961154271} &
\cellcolor{red!0.0} &
\cellcolor{red!0.0} \\

5 & 
\cellcolor{red!0.22197558268590456} &
\cellcolor{red!2.108768035516093} &
\cellcolor{red!6.992230854605993} &
\cellcolor{red!19.08990011098779} &
\cellcolor{red!31.853496115427305} &
\cellcolor{red!0.0} &
\cellcolor{red!12.430632630410656} &
\cellcolor{red!3.884572697003329} &
\cellcolor{red!1.7758046614872365} &
\cellcolor{red!0.6659267480577136} &
\cellcolor{red!0.0} \\

6 & 
\cellcolor{red!0.0} &
\cellcolor{red!0.9988901220865706} &
\cellcolor{red!1.9977802441731412} &
\cellcolor{red!3.3296337402885685} &
\cellcolor{red!9.100998890122087} &
\cellcolor{red!13.762486126526083} &
\cellcolor{red!0.0} &
\cellcolor{red!5.105438401775805} &
\cellcolor{red!0.8879023307436182} &
\cellcolor{red!0.5549389567147613} &
\cellcolor{red!0.0}  \\

7 & 
\cellcolor{red!0.0} &
\cellcolor{red!0.0} &
\cellcolor{red!0.6659267480577136} &
\cellcolor{red!0.8879023307436182} &
\cellcolor{red!1.9977802441731412} &
\cellcolor{red!3.7735849056603774} &
\cellcolor{red!4.772475027746948} &
\cellcolor{red!0.0} &
\cellcolor{red!0.8879023307436182} &
\cellcolor{red!0.11098779134295228} &
\cellcolor{red!0.11098779134295228} \\

8 & 
\cellcolor{red!0.0} &
\cellcolor{red!0.0} &
\cellcolor{red!0.0} &
\cellcolor{red!0.22197558268590456} &
\cellcolor{red!0.6659267480577136} &
\cellcolor{red!0.5549389567147613} &
\cellcolor{red!1.4428412874583796} &
\cellcolor{red!1.2208657047724751} &
\cellcolor{red!0.0} &
\cellcolor{red!0.0} &
\cellcolor{red!0.0}  \\

9 & 
\cellcolor{red!0.0} &
\cellcolor{red!0.0} &
\cellcolor{red!0.0} &
\cellcolor{red!0.11098779134295228} &
\cellcolor{red!0.0} &
\cellcolor{red!0.11098779134295228} &
\cellcolor{red!0.0} &
\cellcolor{red!0.0} &
\cellcolor{red!0.4439511653718091} &
\cellcolor{red!0.0} &
\cellcolor{red!0.0} \\

10 & 
\cellcolor{red!0.0} &
\cellcolor{red!0.0} &
\cellcolor{red!0.0} &
\cellcolor{red!0.0} &
\cellcolor{red!0.22197558268590456} &
\cellcolor{red!0.0} &
\cellcolor{red!0.0} &
\cellcolor{red!0.0} &
\cellcolor{red!0.0} &
\cellcolor{red!0.0} &
\cellcolor{red!0.0} \\
\end{tabular}
}\label{confusion-morphud}}

\caption{Confusion matrix for the depth of arcs where the column represents \texttt{gold}, and the row \texttt{system}.}
\label{confusion-matrix}
\end{figure}

\begin{figure}[!ht]
\centering
\resizebox{.8\textwidth}{!}
{
\includegraphics{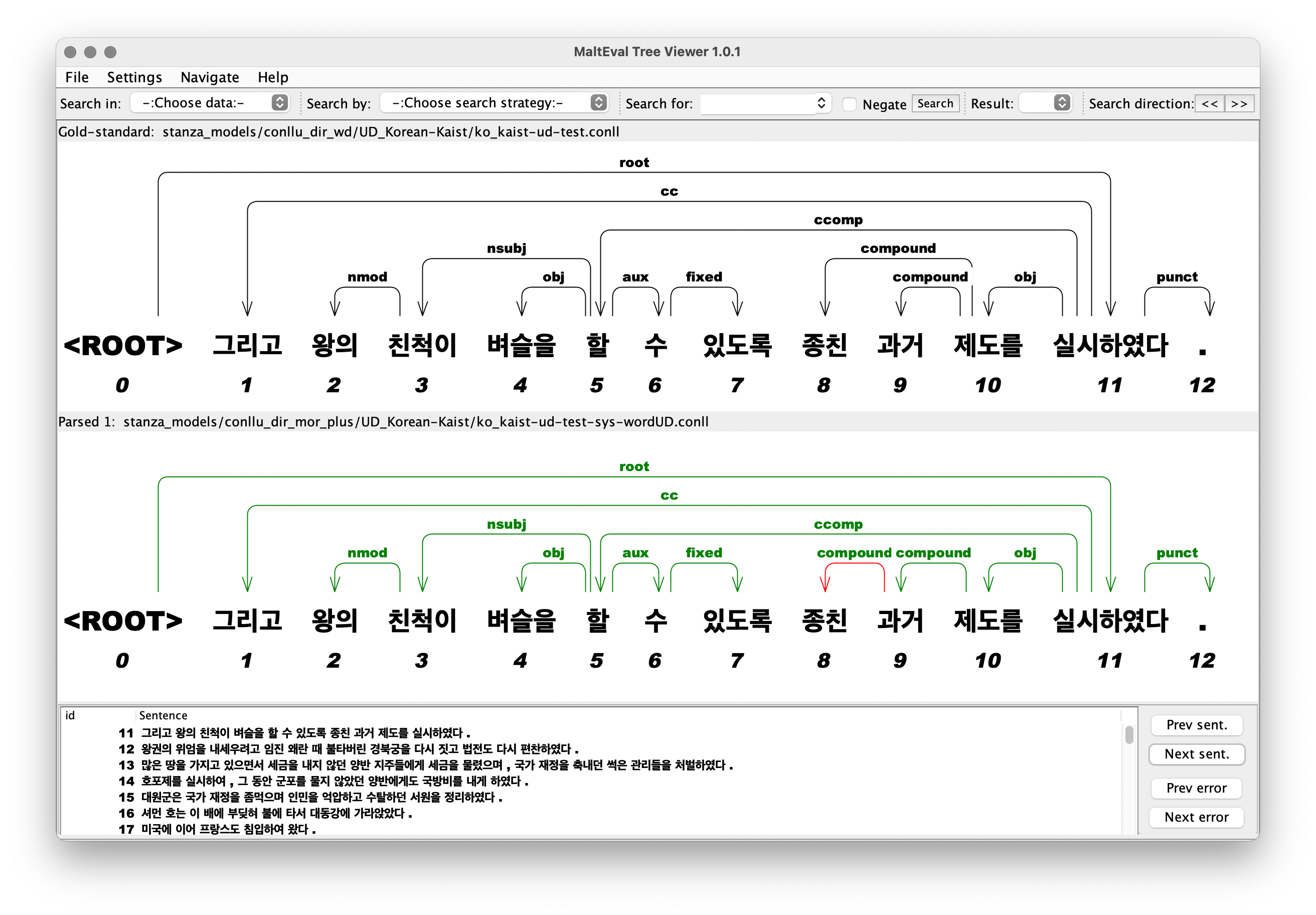} 
}
\caption{Example of the 2/1 error by MaltEval where the \texttt{gold}'s arc depth is 2 and the \texttt{system}'s depth is 1. Note that \textsc{morphUD} results are converted back to \textsc{wordUD}:  \textit{geuligo wang-ui chinjog-i byeoseul-eul hal su issdolog jongchin gwageo jedo-leul silsihayeossda} `And they introduced a clan system to make sure that the king's relatives can obtain the government position'}
\label{error-example}
\end{figure}

\begin{figure}[!ht]
\centering
\footnotesize{
\begin{forest}
where n children=0{tier=word}{}
[NP-AJT [NP [NP [NP [실내 \\\textit{silnae}\\(`interior'),name=silnae]]
[NP [장식용 \\\textit{jangsik-yong}\\(`ornamental'),name=jangsik-yong]]]
[NP [직물 \\\textit{jikmul}\\(`textile'),name=jikmul]]]
[NP [디자이너로 \\\textit{dijaineo-ro}\\(`designer-\textsc{ajt}'),name=dijaineo-ro]]]
\draw[->] (silnae) to[out=south east,in=south] (jangsik-yong);
\draw[->] (jangsik-yong) to[out=south east,in=south] (jikmul);
\draw[->] (jikmul) to[out=south east,in=south] (dijaineo-ro);
\end{forest}
}
\caption{Compound noun with a left-skewed tree for NP modifiers in the Korean treebank} \label{np-modifiers}
\end{figure}

Figure~\ref{confusion-matrix} presents the confusion matrix for the arc depth. 
The most frequent arc depth error is 2 (\texttt{gold}) / 1 (\texttt{system}) (901 arc depth errors in \textsc{wordUD} vs. 855 in \textsc{morphUD}). 
Figure~\ref{error-example} shows an example of parsing errors generated by MaltEval \citep{nilsson-nivre:2008:LREC}. 
The parsing error shows that whereas the gold's arc requires the depth 2, the system predicts the depth 1. 
This is mainly because the analysis of compound nouns for the NP modifier in the Korean treebank prefers a left skewed tree as shown in Figure~\ref{np-modifiers} where some nouns are a verbal noun, and it plays a role as a predicate of the precedent NP modifier. 
This is a quite different from the English treebank where the right skewed tree dominates: $[_{\textsc{np}}$  $[_{\textsc{prps}}$ \textit{its}$]$ $[_{\textsc{n}}$  $[_{\textsc{nn}}$ \textit{Micronite}$]$ $[_{\textsc{n}}$  $[_{\textsc{nn}}$ \textit{cigarette}$]$ $[_{\textsc{nns}}$ \textit{filters}$]]]]]$.
This is a well-known problem when parsing the Korean treebank because it requires the semantics of the noun to distinguish between the right and the left skewed trees.
One possible remedy for this problem was to build a fully lexicalized parsing system \citep{park-EtAl:2013:IWPT}.\footnote{This section is based on "Yet Another Format of Universal Dependencies for Korean" by Yige Chen, Eunkyul Leah Jo, Yundong Yao, KyungTae Lim, Miikka Silfverberg, Francis M. Tyers and Jungyeul Park, published in \textit{Proceedings of the 29th International Conference on Computational Linguistics}, pages 5432–5437 \citep{chen-etal-2022-yet}.}

\section{FrameNet Parsing}

\subsection{Korean FrameNet dataset}\label{section-dataset}
 
The dataset we use was originally developed and published by KAIST \citep{park-EtAl:2014:ISWCPOSTER,kim-etal-2016-korean,hahm-EtAl:2018:LREC}, and it includes multiple sources from which the data are collected.
We choose parts of the whole dataset originating from three sources for the purpose of this study, which are the Korean FrameNet data from Korean PropBank (\textsc{pkfn}), the Japanese FrameNet (\textsc{jkfn}), and the Sejong Dictionary (\textsc{skfn}).
While the Korean FrameNet data from the English PropBank (\textsc{ekfn}) is also available, we noticed that the tokenization scheme does not agree with other datasets, and decided not to adopt it to the current study. Table \ref{framenet-data-lu} {introduces} statistics describing the distribution of the lexical units (LUs). Table \ref{framenet-data-luframe} presents the number of frames per LU, which measures the degree of ambiguity in the lexical units within the three subsets. Table \ref{framenet-data-stats} shows the total number of sentences and instances in each subset, in which identical sentences with different frames count as a single sentence but as separate instances. 

\begin{table}[!th]
\centering
\footnotesize{
\begin{tabular} {r ccc} \hline
\# of LUs&\textsc{pkfn}&\textsc{jkfn}&\textsc{skfn}\\\hline
Noun & 0 & 755 & 0\\
Verb & 644 & 500 & 2,252\\
Adjective & 6 & 155 & 0\\
Others & 0 & 14 & 0\\\hline
Total & 650 & 1,424 & 2,252\\\hline
\end{tabular}
}
\caption{Distributions of the lexical units (LUs) of the targets in 3 Korean FrameNet datasets. An LU is a word with its part-of-speech.
} 
\label{framenet-data-lu}
\end{table}

\begin{table}[!th]
\centering
\footnotesize{
\begin{tabular} {r ccc} \hline
\# of frames per LU&\textsc{pkfn}&\textsc{jkfn}&\textsc{skfn}\\\hline
Noun & 0 & 1.109 & 0\\
Verb & 1.183 & 1.276 & 1.274\\
Adjective & 1.167 & 1.290 & 0\\
Others & 0 & 1.286 & 0\\\hline
Overall & 1.183 & 1.189 & 1.274\\\hline
\end{tabular}
}
\caption{The number of frames per lexical unit for each of the Korean FrameNet datasets} \label{framenet-data-luframe}
\end{table}

\begin{table}[!th]
\centering
\footnotesize{
\begin{tabular} {r ccc} \hline
&\textsc{pkfn}&\textsc{jkfn}&\textsc{skfn}\\\hline
\# of sentences&1,767&1,357&5,703\\
\# of instances&2,350&2,919&5,703\\
\# of frames per sentence&1.330&2.151&1.000\\\hline
\end{tabular}
}
\caption{Numbers of sentences and instances in the 3 Korean FrameNet datasets. 
} \label{framenet-data-stats}
\end{table}

\paragraph{\textsc{pkfn}}
The \textsc{pkfn} data in the Korean FrameNet dataset was sourced from the Korean PropBank.\footnote{\url{https://catalog.ldc.upenn.edu/LDC2006T03}} The dataset contains mainly verbal targets, along with a few adjectival targets. Figure \ref{propbank-ex} illustrates how a single sentence is labeled in the Korean PropBank and the Korean FrameNet dataset respectively, where the FrameNet annotation inherits the predicate-argument relation from PropBank and re-analyzes the sentence using frame semantics.

\begin{figure}[!ht]
\centering
\resizebox{0.95\textwidth}{!}{
\footnotesize{
\begin{tabular}{r cccc}
         &{북한이}& {국무부의 테러지원국 명단에} &{빠지게} &되면 ...\\
         &\textit{{bughan-i}}& \textit{{gugmubuui teleojiwongug myeongdan-e}} &\textit{{ppajige}} &\textit{doemyeon ...}\\
         &{{`North Korea-\textsc{nom}'}}& {`{Department of State's list of state sponsors of terrorism}-\textsc{obl}`} &{`{exclude}'} &{`if/when' ...}\\
PropBank & 
(thing-excluded)$_{\textsc{Arg1}}$ & 
(excluded-from)$_{\textsc{Arg2}}$ & 
(exclude)$_{\textsc{Pred}}$ 
\\
& $\downarrow$ &  $\downarrow$ &  $\downarrow$ &  \\
FrameNet & Theme & Source & 
Removing$_{\textsc{Target}}$
\\
\end{tabular}
}}
\caption{Comparisons between annotations on the same instance in Korean PropBank and the Korean FrameNet dataset. The meaning of the above instance is `if North Korea were excluded from the Department of State's list of state sponsors of terrorism...', which is part of a sentence in the Korean PropBank.} \label{propbank-ex}
\end{figure}

\paragraph{\textsc{jkfn}}
The \textsc{jkfn} data, as presented in \citet{kim-etal-2016-korean}, was projected from the Japanese FrameNet \citep{ohara-etal-2003-japanese}. Given the syntactic similarities between Korean and Japanese, the \textsc{jkfn} data are direct and literal translations from the original word chunks separated by frame data in the Japanese FrameNet, in which way the projected \textsc{jkfn} data preserves the boundaries of the frames \citep{kim-etal-2016-korean} as shown in Figure \ref{jap-frame-ex}. The dataset contains a large number of nominal targets and a considerable number of verbal targets, whereas adjectival targets are also present in the dataset.

\begin{figure}[!ht]
    \centering
    \resizebox{0.95\textwidth}{!}{
    \footnotesize{
    \begin{tabular}{r cccc}
         & \np{小学生が}& \np{青信号で} & \np{横断歩道を}& \np{渡る。}\\
         & \textit{{shōgakusei-ga}} &\textit{aoshingō-de}& \textit{{ōdan hodō-o}} &\textit{{wataru}}\\
         Japanese FrameNet & {Theme}&  & {Path} & {Path\_shape}$_{\textsc{Target}}$\\
& $\downarrow$ & & $\downarrow$ & $\downarrow$ \\
         
Korean FrameNet & {Theme}&  & {Path} &  Path\_shape$_{\textsc{Target}}$\\         
         &초등학생이 &파란 불에 &횡단보도를 &건넌다.\\
         &\textit{{chodeunghagsaeng-i}} &\textit{palan bul-e} &\textit{{hoengdanbodoleul}} &\textit{{geonneonda}}\\
         & {`{elementary school students}-\textsc{nom}'} &{`green light-\textsc{obl}'}& {{`crosswalk-\textsc{acc}'}} &{`{cross}'}\\
    \end{tabular}
    }
    }
    \caption{Comparisons between annotations on the same instance in the Japanese FrameNet dataset and the Korean FrameNet dataset. The meaning of the above instance is `elementary school students cross a crosswalk on the green light'.}
    \label{jap-frame-ex}
\end{figure}

\paragraph{\textsc{skfn}}
The \textsc{skfn} data is based on the example sentences in the Sejong dictionary. The major characteristic that differentiates \textsc{skfn} from the above two subsets is that the example sentences in the dictionary are usually short, and as a result, a sentence in the \textsc{skfn} data carries a single frame only. All frame targets in \textsc{skfn} are verbs with no exception. Figure \ref{sejong-ex} presents an example of the frame-based information in the Sejong dictionary and how its example sentence is annotated in the FrameNet data. Note that 이 (-\textit{i}) denotes any nominative particle in Sejong Dictionary. As a result, \texttt{X} corresponds to the nominative noun phrase \textit{jeo salam-eun} (that person), and \texttt{Y} corresponds to the event \textit{uli il-e} (out affairs), in the example. {The boundaries of frame arguments cannot be inherited from the original source because the Sejong dictionary did not explicitly specify such boundaries. 
Instead, automatic detection and mapping between frame elements and arguments for the frame of the given predicate are conducted.}

\begin{figure}[!ht]
    \centering
    \resizebox{0.95\textwidth}{!}{
\footnotesize{
    \begin{tabular}{r cccc}
        Sejong&\multicolumn{4}{l}{개입하다 (\textit{gaeibhada}, to intervene)}\\
        &\multicolumn{4}{l}{{Frame: X=N0-이 Y=N1-에 V}}\\
        &\multicolumn{4}{l}{{X: AGT (individual|group); Y: LOC (abstract object|event|action)}}\\\\
         &저 사람은 &사사건건 &우리 일에 &개입한다.\\
         &\textit{{jeo salam-eun}} &\textit{{sasageongeon}} &\textit{{uli il-e}} &\textit{{gaeibhanda}}\\
         &{{`that person-\textsc{top}'}} &{{`everything'}} &{{`our affairs-\textsc{obl}'}} &{{`interfere'}}\\
         FrameNet & {Participant}& {Manner} & {Event} & {Participation}$_{\textsc{Target}}$\\
    \end{tabular}
    }}
    \caption{Comparisons between the corresponding information in Sejong Dictionary and the annotation in the Korean FrameNet dataset with regard to a single instance. The meaning of the above instance is ``that person interferes in our affairs constantly and meddles in everything''. }
    \label{sejong-ex}
\end{figure}

\subsection{Morphologically enhanced FrameNet dataset}

We propose a morpheme-based scheme for Korean FrameNet data that leverages the linguistic properties of the Korean language. As an agglutinative language, Korean possesses the feature that the natural segmentation, namely an \textit{eojeol}, can consist of both the lexical morpheme and its postposition, such as a particle that marks tense or case. This poses challenges in Korean FrameNet parsing, as the parser is not able to distinguish the arguments from their functional morphemes given the \textit{eojeol}-based segmentation. In other words, the smallest unit (i.e., \textit{eojeol}) as a single token is a mixture of the lexical part and the functional part, and a sequence labeling model is not able to learn from the \textit{eojeol}-based data and tell what the lexical morphemes are in an \textit{eojeol}. Since the lexical morphemes contribute to the semantic meaning of the \textit{eojeol} on a large scale and determine the lexical units the targets instantiate and the semantic frame they evoke, it is essential to separate them from their postpositions during processing. 

As illustrated in Figure \ref{korean-srl-rev}, the sentence is decomposed into morphemes as the basic unit of tokens. On the other hand, the information on its natural segmentation is preserved by keeping the \textit{eojeols} at the top of the morphemes that are split from the corresponding \textit{eojeol} {following the CoNLL-U format}. The frames are therefore annotated on morphemes instead of \textit{eojeols}, and lexical morphemes and functional morphemes are split into separate tokens.
Although whether a token is lexical or functional is not explicitly annotated, the morphologically enhanced annotation scheme allows the parser to {subconsciously} distinguish functional components from the lexical morphemes that trigger semantic frames. This is in line with the aforementioned agglutinative feature of the Korean language.

We neither exclude the functional morphemes from the annotated targets or arguments, nor do we introduce additional labels to annotate them. This is because (1) functional morphemes are parts of the targets/arguments \citep{park-kim-2023-role} that a parser should identify (therefore must not be labeled as \texttt{O}'s), (2) introducing additional labels would potentially confuse the parser, worsening the model performance, and (3) separation between lexical morphemes and functional morphemes can be performed in postprocessing steps if necessary. {Based on the above, we implement a script that automatically converts existing Korean FrameNet datasets into the morpheme-based format, and back-converts our morpheme-based format into the original format. Conversions in both directions rely on alignments between \textit{eojeols} and morphemes and assignments of tags on the aligned tokens. The morphologically enhanced FrameNet datasets are therefore prepared using the aforementioned script for further experiments. }

\begin{figure}[!ht]
\centering
\scriptsize{
\begin{tabular} {|l lllll|} \hline
\texttt{index} & \texttt{word} & \texttt{lexeme} & \texttt{target} & \texttt{frame} & \texttt{annotation}\\ \hdashline 
... &&&&& \\
16&30&30&\_&\_&B-Time\\
17-19&여년간&\_&\_&\_&\_\\
17&여&여&\_&\_&I-Time\\
18&년&년&\_&\_&I-Time\\
19&간&간&\_&\_&I-Time\\
20-21&오스트리아를&\_&\_&\_&\_\\
20&오스트리아&오스트리아&\_&\_&B-Dependent\_entity\\
21&를&을&\_&\_&{I-Dependent\_entity}\\
22-24&통치한&\_&\_&\_&\_\\
22&통치&통치&{통치하다.v}&{Being\_in\_control}&B-FrameTarget\\
23&하&하&\_&\_&I-FrameTarget\\
24&ㄴ&은&\_&\_&{I-FrameTarget}\\
25-26&좌익이&\_&\_&\_&\_\\
25&좌익&좌익&\_&\_&B-Controlling\_entity\\
26&이&이&\_&\_&{I-Controlling\_entity}\\
... &&&&& \\\hline
\end{tabular}
}
\caption{Example of the morphologically enhanced FrameNet data: \textit{30yeonyeongan oseuteulialeul jibaehan jwaigi...} (`The left wing that ruled Austria for over 30 years...')} \label{korean-srl-rev}
\end{figure}

\begin{table*}[!ht]
\centering
\resizebox{\textwidth}{!}{
\scriptsize{
\begin{tabular} {c r | ccc | ccc} \hline
&&\multicolumn{3}{c}{\texttt{KoELECTRA-Base}} & \multicolumn{3}{|c}{\texttt{KR-BERT-char16424}}\\
\ & & \textsc{pkfn} & \textsc{jkfn} & \textsc{skfn} & \textsc{pkfn} & \textsc{jkfn} & \textsc{skfn}\\\hline
exact & \textit{eojeol} & $0.2523\pm0.0215$ & $0.3968\pm0.0445$ & $0.8091\pm0.0003$ &$0.2964\pm0.0229$ & $0.3493\pm0.0281$ & $0.8041\pm0.0009$\\
& morph & $0.3319\pm0.0807$ & $0.6528\pm0.0135$ & $0.6054\pm0.0056$ & $0.3070\pm0.0868$ & $0.6256\pm0.0127$ & $0.5343\pm0.0042$\\\hline
partial
& \textit{eojeol} & $0.3051\pm0.0224$ & $0.4438\pm0.0444$ & $0.8279\pm0.0003$ & $0.3475\pm0.0226$ & $0.4010\pm0.0267$ & $0.8241\pm0.0008$\\
& morph & $0.4091\pm0.0694$ & $0.7152\pm0.0096$ & $0.7373\pm0.0047$ & $0.4094\pm0.0677$ & $0.6929\pm0.0083$ & $0.6627\pm0.0036$\\
\hline
\end{tabular}
}}
\caption{The cross validation mean plus-minus standard deviation of exact and partial F scores on \textit{eojeol}- and morpheme-based variants of \textsc{pkfn}, \textsc{jkfn} and \textsc{skfn} datasets.} \label{framenet-experiment-results}
\end{table*}

\subsection{Experiments and results}

We perform semantic frame parsing on the proposed datasets {and the original datasets respectively}. {Specifically,} we focus only on the argument extraction {task with} the assumption that the frame target and the frame itself have already been given to the {parsers} as inputs. This allows us to approach {the} problem as a {sequence labeling} task, where the tokens are the lexical units and the classes are frame elements. {We remap the frame-specific elements into general arguments given that the Korean FrameNet datasets contain more than 2,000 unique frame elements which are hard to be classified with the limited instances. }
Hence, our classification is over five classes: \texttt{O}, \texttt{B-FrameTarget}, \texttt{I-FrameTarget}, \texttt{B-Argument}, and \texttt{I-Argument}, following the BIO tagging scheme.  Our {parsers are} based on the pre-trained  {\texttt{KoELECTRA-Base-v3} discriminator} model\footnote{\url{https://github.com/monologg/KoELECTRA}} {and the \texttt{KR-BERT-char16424} model\footnote{\url{https://github.com/snunlp/KR-BERT}}}, and {are} fine-tuned for the argument {detection} task using our proposed datasets. The models have their own tokenizers whereas they process the already segmented \textit{eojeols} and morphemes from our proposed datasets. 
The hyperparameter settings {are as follows}: 

\begin{center}
\footnotesize{
\begin{tabular} {r c} 
\hline
Epochs & 3 \\
Learning Rate & 5e-5 \\
Batch Size (train) & 128 \\
Batch Size (eval) & 256\\
Evaluation Strategy & epoch \\
\hline
\end{tabular}
}
\end{center}
For evaluation of the {parsers'} performance, we use measurements {as suggested} in SemEval'13 \citep{jurgens-klapaftis-2013-semeval}. Specifically, we use the exact $F_1$ score to choose our best epoch out of three training epochs.
The morpheme-based {outputs} {are converted back} into the \textit{eojeol}-based {format for fair comparisons of the results}. 

The exact and partial $F_1$ scores of parsers trained on \textit{eojeol}- and morpheme-based 
data using 2-fold cross-validation is summarized in Table \ref{framenet-experiment-results}. 
It is observed that the parsers trained on the morpheme-based datasets {substantially} outperform those trained on the \textit{eojeol}-based alternatives with regard to the \textsc{pkfn} and \textsc{jkfn} data. 
The disagreement from \textsc{skfn} may be {owning to the fact that the argument boundaries are not direct inheritances from its source data, as discussed in Section \ref{section-dataset}. 
This potentially causes some discrepancies within the \textsc{skfn} dataset, and the discrepancies further hinder the morpheme-based parsers from obtaining satisfactory performance since morphemes as smaller units than \textit{eojeols} are more sensitive to the boundaries. Overall, we find our proposed scheme an effective approach to representing Korean FrameNet data {as previous work suggested in other Korean language processing tasks}.}
As future work, resolving the discrepancies within \textsc{skfn} will necessitate a comprehensive strategy. Primarily, it is essential to conduct a more thorough investigation into the underlying causes of these inconsistencies, as detailed in Section \ref{section-dataset}, with the goal of fortifying the dataset's reliability. This may involve the refinement of argument boundary derivation processes or the exploration of alternative methods to ensure greater precision and consistency in annotations.\footnote{This section is based on "Towards Standardized Annotation and Parsing for Korean FrameNet" by Yige Chen, Jae Ihn, KyungTae Lim,  and Jungyeul Park, published in \textit{Proceedings of the 2024 Joint International Conference on Computational Linguistics, Language Resources and Evaluation (LREC-COLING 2024)}, pages 16653–16658 \citep{chen-etal-2024-towards-standardized}.}

\section{Summary of contributions}

This chapter develops a single representational line from morpheme level analysis to sentence level and frame semantic annotation for Korean, arguing that eojeol based segmentation conflates lexical material with function marking in ways that hinder downstream modeling. The chapter’s main contribution is to articulate a morpheme based representation that preserves the original surface segmentation as recoverable metadata, making tokenization an explicit, controllable variable rather than an implicit preprocessing artifact. By adopting a CoNLL-U style format in which eojeols are treated as multi word tokens and their internal morpheme sequences form the primary annotation layer, the chapter establishes a stable interface that supports reversible conversion procedures, controlled comparisons across tasks, and interpretable error analysis grounded in representational adequacy.

Within this unified format, the chapter contributes an end to end morphology and part of speech tagging scheme for Korean built from Sejong resources, including a principled mapping from Sejong part of speech tags to universal part of speech categories and a concrete conversion strategy that expands multi morpheme eojeols into CoNLL-U multi word token structures. It further specifies suffix normalization rules and practical tokenization heuristics for symbol disambiguation, enabling large scale processing in a well formed CoNLL-U pipeline environment. A central empirical and methodological point is that the morpheme based scheme avoids the exponential growth of eojeol level part of speech pattern types that arises under concatenated tag representations, thereby improving tractability while retaining linguistically informative structure.

Building on the same backbone, the chapter contributes a morpheme based named entity recognition framework for Korean that redefines named entity boundaries at the morpheme level rather than at the eojeol level. It introduces conversion and back conversion procedures that map between eojeol based and syllable based named entity corpora and the proposed morpheme based CoNLL-U representation, with rule guided decisions about when functional material should be excluded from named entities and when it must be retained as part of the entity span. The modeling results reported in the chapter support the representational claim: training and prediction in the morpheme based format, followed by conversion back to the original evaluation format, yields consistent gains across datasets and model families, while also enabling a controlled study of the contribution of part of speech features and tagging conventions.

The chapter extends the same representational commitment to dependency parsing by defining morphUD, a morpheme level dependency representation derived from existing Korean Universal Dependencies treebanks. It contributes a linguistically motivated head selection procedure inside non monomorphemic eojeols, a set of rules for attaching remaining morphemes to that head, and bidirectional conversion scripts that allow training in morphUD while reporting results in the original word based representation for comparability. The reported parsing experiments with a baseline pipeline and a strong neural parser show that morphUD training improves labeled attachment scores on both Korean treebanks, and the accompanying error analysis clarifies which structural phenomena are affected, including systematic changes in arc direction distributions and reductions in common depth related errors.

Finally, the chapter contributes a morpheme based reformulation of Korean FrameNet parsing, motivated by the observation that eojeol level tokens obscure the lexical material that determines frame targets and frame element boundaries. It characterizes the internal heterogeneity of the Korean FrameNet resource by distinguishing subsets derived from Korean PropBank, projected Japanese FrameNet data, and Sejong dictionary examples, and it motivates morpheme based annotation as a way to make lexical and functional components separable for sequence labeling models. By proposing a morphologically enhanced FrameNet dataset design that follows the same CoNLL-U multi word token principle, the chapter completes the intended representational thread: one morpheme centered interface supporting morphology, named entities, syntactic dependency structure, and frame semantic annotation under conversion stable, evaluation compatible workflows.

\part{How to \textit{Evaluate} Results of NLP}
\singlespacing
\chapter{Evaluation by Alignment} \label{evaluation-by-alignment}
\doublespacing

\section{Introduction}

Accurate evaluation is a cornerstone in the field of Natural Language Processing (NLP), playing an important role in determining how effectively a system meets its intended goals. Traditionally, evaluation in NLP involves comparing system outputs against gold-standard answers to measure performance. This method has been particularly useful in assessing component-based systems, where individual systems are evaluated separately to prevent errors from propagating through subsequent preprocessing steps.
However, this traditional approach may not fully capture the complexities of real-world scenarios. The rise of end-to-end systems in NLP, which manage entire tasks from start to finish without relying on separate modular components, has highlighted the need for more flexible and robust evaluation methodologies. While these systems offer streamlined solutions, they also introduce new challenges for evaluation, particularly when working with real-world data and its outcomes.
One major challenge is that traditional evaluation metrics often assume predefined, consistent sentence boundaries between the gold standard and the system outputs. In practice, this assumption often does not hold, leading to significant evaluation inaccuracies and potential failures. Differences in sentence boundaries, especially when processing raw text, can result in mismatches that traditional evaluation methods are ill-equipped to handle.

In response to these issues, we introduce a new evaluation algorithm, which we refer to as the jp-algorithm (jointly preprocessed evaluation algorithm).
This algorithm employs an alignment-based approach to improve evaluation accuracy by directly addressing the mismatches that occur during preprocessing, particularly in tasks like tokenization and sentence boundary detection (SBD). Drawing inspiration from alignment techniques commonly used in machine translation (MT), our algorithm simplifies the evaluation process for mismatches that occur during preprocessing tasks, making it both more efficient and accurate for subsequent evaluation of language processing tasks.
This allows us to reevaluate key edge cases, refine precision, and improve recall for F measures in the final evaluation results. 
To emphasize the significance of reevaluating language processing tasks, we begin by addressing certain overlooked aspects of language processing evaluation. We propose that the mismatches between sentence and word segmentation be reconsidered to improve the accuracy of evaluation for NLP systems, particularly those that rely on precise sentence and word boundaries for downstream tasks like constituency parsing and grammatical error correction.

In constituent parsing, for example, the \texttt{EVALB} implementation has long been the standard tool for evaluating parser performance, using PARSEVAL measures \citep{black-etal-1991-procedure} to compare predicted parse trees against human-labeled reference trees. A constituent in a hypothesis (system) parse is considered correct if it matches a constituent in the reference (gold) parse with the same non-terminal symbol and span. Despite \texttt{EVALB}’s success in evaluating constituency parsing results, it faces unresolved issues, such as the requirement for consistent tokenization results and equal-length gold and system parse trees.
Similarly, in GEC, evaluation metrics such as \texttt{errant}‘s F$.5$ scoring \citep{bryant-etal-2017-automatic,bryant:2019} often share the same limitation: they require consistent sentence boundaries between the gold standard and system outputs. This requirement is problematic when dealing with raw text input, where the sentence boundaries in learners’ writing may not align with the predefined corrections.

The contributions of our work are threefold. First, we introduce an alignment-based method that enhances the robustness and accuracy of end-to-end evaluation. Second, we extend the applicability of this method to several NLP tasks, including preprocessing, constituent parsing, and GEC. Third, we emphasize the real-world relevance of our approach by addressing the complexities of writing structures, regardless of their preprocessing—such as SBD and tokenization—ensuring reliable evaluations across diverse text inputs.
In conclusion, the \textit{jointly preprocessed} algorithm represents a significant advancement in NLP evaluation methodologies. By addressing the limitations of traditional methods, particularly in the context of end-to-end systems, our approach offers a more accurate and reliable way to assess the performance of NLP systems. Whether applied to constituency parsing, GEC, or other NLP tasks where predefined sentence boundaries and consistent tokenization results are required, the algorithm ensures that evaluations reflect the true capabilities of the system, taking into account the complexities of real-world data.

\section{Jointly Preprocessed Evaluation Algorithm}

The Jointly Preprocessed (JP) alignment-based evaluation algorithm introduces a pattern-based approach for sentence and word alignment in a monolingual context. Algorithm~\ref{alignment-algorithm} outlines the general procedure of the jp-algorithm, facilitating both sentence and word alignment processes. The input and output for the jp-alignment algorithm are defined as follows:

\begin{description}
\item[Input] $\mathcal{L}$ and $\mathcal{R}$ are unaligned texts.
\item[Output] $\mathcal{L}^{\prime}$ and $\mathcal{R}^{\prime}$ are aligned texts, with equal lengths of either sentences or words.
\end{description}

\begin{algorithm}[!ht]
\caption{Pseudo-code for JP-alignment}\label{alignment-algorithm}
{
\begin{algorithmic}[1]
\STATE{\textbf{function} \textsc{Alignment} ($\mathcal{L}$, $\mathcal{R}$):}
\begin{ALC@g}
\WHILE{$\mathcal{L}$ and $\mathcal{R}$}
\IF{ \textsc{Matched cases$_{(i,j)}$}} 
\STATE{$\mathcal{L^{\prime}}$, $\mathcal{R^{\prime}}$ $\gets$ $\mathcal{L^{\prime}}+{L}_i, \mathcal{R^{\prime}}+{R}_j$ }
\ELSE
\WHILE{$\neg$(\textsc{Matched cases$_{(i+1,j+1)}$}}
\IF{$\textsc{len}({L}_{i}) > \textsc{len}({R}_{j})$}
\STATE{${L}^{\prime}$ $\gets$ ${L}^{\prime}+{L}_{i}$} 
\STATE{$i \gets i+1 $}
\ELSE
\STATE{${R}^{\prime}$ $\gets$ ${R}^{\prime}+{R}_{j}$}
\STATE{$j \gets j+1 $}
\ENDIF

\ENDWHILE
\STATE{
$\mathcal{L^{\prime}}$, $\mathcal{R^{\prime}}$ 
$\gets$ 
$\mathcal{L^{\prime}}+L^{\prime}, \mathcal{R^{\prime}}+R^{\prime}$
}
\ENDIF
\ENDWHILE
\RETURN {$\mathcal{L^{\prime}}$, $\mathcal{R^{\prime}}$}
\end{ALC@g}
\end{algorithmic} 
}
\end{algorithm} 

$\mathcal{L}$ and $\mathcal{R}$ represent either a list of sentences or words, depending on whether sentence alignment or word alignment is being performed. $\mathcal{L}^{\prime}$ and $\mathcal{R}^{\prime}$ denote the aligned lists of sentences or words, where \textsc{len}($\mathcal{L}^{\prime}$) and \textsc{len}($\mathcal{R}^{\prime}$) are equal. \textsc{Matched cases$_{(i,j)}$} may also vary depending on whether sentence alignment or word alignment is being applied.

\subsection{Sentence alignment}

For the \textsc{SentenceAlignment} algorithm, we define the input and output as follows:
\begin{description}
\item[Input:] $\mathcal{L}$ and $\mathcal{R}$, representing the list of sentences.
\item[Output:] $\mathcal{L}^{\prime}$ and $\mathcal{R}^{\prime}$, representing the aligned list of sentences, where both lists have equal lengths.
\end{description}

We designate two sets of inputs, $\mathcal{L}$ (representing the gold data) and $\mathcal{R}$ (representing the system outputs), where $\mathcal{L}$ is expressed as $[L_{1}, L_{2}, L_{3}, \dots, L_{m}]$ and $\mathcal{R}$ as $[R_{1}, R_{2}, R_{3}, \dots, R_{n}]$. After aligning the sentences, we generate aligned document outputs such as $\mathcal{L}^{\prime} = [[L_{1}], [L_{2}, L_{3}], \dots, [L_{m}]]$ and $\mathcal{R}^{\prime} = [[R_{1}, R_{2}], [R_{3}], \dots, [R_{n}]]$, where $\textsc{len}(\mathcal{L}^{\prime})$ and $\textsc{len}(\mathcal{R}^{\prime})$ are equal.
We define the following two cases for \textsc{Matched cases$_{(i,j)}$} in sentence alignment:
\begin{align}
\label{sa-case1} {L}_{i({\not\sqcup})} = {R}_{j({\not\sqcup})} \\
\label{sa-case2} ({L}_{i({\not\sqcup})} \simeq {R}_{j({\not\sqcup})}) ~\land 
({L}_{i+1({\not\sqcup})} = {R}_{j+1({\not\sqcup})} \lor {L}_{i+1({\not\sqcup})} \simeq {R}_{j+1({\not\sqcup})})
\end{align}

We define that $L_{i}$ and $R_{j}$ can be aligned if their respective sequences, denoted as $L_{i\not\sqcup}$ and $R_{j\not\sqcup}$, match when all spaces are removed, as shown in \eqref{sa-case1}. This comparison accounts for tokenization results in the monolingual context, ensuring sentence alignment considers both tokenized systems and gold sentences.
However, simply concatenating words by removing spaces is insufficient for comparing tokenized systems and gold sentences, as tokenization may introduce grammatical morphemes that are absent in the gold sentence, and vice versa. Therefore, we also allow $L_{i}$ and $R_{j}$ to be aligned when their respective sequences, $L_{i\not\sqcup}$ and $R_{j\not\sqcup}$, exhibit high character similarity ($>\alpha$), provided that $L_{i+1}$ and $R_{j+1}$ also match or are sufficiently similar. We define the similarity measure $\alpha$ as follows:

\begin{equation}
\alpha = 2 \times \frac{|\textsc{len}(L_{i\not\sqcup}) - \textsc{len}(R_{i\not\sqcup})|}{\textsc{len}(L_{i\not\sqcup}) + \textsc{len}(R_{i\not\sqcup})}    
\end{equation}
\noindent where \textsc{len} denotes the function returning the number of characters.

These two conditions form the sentence matching criteria. If $L_i$ and $R_j$ cannot be aligned, we concatenate either {$L_i$, $L_{i+1}$} or {$R_j$, $R_{j+1}$} based on their respective lengths. If the length of $L_{i:m}$ exceeds that of $R_{j:n}$, we concatenate $L_i$ and $L_{i+1}$, and vice versa if $R_{j:n}$ is longer. This process continues until $L_{i+1}$ and $R_{j+1}$ meet the sentence matching criteria.

\subsection{Word alignment}

The \textsc{WordAlignment} algorithm follows a logic similar to sentence alignment and is defined as follows for input and output:

\begin{description}
\item[Input] $l$ and $r$ represent lists of words.
\item[Output] $l'$ and $r'$ are the aligned lists of words with equal lengths.
\end{description}

Upon aligning the words, we generate aligned word outputs, such as ${l}^{\prime} = [[l_{1}], [l_{2}, l_{3}], \dots, [l_{m}]]$ and ${r}^{\prime} = [[r_{1}, r_{2}], [r_{3}], \dots, [r_{n}]]$, where $\textsc{len}({l}^{\prime}) == \textsc{len}({r}^{\prime})$. This involves accumulating words in $l^{\prime}$ and $r^{\prime}$ when pairs of $l_i$ and $r_j$ do not match, often due to tokenization mismatches. We assume interchangeability between the notations used for sentence alignment ($\mathcal{L}$) and word alignment ($l_i$).
We define the following two cases for matched \textsc{cases$_{(i,j)}$} in word alignment:

\begin{align}
\label{wa-case1} l_{i} = r_{j} \\
\label{wa-case2} (l_{i} \neq r_{j}) \land (l_{i+1} = r_{j+1}) 
\end{align}

When deciding whether to accumulate the token from $l_{i+1}$ or $r_{j+1}$ in the case of a word mismatch, we base our decision on the following condition, rather than a straightforward comparison between the lengths of the current tokens $l_{i}$ and $r_{j}$:
$(\textsc{len}(l) - \textsc{len}(l_{0..i}))  > (\textsc{len}(r) - \textsc{len}(r_{0..j}))$. 
This condition compares the remaining lengths of the two lists starting from their respective current positions. Specifically, we calculate the difference between the total length of the list $l$ and the cumulative length of tokens up to $l_i$ for the left side, and similarly for the right side with $r_j$. If the remaining length of $l$ is greater than that of $r$, the algorithm decides to accumulate the next token from $l_{i+1}$; otherwise, it accumulates from $r_{j+1}$.

For word-aligned sentence pairs, we re-index the words after alignment. For example, given two sentences $[S_{i}, S_{j}]$ where $S_{i} = [w_{1}, w_{2}, \dots, w_{m}]$ and $S_{j} = [w_{1}, w_{2}, \dots, w_{n}]$, when they are concatenated, the word indices become $[w_{1}, w_{1}, \dots, w_{m}, w_{m+1}, w_{m+2}, \dots, w_{m+n}]$. This re-indexing ensures that each word in the aligned sentences has a distinct index.
For cases where words are concatenated within a sentence, we also re-index. For example, if two words $[w_{i}, w_{j}]$ are combined in a sequence like $[\dots, [w_{i-1}], [w_{i}, w_{j}], [w_{j+1}], \dots]$, their indices are re-assigned as $[\dots, [w_{(i-1)}], [w_{(i,1)}, w_{(i,2)}], [w_{(j)}], \dots]$, where $w_{(i,1)}$ and $w_{(i,2)}$ represent the re-indexed words in a tuple.
In terms of results, we ensure that both the aligned sentences $L'$ and $R'$ have an equal number of sentence lists, and similarly, the aligned words $l$ and $r$ have an equal number of word lists after re-indexing.

\subsection{Proof}

We define a perfect sentence alignment pair as $L_{\not\sqcup} = R_{\not\sqcup}$. Additionally, we define another perfect matched pair as $L'{\not\sqcup} = R'{\not\sqcup}$, obtained by accumulating unmatched lines from $L$ and $R$ when $L_{\not\sqcup} \neq R_{\not\sqcup}$. Our goal is to demonstrate that such an alignment process can lead to $L'$ and $R'$, where ${L_{\not\sqcup}}' = {R_{\not\sqcup}}'$.

\paragraph{Soundness}
The algorithm is sound if, given true premises, it produces a true result, i.e., it leads to a tautology. According to the Soundness Theorem, Algorithm $A$ is sound under the following three cases:
Let $\mathcal{L}, \mathcal{R}, L'$, and $R'$ be elements of $A$.

\begin{case}
Algorithm~\ref{alignment-algorithm} is sound if
$L_{\not\sqcup} \vdash L$ implies $L_{\not\sqcup} \models_{\text{taut}} L$ and
$R_{\not\sqcup} \vdash R$ implies $R_{\not\sqcup} \models_{\text{taut}} R$.
\end{case}
\textit{Explanation:} This case demonstrates that if $L_{\not\sqcup}$ and $R_{\not\sqcup}$, which are the non-tokenized representations of $L$ and $R$, respectively, are consistent with their tokenized forms, the algorithm holds soundness.

\begin{case}
Algorithm~\ref{alignment-algorithm} is sound if $({L_{\not\sqcup}} \land L_{\sqcup}) \vdash L$ implies $({L_{\not\sqcup}} \land L_{\sqcup}) \models_{\text{taut}} L$ and
$({R_{\not\sqcup}} \land R_{\sqcup}) \vdash R$ implies $({R_{\not\sqcup}} \land R_{\sqcup}) \models_{\text{taut}} R$.
\end{case}
\textit{Explanation:} This case shows that combining tokenized ($L_{\sqcup}$ and $R_{\sqcup}$) and non-tokenized ($L_{\not\sqcup}$ and $R_{\not\sqcup}$) representations for both $L$ and $R$ still preserves soundness in the algorithm. If both the tokenized and non-tokenized versions align with the original, the alignment process remains valid.

\begin{case}
Algorithm~\ref{alignment-algorithm} is sound if
${L_{\not\sqcup}} \vdash L'$ implies ${L_{\not\sqcup}} \models_{\text{taut}} L'$ and
${R_{\not\sqcup}} \vdash R'$ implies ${R_{\not\sqcup}} \models_{\text{taut}} R'$.
\end{case}
\textit{Explanation:} This case illustrates that even after accumulating and aligning lines, the resulting $L'$ and $R'$ remain consistent with their original non-tokenized representations, $L_{\not\sqcup}$ and $R_{\not\sqcup}$. Therefore, the alignment is sound.

\paragraph{Correctness}
To prove the correctness of the statement $L' = R'$, we assume that the algorithm aligns two sequences $L$ and $R$ such that the final outputs $L'$ and $R'$ have equal values.

Let $L' = L_{i}, L_{i+1}, \cdots, L_{n}$ and $R' = R_{j}, R_{j+1}, \cdots, R_{m}$ for some $i, j, n, m \in \mathbb{Z}^{\geq 0}$.

\begin{case}
${L_{i+1} = R_{j+1}}$.
For any $i$ and $j$, if $i+1$ and $j+1$ are such that $i, j \in \mathbb{Z}^{\geq 0}$, then $L_{i+1}$ and $R_{j+1}$ have the same value.
This implies that for each step of the algorithm, if $L_{i+1}$ and $R_{j+1}$ are equal, they are aligned together. Since both $L$ and $R$ remain in the domain of the input space, $L_{i+1}$ and $R_{j+1}$ will be accumulated together into $L'$ and $R'$.
\end{case}

\begin{case}
$L_{n+1} = R_{m+1}$.
For any $n$ and $m$, if $n+1$ and $m+1$ are such that $n, m \in \mathbb{Z}^{\geq 0}$, then $L_{n+1}$ and $R_{m+1}$ are aligned.
This implies that as the algorithm processes $L$ and $R$, the elements $L_{n+1}$ and $R_{m+1}$ will be accumulated together, ensuring that the alignment is maintained through the process.
\end{case}

\noindent Therefore, for all indices $i, j, n, m \in \mathbb{Z}^{\geq 0}$, the output sequences $L'$ and $R'$ are correctly aligned, and $L' = R'$ holds.

\section{Limitations and notes}

The jp-algorithm targets monolingual alignment at the sentence and word levels.
Within the jointly preprocessed evaluation framework, its role is to reconcile segmentation differences prior to scoring, so that system outputs and gold references are compared on equivalent structural units.

The pattern-based alignment procedure is therefore an implementation choice rather than a defining property of the framework.
In principle, any sentence alignment method may be substituted, provided that it produces aligned sentence pairs that support the same jointly preprocessed workflow.
The same modularity holds at the word level.

Alternative alignment strategies include edit-distance-based alignment and probabilistic alignment models.
Edit-distance methods are naturally suited to monotonic, linearly edited strings and are less appropriate when mismatches involve non-local reconfiguration.
Probabilistic models such as IBM-style alignment require training and parameter estimation.
By design, the present method is deterministic and parameter free, which makes it convenient as a default instantiation.

The jp-framework should thus be read as a general evaluation methodology centered on joint preprocessing, not as a commitment to a particular alignment algorithm.
The procedure in this chapter is one concrete realization of that methodology.

\singlespacing
\chapter{Applications of \textsc{JP-Algorithm}} \label{applications-of-jp-algorithms}
\doublespacing

This chapter demonstrates how {JP-Algorithm} functions as a shared alignment layer for evaluation settings in which gold and system outputs are intended to represent the same text but diverge due to segmentation and preprocessing decisions. We instantiate the same core idea across three applications. {JP-preprocessing} evaluates tokenization and sentence boundary detection by aligning system outputs to gold references to prevent partial matches from being miscounted. {JP-evalb} extends PARSEVAL evaluation for constituent parsing by inserting sentence and token alignment as a preprocessing step, enabling robust scoring under mismatched inputs without changing the underlying metric definition. {JP-errant} adapts grammatical error correction evaluation by jointly preprocessing gold and system m2 files through sentence alignment and re indexing, preserving the original annotation scheme while supporting end to end evaluation from raw text.

\section{\textsc{JP-preprocessing}} \label{jp-preprocessing}

\subsection{Introduction}

Recognizing and adjusting for errors is crucial to achieve dependable outcomes in most computer-based language tasks. However, this step often does not get the attention it deserves because there is a common practice of accepting some level of errors. This lack of focus becomes even more problematic during the initial stages of text processing, particularly when dealing with tasks like tokenization and sentence boundary detection (SBD).
Given that most commonly used preprocessing methods are generally recognized as well-established and can be easily reused, people tend to think that challenges like tokenization and SBD have already been completely solved. However, what often goes unnoticed is the fact that there has been a longstanding lack of attention to the evaluation process, which has been underestimated despite the apparent success of these tools.

In response to these challenges, the literature has seen various efforts aimed at addressing and improving the limitations of tokenization and SBD. These efforts include extensive comparisons and detailed reevaluations for tokenization \citep{dridan-oepen-2012-tokenization} and  for SBD \citep{read-EtAl:2012:COLING}. Although these reevaluations provide comprehensive and convincing results, they often fall short in offering clear and adaptable configurations that can be easily re-implemented by others.

When you examine these efforts side by side, it becomes evident that the tasks of   sentence and word segmentation are closely related and complement each other more than they are typically considered as separate steps. In response to this, some innovative sequence labeling methods \citep{evang-EtAl:2013:EMNLP}, have been proposed, showing promising results in terms of reproducibility across various domains. However, there are still noticeable challenges related to sentence boundaries that may remain unresolved in the evaluation process.

\subsection{Mismatches in preprocessing} \label{issues}

\subsubsection{Tokenization} 

The primary challenge in tokenization, as demonstrated by several methods, revolves around handling the ambiguity related to sentence periods and contractions \citep{dridan-oepen-2012-tokenization}. For instance, consider the sentence.\textit{When No. 1 Isn't the Best}\footnote{\url{https://www.washingtonpost.com/archive/sports/2004/05/26/when-no-1-isnt-the-best/fa34156f-881a-4181-b0fe-4447e2f36f0e/}} Different tokenization schemes, such as \texttt{tok.sed}\footnote{\url{ftp://ftp.cis.upenn.edu/pub/treebank/public_html/tokenization.html}}, \texttt{Moses} \citep{koehn-EtAl:2007:PosterDemo}, and \texttt{CoreNLP} \citep{manning-etal-2014-stanford}, tokenize the contraction and abbreviation of common words in this sentence differently.

\begin{center}
\footnotesize{
\begin{tabular} {r l}
\texttt{tok.sed} & When No$\sqcup$. 1 is$\sqcup$n't the Best \\
\texttt{Moses} & When No. 1 isn$\sqcup$'t the Best \\
\texttt{CoreNLP} & When No. 1 is$\sqcup$n't the Best 
\end{tabular}
}
\end{center}
\noindent where $\sqcup$ is a symbol for a token delimiter.
Typically, periods are treated as individual tokens in text processing. However, due to the common use of word-final periods in conjunction with abbreviations, acronyms, and named entities, there is a lack of consistency in how these cases are handled. This inconsistency results in variations in tokenization approaches for these words.
While there are various tokenization methods available, \texttt{CoreNLP} results suggest that, in practice, the terminal leaf nodes in the Penn treebank \citep{marcus-etal-1993-building,taylor-marcus-santorini:2003} have become a widely accepted standard tokenization scheme for English.

\subsubsection{Sentence boundary detection} 

The task of sentence boundary detection involves identifying the points where sentences start and end. In many prior studies \citep{palmer-hearst:1997,reynar-ratnaparkhi:1997:ANLP,kiss-strunk:2006,gillick:2009:NAACL,lu-ng:2010:EMNLP}, sentence boundary disambiguation has been treated as a classification problem. It is important to note that sentence boundary disambiguation is distinct from sentence boundary detection. The former requires the use of punctuation marks for classification, while the latter does not necessarily rely on punctuation to determine sentence boundaries.
A significant challenge in sentence boundary disambiguation arises when dealing with text that lacks consistent punctuation marks. To tackle this issue, \citet{evang-EtAl:2013:EMNLP} proposed a method involving supervised character-level sequence labeling for both sentence and word segmentation. Additionally, \citet{lim-park-2024-NLE} annotated a dataset with rich linguistic information to improve sentence boundary detection.
However, many common sentence boundary detection systems, such as \texttt{splitta} \citep{gillick:2009:NAACL}, \texttt{CoreNLP} \citep{manning-etal-2014-stanford}, and \texttt{Elephant} \citep{evang-EtAl:2013:EMNLP}, often fail to correctly identify sentence boundaries in cases where there are no punctuation marks between two sentences, even when their methods are not designed to rely on punctuation marks. For instance, this issue can be observed between \textit{... session} and \textit{I ...} in the following text extracted from the Europarl parallel corpus \citep{koehn:2005}:

\begin{center}
\footnotesize{
\begin{tabularx}{.8\textwidth}{|X|}\hline
\texttt{Opening of the session I declare resumed the 2000-2001 session of the European Parliament.} \\ \hline
\end{tabularx}
}
\end{center}

Additionally, textual content often exhibits improper formatting, including issues such as missing or incorrect punctuation, inconsistent casing, and the presence of unnecessary symbols. For instance, in social media text,  individuals may write in all uppercase or with incorrect punctuation and casing, or entirely omit punctuation. 
In the domain of automatic speech recognition (ASR) where there are no punctuation marks, several SBD-related works have been proposed \citep{treviso-shulby-aluisio:2017,gonzalezGallardo-torresMoreno:2018}. 
Punctuation restoration in ASR has also been explored \citep{fu-etal-2021-improving,alam-etal-2020-punctuation,polacek-et-al-2023-online}.

\subsubsection{Evaluation}

Methods for evaluating sentence boundary disambiguation and tokenization can be broadly categorized into three main approaches:

\begin{description} \setlength\itemsep{0em}
\item[Accuracy and error rate] This method assesses how accurately the boundaries are detected or tokens are separated, and quantifies the rate of errors made.

\item[Precision and recall for F1 Measure] This approach focuses on measuring the precision (how many of the detected boundaries or tokens are correct) and recall (how many of the actual boundaries or tokens are detected). The F1 measure is then calculated to balance these two factors.

\item[Levenshtein distance] This method involves calculating the Levenshtein distance, which represents the number of edits (insertions, deletions, or substitutions) needed to transform the detected boundaries or tokens into the correct ones. 
\end{description}
\noindent These three evaluation approaches help assess the performance and accuracy of sentence boundary detection and tokenization methods.

In previous SBD tasks, such as the one mentioned in \citet{dridan-oepen-2012-tokenization}, each character position is considered a potential boundary point. If a gold-standard boundary is missed, it is counted as a false negative. This imbalance between negative and positive cases makes precision and recall for F1 evaluation metrics more informative than accuracy or error rate metrics. 
To address these issues, character-level methods aim to facilitate the integration of evaluations between the two tasks, as proposed by \citet{evang-EtAl:2013:EMNLP}. However, even at the character level, the imbalance problem persists, especially in cases where there are partial matches, which can lead to miscounted true positive cases.
For example, in Figure~\ref{evaluation-previous-work}, within the context of tokenization, a perfect alignment between the system output and the reference gold standard is observed, yielding an F1 measure of 1.0 (100\%).
However, when it comes to sentence boundary detection (SBD), there's a difference. Using previous methods, \textit{Click here} and \textit{To view it.} are identified as two separate sentences in SBD, with two "S" labels marking sentence boundaries under \textit{Click} and \textit{To}, as shown in Figure~\ref{evaluation-previous-work}. 
However, the first sentence boundary marked by \textit{Click} is essentially a partial match, while the second part is separated into another sentence by the second sentence boundary marked by \textit{To}.

\begin{figure}[!ht]
\centering
\footnotesize{
\begin{tabular}{rl}
& \texttt{Click here To view it.} \\
\textsc{sys} & \texttt{{\color{red}S}IIIIOTIIIO{\color{red}S}IOTIIIOTIT}\\
\textsc{gold} & \texttt{{\color{red}S}IIIIOTIIIOTIOTIIIOTIT}\\
\end{tabular} }
\caption{BIO-style evaluation in previous work where partial matched sentences ({e.g} \textit{Click here}) could be considered as \texttt{true positive}.}
\label{evaluation-previous-work}
\end{figure}

The proposed evaluation using the alignment algorithm aims to prevent such partially matched sentences from being incorrectly counted during evaluation. This is achieved by incorporating sentence alignment methods borrowed from machine translation (MT), addressing various preprocessing issues along the way.

\subsection{Alignment-based evaluation} \label{evaluation-by-alignment-section}

We begin by employing a fundamental algorithm that leverages sentence and word alignment techniques typically used in MT.
In MT and other cross-language applications, sentence and word alignment have been crucial sub-tasks. They are often employed when working with parallel corpora, where sentences in one language need to be aligned with their translations in another language. This necessity arises from the fact that when two sentences are direct translations of each other, they can differ significantly in terms of word order, length, and structure. Therefore, to enable downstream cross-language tasks, it's often essential to perform comprehensive structural adjustments by establishing sentence and word alignments between the source and target languages. These alignment methods, used to address inter-lingual differences, shed light on the challenges we discuss in Section~\ref{issues} regarding evaluation.

However, when dealing with monolingual inputs from both the system and gold results, there are no inter-lingual differences between sequences on each side ($L$ and $R$). Instead, these sequences are essentially identical sentences, differing only in terms of token and sentence boundaries. Nevertheless, when evaluating tokenization and SBD tasks, they share a commonality with cross-language tasks: for a corresponding sentence pair from the system and gold results, even if they are identical at the character level, they can still vary in length across lines due to differences in tokenization and SBD results, as demonstrated in the \texttt{system result} and \texttt{gold file} in Figure~\ref{evaluation-by-alignment-example}.
This is where the alignment approach used in MT becomes useful in the evaluation algorithm we propose. Figure~\ref{evaluation-by-alignment-example} provides a simplified example of how our proposed evaluation, facilitated by the alignment process, would resolve the partial match issue discussed in the previous section.

\begin{figure}[!ht]
\centering
{\footnotesize
\begin{tabularx}{\textwidth}{|XX|} \hline
\multicolumn{2}{|l|}{\textbf{\textsc{input file}:}} \\
\multicolumn{2}{|l|}{\texttt{Click here To view it. He makes some good} $\sqcap$}\\
\multicolumn{2}{|l|}{\texttt{observations on a few of the picture's.} $\sqcap$} \\ \hline

\multicolumn{2}{|l|}{\textbf{\textsc{system result}:}} \\
\multicolumn{2}{|l|}{\texttt{Click here} $\sqcap$}\\
\multicolumn{2}{|l|}{\texttt{To view it$\sqcup$. $\sqcap$}}\\
\multicolumn{2}{|l|}{\texttt{He makes some good observations on a few of the picture$\sqcup$'s$\sqcup$. $\sqcap$}} \\ \hline
\multicolumn{2}{|l|}{\textbf{\textsc{gold file}:}} \\
\multicolumn{2}{|l|}{\texttt{Click here To view it$\sqcup$. $\sqcap$}}\\
\multicolumn{2}{|l|}{\texttt{He makes some good observations on a few of the picture$\sqcup$'s$\sqcup$. $\sqcap$}} \\ \hline

\multicolumn{2}{|l|}{\textbf{\textsc{sentence-aligned result for preprocessing evaluation}:}} \\
\texttt{Click here $\sim\sim\sim$ To view it$\sqcup$. $\sqcap$} & \texttt{Click here To view it$\sqcup$. $\sqcap$}\\
\texttt{He makes some good observations on a few of the picture$\sqcup$'s$\sqcup$. $\sqcap$} & \texttt{He makes some good observations on a few of the picture$\sqcup$'s$\sqcup$. $\sqcap$} \\ \hline
\end{tabularx}
}
\caption{Example of intermediate results of the evaluation by alignment algorithm where \textit{Click here} and \textit{To view it.} In the system result are the realigned, produced by merging after the implementation of sentence alignment: {$\sqcup$ is a symbol for tokenization, $\sqcap$ is a symbol for SBD, and $\sim\sim\sim$ is a symbol for merged sentences by sentence alignment.}}
\label{evaluation-by-alignment-example}
\end{figure}

While Figure~\ref{evaluation-by-alignment-example} presents sentence alignment for the evaluation of sentence boundary detection, a same approach will be employed for tokenization evaluation, utilizing word alignment. 

\subsubsection{Algorithm description}

\paragraph{Notations}
To describe our proposed algorithms more concisely and clearly, we'll use the following notations:
$\mathcal{L}$ and $\mathcal{R}$ represent the system output and the gold files for preprocessing, respectively.
We assume \textsc{len}(${\mathcal{L}_{\not\sqcup}}$) and \textsc{len}(${\mathcal{R}_{\not\sqcup}}$) are equal, where ${\mathcal{L}_{\not\sqcup}}$ and ${\mathcal{R}_{\not\sqcup}}$ are sequences of characters with all spaces removed from $\mathcal{L}$ and $\mathcal{R}$.
$L$ and $R$ denote the current sentence pair from the system output and the gold files. The following notations are also used interchangeably for both $L$ and $R$:

\begin{description} 
\setlength{\itemsep}{0pt} 
\item $\mathcal{C}_{sb}(\mathcal{L})$ and $\mathcal{C}_{tk}(\mathcal{L})$: These represent the total number of sentence boundaries (${sb}$) and the total number of tokens (${tk}$) in $\mathcal{L}$.
\item $\texttt{TP}_{\texttt{sb}}$ and $\texttt{TP}_{\texttt{tk}}$: These denote the number of true positives (\texttt{tp}) for sentence boundaries and tokens, respectively.
\item $L_{\sqcup}$: This represents $L$ where spaces between tokens have not been removed.
\item $L_{\not\sqcup}$: This represents $L$ where spaces between tokens have been removed.
\item $L_{i}$: It stands for the \textit{i}th token in $L_{\sqcup}$.
\end{description}

\paragraph{Evaluation measures}
To calculate an F1 score for SBD, we need to have the total counts of sentence boundaries in both the system output and the gold standard ($\mathcal{C}_{sb}(\mathcal{L})$ and $\mathcal{C}_{sb}(\mathcal{R})$, respectively), as well as the number of true positives for sentence boundaries ($\texttt{TP}_{\texttt{sb}}$).
Precision and recall, for example, are defined as follows, considering sentence boundary detection:

\begin{align*}
\text{precision} &= \frac{\text{relevant \# of \texttt{sb}} \cap \text{retrieved \# of \texttt{sb}}}{\text{retrieved \# of \texttt{sb}}}\\
&= \frac{\mathcal{C}_{sb}(\mathcal{L}) \cap \mathcal{C}_{sb}(\mathcal{R})}{\mathcal{C}_{sb}(\mathcal{L})} = \frac{\texttt{TP}_{\texttt{sb}}}{\mathcal{C}_{sb}(\mathcal{L})} \\
\text{recall} &= \frac{\text{relevant \# of \texttt{sb}} \cap \text{retrieved \# of \texttt{sb}}}{\text{relevant \# of \texttt{sb}}}\\
&= \frac{\mathcal{C}_{sb}(\mathcal{L}) \cap \mathcal{C}_{sb}(\mathcal{R})}{\mathcal{C}_{sb}(\mathcal{R})} = \frac{\texttt{TP}_{\texttt{sb}}}{\mathcal{C}_{sb}(\mathcal{R})} 
\end{align*}
To apply in the algorithm, `$\text{relevant \# of \texttt{sb}} \cap \text{retrieved \# of \texttt{sb}}$' represents $\texttt{TP}_{\texttt{sb}}$, `retrieved \# of \texttt{sb}' is $\mathcal{C}_{sb}(\mathcal{L})$ by a system result, and `relevant \# of \texttt{sb}' is $\mathcal{C}_{sb}(\mathcal{R})$ by a gold file.
The same precision and recall can be defined for tokenization using $\mathcal{C}_{tk}(\mathcal{L})$, $\mathcal{C}_{tk}(\mathcal{R})$ and $\texttt{TP}_{\texttt{tk}}$.

\paragraph{Sentence alignment for SBD evaluation}
The proposed alignment algorithm processes $\mathcal{L}$ and $\mathcal{R}$ and increments the \texttt{TP$_{sb}$} count if $L_{\not\sqcup}$ is equal to $R_{\not\sqcup}$. However, when they do not match, the algorithm merges sentences in $L'$ and $R'$ for mismatched sentence pairs until $L'$ and $R'$ are identical (practically, this is done until the following $L_{\not\sqcup}$ and the following $R_{\not\sqcup}$ are identical instead of verifying $L'$ and $R'$ being identical). After this merging process, the algorithm continues to increment \texttt{TP$_{sb}$} if $L_{\not\sqcup}$ is equal to $R_{\not\sqcup}$.

\paragraph{Word alignment for tokenization evaluation}
In the context of our evaluation algorithm, we increment the \texttt{TP$_{tk}$} count in the word-aligned $L_{\sqcup}$ and $R_{\sqcup}$ pair if $L_{i}$ is equal to $R_{j}$, where $L_{i}$ and $R_{j}$ represent the \textit{i}-th and \textit{j}-th tokens in $L_{\sqcup}$ and $R_{\sqcup}$, respectively.
While IBM word alignment can be adapted for token alignment within our evaluation approach, our focus lies in aligning tokens within the specific sentence pair expected to be identical, rather than traversing the entire document as IBM models are designed to find the proper lexical translation pair between sentence pairs.
For our evaluation algorithm, token alignment only needs to traverse each aligned sentence pair within $L_{\sqcup}$ and $R_{\sqcup}$, directly assessing token equivalence within the corresponding sentence pair.

\paragraph{Summary}
Drawing an analogy from sentence and word alignment, we introduced an alignment-based algorithm that significantly enhances the efficiency of evaluating  sentence boundary detection (SBD) and tokenization results. 
This algorithm allows us to evaluate both tasks in a single pass through the input. Through the alignment of sentences and tokens, we can streamline the evaluation process, resulting in more precise counts of true positive cases for both tokenization and SBD results.

\subsubsection{Pseudo-codes}

\paragraph{Alignment algorithm}
The alignment algorithm is employed in two key scenarios: (a) to align sentences between the system and the gold results, and (b) to align tokens within the current sentence pair.
Once the alignment is established, we can conduct a side-by-side comparison between the system and gold results. This comparison allows us to discern and integrate all the intertwined matched and mismatched cases between sentence boundary detection (SBD) and tokenization.
Algorithm~\ref{alignment-algorithm} provided a generic representation of the alignment algorithm, which we can apply to both sentence boundary and tokenization evaluation. The inner \texttt{while} statement, where mismatches are addressed, is part of an iterative process controlled by the outer \texttt{while} loop, which iterates through each element. The time complexity of this algorithm is $\mathcal{O}(N + N')$, where $N$ and $N'$ represent the lengths of the left and right sentences ($L$ and $R$), respectively. Additionally, this algorithm is applicable to tokens.

\paragraph{Implementations}
We provide two comprehensive implementations of the algorithm: one as a basic algorithm (Algorithm~\ref{basic-algorithm}) and another as a joint algorithm (Algorithm~\ref{joint-algorithm}).
In Algorithm~\ref{basic-algorithm}, the evaluation procedure distinguishes between sentence boundary detection (SBD) and tokenization. Its time complexity is $\mathcal{O}(2N + 4NM)$, where $N$ represents the number of sentences and $M$ denotes the number of tokens in each sentence. We assume that $N \simeq N'$ and $M \simeq M'$ for the left ($L$) and right ($R$) sentences, respectively.
This version simplifies notations by removing indexes $i$ and $j$. During the alignment of sentences, we obtain $\texttt{TP}{\texttt{sb}}$. Subsequently, in the second outer \texttt{while} loop, we obtain $\texttt{TP}{\texttt{tk}}$ during word alignment, where $L'_{i,i'}$ represents the $i'$-th token in the $i$-th sentence of $L'$. Both iteration processes resemble those in Algorithm~\ref{alignment-algorithm}.
In contrast, Algorithm~\ref{joint-algorithm} processes SBD and tokenization jointly, reducing the time complexity to $\mathcal{O}(4NM)$. We maintain the assumption that $N \simeq N'$ and $M \simeq M'$ for $L$ and $R$.

\begin{algorithm}[!ht]
\caption{\textsc{Basic Algorithm}}\label{basic-algorithm}
{
\begin{algorithmic}[1]
\STATE{\textbf{function} \textsc{Basic Algorithm} ($\mathcal{L}$, $\mathcal{R}$):}
\STATE{\textbf{Input}: $\mathcal{L}$, $\mathcal{R}$}
\STATE{\textbf{Output}: $\mathcal{C}_{sb}(\mathcal{L})$, $\mathcal{C}_{tk}(\mathcal{L})$, $\mathcal{C}_{sb}(\mathcal{R})$, $\mathcal{C}_{tk}(\mathcal{R})$, $\texttt{TP}_{\texttt{sb}}$, $\texttt{TP}_{\texttt{tk}}$}
\STATE{Obtain $\mathcal{C}_{sb}(\mathcal{L})$, $\mathcal{C}_{tk}(\mathcal{L})$, $\mathcal{C}_{sb}(\mathcal{R})$, and $\mathcal{C}_{tk}(\mathcal{R})$ from $\mathcal{L}$ and $\mathcal{R}$}

\WHILE{$\mathcal{L}$ and $\mathcal{R}$}
    \STATE{/* Obtain $\texttt{TP}_{\texttt{sb}}$ */}
    \IF{$L_{i{\not\sqcup}}$ == $R_{j\not\sqcup}$}
        \STATE{$\texttt{TP}_{\texttt{sb}} \gets \texttt{TP}_{\texttt{sb}} + 1$}
        \STATE{$\mathcal{L^{\prime}}$, $\mathcal{R^{\prime}}$ $\gets$ $\mathcal{L^{\prime}}+{L}_i, \mathcal{R^{\prime}}+{R}_j$}
    \ELSE
        \WHILE{${L_{i+1\not\sqcup}} \neq {R_{j+1\not\sqcup}}$}
            \STATE{$L' \gets L' + L_i$}
            \STATE{$R' \gets R' + R_j$}
        \ENDWHILE
        \STATE{$\mathcal{L^{\prime}}$, $\mathcal{R^{\prime}}$ $\gets$ $\mathcal{L^{\prime}} + L', \mathcal{R^{\prime}} + R'$}
    \ENDIF
\ENDWHILE

\WHILE{$\mathcal{L^{\prime}}$ and $\mathcal{R^{\prime}}$}
    \STATE{/* Obtain $\texttt{TP}_{\texttt{tk}}$ */}
    \IF{$L_{i\sqcup} == R_{j\sqcup}$}
        \STATE{$\texttt{TP}_{\texttt{tk}} \gets \texttt{TP}_{\texttt{tk}} + \textsc{len}(L_i\sqcup)$}
    \ELSE
        \WHILE{$L_{i\sqcup}$ and $R_{j\sqcup}$}
            \STATE{$\texttt{TP}_{\texttt{tk}} \gets \texttt{TP}_{\texttt{tk}} + 1$ \textbf{if} $L_{i,i'} == R_{j,j'}$}
        \ENDWHILE
    \ENDIF
\ENDWHILE
\RETURN {$\mathcal{C}_{sb}(\mathcal{L})$, $\mathcal{C}_{tk}(\mathcal{L})$, $\mathcal{C}_{sb}(\mathcal{R})$, $\mathcal{C}_{tk}(\mathcal{R})$, $\texttt{TP}_{\texttt{sb}}$, $\texttt{TP}_{\texttt{tk}}$}
\end{algorithmic} 
}
\end{algorithm}

\begin{algorithm}[!ht]
\caption{\textsc{Joint Algorithm}}\label{joint-algorithm}
{
\begin{algorithmic}[1]
\STATE{\textbf{function} \textsc{Joint Algorithm} ($\mathcal{L}$, $\mathcal{R}$):}
\STATE{\textbf{Input}: $\mathcal{L}$, $\mathcal{R}$}
\STATE{\textbf{Output}: $\mathcal{C}_{sb}(\mathcal{L})$, $\mathcal{C}_{tk}(\mathcal{L})$, $\mathcal{C}_{sb}(\mathcal{R})$, $\mathcal{C}_{tk}(\mathcal{R})$, $\texttt{TP}_{\texttt{sb}}$, $\texttt{TP}_{\texttt{tk}}$}
\STATE{Obtain $\mathcal{C}_{sb}(\mathcal{L})$, $\mathcal{C}_{tk}(\mathcal{L})$ from $\mathcal{L}$ and $\mathcal{C}_{sb}(\mathcal{R})$, $\mathcal{C}_{tk}(\mathcal{R})$ from $\mathcal{R}$}

\WHILE{$\mathcal{L}$ and $\mathcal{R}$}
    \STATE{/* Obtain $\texttt{TP}_{\texttt{sb}}$ */}
    \IF{$L_{i{\not\sqcup}}$ == $R_{j\not\sqcup}$}
        \STATE{$\texttt{TP}_{\texttt{sb}} \gets \texttt{TP}_{\texttt{sb}} + 1$}
        \STATE{/* Obtain $\texttt{TP}_{\texttt{tk}}$ */}
        \IF{$L_{i\sqcup} == R_{j\sqcup}$}
            \STATE{$\texttt{TP}_{\texttt{tk}} \gets \texttt{TP}_{\texttt{tk}} + \textsc{len}(L)$}
        \ELSE
            \WHILE{$L_{i\sqcup}$ and $R_{j\sqcup}$}
                \STATE{$\texttt{TP}_{\texttt{tk}} \gets \texttt{TP}_{\texttt{tk}} + 1$ \textbf{if} $L_{i,i'} == R_{j,j'}$}
            \ENDWHILE
        \ENDIF
    \ELSE
        \WHILE{${L_{i+1\not\sqcup}} \neq {R_{j+1\not\sqcup}}$}
            \STATE{$L' \gets L' + L_i$}
            \STATE{$R' \gets R' + R_j$}
        \ENDWHILE
        \WHILE{$L'_{\sqcup}$ and $R'_{\sqcup}$}
            \STATE{$\texttt{TP}_{\texttt{tk}} \gets \texttt{TP}_{\texttt{tk}} + 1$ \textbf{if} $L'_{i'} == R'_{j'}$}
        \ENDWHILE
    \ENDIF
\ENDWHILE
\RETURN {$\mathcal{C}_{sb}(\mathcal{L})$, $\mathcal{C}_{tk}(\mathcal{L})$, $\mathcal{C}_{sb}(\mathcal{R})$, $\mathcal{C}_{tk}(\mathcal{R})$, $\texttt{TP}_{\texttt{sb}}$, $\texttt{TP}_{\texttt{tk}}$}
\end{algorithmic}
}
\end{algorithm}

\paragraph{Examples}
Figure~\ref{algorithm-example} provides examples to illustrate the jp-algorithm in different scenarios.
In the first case, where $L_{\not\sqcup}$ matches $R_{\not\sqcup}$, we obtain true positive instances for sentence boundaries, denoted as $\texttt{TP}_{\texttt{sb}}$.
The second case depicts conditions where true positive sentence boundaries are not present. In such instances, the algorithm proceeds to build $L'$ and $R'$ if a sentence is segmented and requires merging (which we consider sentence alignment). After merging, these sentences should align correctly, and we identify the true positives among these pairs.
To count the number of true positives for tokens between $L$ and $R$, we obtain $\texttt{TP}_{\texttt{tk}}$ if $L_{i} == R_{j}$. 
Figure~\ref{algorithm-example-token} provides a detailed breakdown of how to evaluate and count the number of true positives (\texttt{TP}) for tokens.
This is done using the same alignment method as previously explained for sentence alignment.

\begin{figure}[!ht]
 \centering
\scriptsize{
\begin{tabularx}{\textwidth}{| c | XX | c|} \hline 
& \multicolumn{1}{c}{$\mathcal{L}$ (\textsc{system})} & \multicolumn{1}{c|}{$\mathcal{R}$ (\textsc{gold})} & \\ 
\hline 
$L_{\not\sqcup}$ == $R_{\not\sqcup}$ & \texttt{When No. 1 Isn 't the Best $\sqcap$} & \texttt{When No. 1 Is n't the Best $\sqcap$} & $\texttt{TP}$ \\
\hline

$L_{\not\sqcup}$ $\neq$ $R_{\not\sqcup}$ & \texttt{Mike McConnell 07/06/2000 14:57 John ,} & \texttt{Mike McConnell $\sqcap$} & \\
(before alignment) & \texttt{Hello from South America . $\sqcap$} & \texttt{07/06/2000 14:57 $\sqcap$} & \\
 & \texttt{} & \texttt{John , Hello from South America . $\sqcap$} & \\\hdashline
(after alignment) & \texttt{Mike McConnell 07/06/2000 14:57 John ,} & \texttt{Mike McConnell $\sim\sim\sim$ 07/06/2000 14:57} & -\\ 
 & \texttt{Hello from South America . $\sqcap$} & \texttt{$\sim\sim\sim$ John , Hello from South America. $\sqcap$} & \\ 
\hline 
\end{tabularx}
}
\caption{Examples with the jp-algorithm in Algorithm~\ref{joint-algorithm} {for \texttt{true positive} of sentences. If not, merging sentence boundaries.}}
\label{algorithm-example}
\end{figure}

\begin{figure}[!ht]
\centering
{\footnotesize 
\begin{tabular}{ |r | cccccc |} \hline
$\mathcal{L}$ (\textsc{system}) & \texttt{When} & \texttt{No.}& \texttt{1} & \texttt{Isn} $\sim\sim\sim$ \texttt{'t} & \texttt{the} & \texttt{Best} \\
{$\mathcal{R}$ (\textsc{gold})} & \texttt{When} & \texttt{No.} & \texttt{1} & \texttt{Is} $\sim\sim\sim$ \texttt{n't} & \texttt{the} & \texttt{Best} \\ \hdashline
& $\texttt{TP}$ & $\texttt{TP}$& $\texttt{TP}$& - & $\texttt{TP}$& $\texttt{TP}$ \\\hline
\end{tabular}
}
\caption{Examples {for \texttt{true positive} of tokens. If not, merging tokens.}}
\label{algorithm-example-token}
\end{figure}

\paragraph{Discussion}
The length-based sentence alignment algorithms, like the one described by \citet[p.83]{gale-church:1993:CL}, typically consider matches in the ratios of 1:0, 0:1, 1:1, 2:1, 1:2, and 2:2. However, we need to account for cases where the system segments a sentence into more than two sentences or where gold sentence boundaries are segmented into more than two sentences.
In other words, even after using sentence alignment, the segmented results of $L_{\not\sqcup}$ and $R_{\not\sqcup}$ may still differ from each other if we apply pre-existing sentence alignment algorithms from MT. 
To address this, we accumulate and merge $L$ and $R$ together until their characters match, resulting in $L' == R'$ instead of using MT's alignment algorithm.
As described in Algorithm~\ref{joint-algorithm}, the 'else' condition (when $L_{\not\sqcup} \neq R_{\not\sqcup}$) entails aggregating $L'$ and $R'$ to form a matched sentence pair between $L$ and $R$. This process allows for the accumulation of pairs such as $m$:1, 1:$n$, or $m$:$n$ sentence segments.

\subsection{Experiments and results} \label{jp-preprocessing-case-studies}

\subsubsection{Case study on English corpora}
For our case studies, we conduct preprocessing on five raw input files sourced from English Universal Dependencies \citep{nivre-EtAl:2016:LREC,nivre-etal-2020-universal}. These files include: 
(1) Universal Dependencies syntax annotations from the \texttt{GUM} corpus.
(2) A multilingual parallel treebank known as \texttt{ParTUT}, developed at the University of Turin.
(3) A gold standard Universal Dependencies corpus for English, constructed using the source material of the English Web Treebank (\texttt{EWT}).
(4) The English portion of the parallel Universal Dependencies (\texttt{PUD}) treebanks.
(5) The English half of the \texttt{LinES} Parallel Treebank.
(6) A dataset specifically created for pronoun identification (\texttt{Pronouns}).

We employ the \texttt{nltk} library \citep{loper-bird:2002:ACL02-ETMTNLP, bird-klein-loper:2009} along with its \texttt{word\_tokenizer} and \texttt{sent\_tokenizer} for both sentence boundary detection (SBD) and tokenization tasks.
We also utilize the \texttt{stanza} toolkit \citep{qi-etal-2020-stanza}, which is a natural language processing toolkit based on Dozat's biaffine attention dependency parser \citep{dozat-manning:2017:ICLR}. This toolkit includes a standard tokenizer with built-in sentence boundary detection, enabling us to generate text and CoNLL-U format\footnote{\url{https://universaldependencies.org/format.html}}  outputs. We use these outputs to evaluate preprocessing results using our alignment-based evaluation algorithm. 
Furthermore, we juxtapose our evaluation findings with those of prior methods, such as the evaluation script utilized in the CoNLL 2018 Shared Task \citep{zeman-etal-2018-conll}, employing the CoNLL-U format outputs generated by \texttt{stanza}.

Both \texttt{nltk} and \texttt{stanza} handles tokenization and sentence boundary detection similarly in some aspects. 
For instance, they both split words like \textit{gift's} into two tokens, \textit{gift} and \textit{'s}, and treat periods as separate tokens. However, they differ in their treatment of certain contentious areas, such as words containing dashes. For example, in the case of \textit{search-engine} found in EWT, \texttt{nltk} considers it as one token, while \texttt{stanza} separates it into three tokens: \textit{search}$\sqcup$\textit{-}$\sqcup$\textit{engine}.
Another difference arises in how they identify sentence boundaries, especially in cases where the text lacks capitalized beginnings or period endings. This ambiguity is more pronounced in sentences without clear markers.
Table~\ref{results-case-studies} presents the results of case studies, including the numbers of true positives (TP), false positives (FP), and false negatives (FN) for both sentence boundaries and tokens. These results were obtained using the proposed alignment-based evaluation algorithm with both \texttt{nltk} and \texttt{stanza}. Additionally, we provide the results of the CoNLL evaluation script, which was applied to the SBD and tokenization results produced by \texttt{stanza} and formatted in CoNLL-U format.

\begin{table}[!ht]
\centering
\resizebox{\textwidth}{!}
{\footnotesize
\begin{tabular}{r l| ccc | ccc | ccc | ccc | ccc | ccc} \hline
 & & \multicolumn{3}{c|}{GUM} & \multicolumn{3}{c|}{ParTUT} & \multicolumn{3}{c|}{EWT} & \multicolumn{3}{c|}{PUD} & \multicolumn{3}{c|}{LinES}& \multicolumn{3}{c}{Pronouns}\\ 
& & TP & FP & FN & TP & FP & FN& TP & FP & FN& TP & FP & FN& TP & FP & FN& TP & FP & FN \\ \hline 
sbd & \texttt{nltk \& jp}  
 & 845 &        137 &        251
 & 149 &          2 &          4
 & 1084 &        349 &        993
 & 976 &         26 &         24
 & 865 &        141 &        170
 & 285 & 0 & 0 \\
&\texttt{stanza \& jp} 
 & 1038 &         38 &         58
 & 144 &          6 &          9
 & 1805 &        160 &        272
 & 998 &          4 &          2
 & 912 &        128 &        123
 & 285 &          0 &          0 \\
\hdashline

&{\texttt{stanza}* \& conllu}
 & 1038 &         38 &         58
 & 144 &          6 &          9
 & 1805 &        160 &        272
 & 998 &          4 &          2
 & 912 &        128 &        123
 & 285 &          0 &          0 \\
\hline 

tk & \texttt{nltk \& jp} 
 & 19443 &        351 &        462
 & 3345 &         27 &         63
 & 24176 &       1109 &        918
 & 20733 &        251 &        443
 & 17571 &         56 &        104
 & 1673 &         16 &         32 \\

& \texttt{stanza \& jp} 
 & 19791 &        118 &        114
 & 3381 &         20 &         27
 & 24724 &        286 &        370
 & 21162 &         18 &         14
 & 17517 &        380 &        158
 & 1649 &         28 &         56 \\
\hdashline

&{\texttt{stanza}* \& conllu} 
 & 19791 &        118 &        114
 & 3381 &         20 &         27
 & 24724 &        286 &        370
 & 21162 &         18 &         14
 & 17517 &        380 &        158
 & 1649 &         28 &         56 \\
\hline 

\end{tabular}
}
\caption{Numbers of TP (true positive), FP (false positive) and FN (false negative) using \texttt{nltk} and \texttt{stanza} for sentence boundaries and tokens. {The \texttt{stanza*} line provides result numbers by the CoNLL-U evaluation script.} }
\label{results-case-studies}
\end{table}

The previous evaluation method also utilized precision and recall for F1 measures as evaluation metrics, relying on true positives (TP), false positives (FP), and false negatives (FN). 
In the CoNLL 2018 Shared Task evaluation script,\footnote{\url{https://universaldependencies.org/conll18/conll18_ud_eval.py}} both tokens and sentences are treated as spans. 
In the case of a character-level mismatch in the positions of spans between the system output and the gold file, the script adjusts by skipping to the next token in the file with the smaller start value until the positions align. This process is also applied to sentence boundaries. The start and end values of sentence spans are compared between the system and the gold file, with matching values incrementing the count of correctly matched sentences (true positive sentence boundaries).
However, in our alignment-based evaluation method, this process is limited to the aligned sentence pair, ensuring accuracy in the counts. Any miscounted true positives can significantly impact the evaluation results negatively. Therefore, we address these issues in the existing evaluation scripts, highlighting them as sources of mismatches, and suggests adjusted alternatives to enhance the reliability of sentence preprocessing evaluation.

Due to the differing characters in these representations, our character-level evaluation of both \texttt{nltk} and \texttt{stanza} preprocessing results may not capture such cases accurately.
The lack of consensus on tokenization across different corpora, including Universal Dependencies, contributes to the mismatch issue. Notably, EWT tokenizes \textit{can't} as \textit{ca} and \textit{n't}, while ParTUT tokenizes it as \textit{can} and \textit{not}.
We identified variations in the representation of contractions like \textit{can't} and \textit{ain't}. These contractions can be expressed in multiple ways, where EWT tokenizes \textit{can't} as \textit{ca} and \textit{n't}, while ParTUT tokenizes it as \textit{can} and \textit{not}.
The same issue can arise when converting "starting quotes" (\verb+``+) and "ending quotes" (\verb+''+) in the corpus into straight quotes (\verb+"+) in \texttt{nltk}, resulting in discrepancies. 
Since the number of contractions and symbols to convert, such as quotes, in a language is limited, we have created an exception list for our system to capture such cases in English.
During the final stages of our implementation, we cross-check against the exception list to ensure that every case can be correctly handled by the proposed algorithm.
As a result, the algorithm effectively addresses the preprocessing mismatches discussed in Section~\ref{issues}, which could otherwise disrupt our evaluation procedures.

\subsubsection{Case study on Multilingual corpora}
While we have addressed tokenization mismatches, such as the representation of contractions, other tokenization issues may arise from morphological segmentation or analysis, where additional morphemes can be introduced during the analysis rather than through mechanical token segmentation. Table~\ref{multilingual} presents the results of case studies using seven GSD corpora \citep{mcdonald-etal-2013-universal} in UD, which have been provided by Google. Our results closely align with those suggested by the D evaluation script, except for French.
In French GSD, tokenization often combines several units into a single token. For example, the expression \textit{de 1 000 mètres} (`of 1,000 meters') is treated as three tokens instead of four, as spaces are used to separate the words. Notably, this discrepancy is not a limitation of the proposed algorithm but rather stems from differences in tokenization conventions between plain text and the \textit{rich} CoNLL-U format. Even when the input text is \textit{de 1 000 mètres}, with \textit{1} and \textit{000} separated, it is tokenized as separated in plain text.
Such examples occur across various treebanks. For example, in \texttt{UD\_Kurmanji-MG}, phrases like \textit{dagir kiriye} (`occupied') are tokenized as a single unit.

\begin{table}[!ht]
\centering
\resizebox{\textwidth}{!}
{\footnotesize
\begin{tabular}{r l| ccc | ccc | ccc | ccc | ccc | ccc | ccc } \hline
 & & \multicolumn{3}{c|}{DE} & \multicolumn{3}{c|}{ES} & \multicolumn{3}{c|}{FR} & \multicolumn{3}{c|}{ID} & \multicolumn{3}{c|}{JA}& \multicolumn{3}{c|}{KO} 
 & \multicolumn{3}{c}{PT} \\
& & TP & FP & FN & TP & FP & FN& TP & FP & FN& TP & FP & FN& TP & FP & FN& TP & FP & FN 
& TP & FP & FN\\\hline 
sbd 
&\texttt{stanza \& jp} 
 & 787& 110& 190
 & 373& 35& 53
 & 393& 23& 23
 & 515& 35& 42
 & 541& 5& 2
 & 629& 143& 360
 & 1161& 71& 39\\
\hdashline

&{\texttt{stanza}* \& conllu}
 & 787 & 110 & 190
 & 373 & 35 & 53
 & 393 & 23 & 23
 & 515 & 35 & 42
 & 541 & 5 & 2
 & 629 & 143 & 360
 & 1161 & 71 & 39 \\
\hline 

tk 
& \texttt{stanza \& jp} 
 & 16172 &         61 &         52
 & 11705 &         33 &         28
 & 9714 &         17 &         22
 & 11523 &          9 &         18
 & 12750 &        361 &        284
 & 11332 &        184 &        345
 & 29311 &         52 &         50\\
\hdashline

&{\texttt{stanza}* \& conllu} 
 & 16172 & 61 & 52
 & 11705 & 33 & 28
 & 9710 & 19 & 23
 & 11523 & 9 & 18
 & 12750 & 361 & 284
 & 11332 & 184 & 345
 & 29311 & 52 & 50\\
\hline 

\end{tabular}
}
\caption{Numbers of TP , FP and FN using \texttt{stanza} for sentence boundaries and tokens for multilingual case studies using \texttt{UD\_*-GSD} where \texttt{*} is German, Spanish, French, Indonesian, Japanese, Korean, and  Portuguese.}
\label{multilingual}
\end{table}

\subsubsection{Discussion and limitation}

The effectiveness of the proposed word alignment approach would remain unaffected even in the presence of significant morphological mismatches. 
For example, we trace back to the sentence in Hebrew \citep{tsarfaty-etal-2012-joint} as a word mismatch example caused by morphological analyses:
\begin{center}
\footnotesize{
\begin{tabular}{r ccc cccc}
gold & $^{0}$\textit{B} & \multicolumn{2}{c}{$^{1.0}$\textit{H}  $\sim\sim\sim$ $^{1.1}$\textit{CL}} & $^{2}$\textit{FL} & $^{3}$\textit{HM} & \multicolumn{2}{c}{$^{4.0}$\textit{H} $\sim\sim\sim$ $^{4.1}$\textit{NEIM}} \\
& 'in' & \multicolumn{2}{c}{'the' 'shadow'} & 'of' & 'them' & \multicolumn{2}{c}{'the' 'pleasant'} \\
sys & $^{0}$\textit{B} & \multicolumn{2}{c}{$^{1}$\textit{CL}} & $^{2}$\textit{FL} & $^{3}$\textit{HM} & \multicolumn{2}{c}{$^{4}$\textit{HNEIM}} \\ 
& 'in' & \multicolumn{2}{c}{'shadow'} & 'of' & 'them' & \multicolumn{2}{c}{'made-pleasant'} \\
\end{tabular}
}
\end{center}
\noindent Pairs of \{$^{1.0}$\textit{H} $^{1.1}$\textit{CL}, $^{1}$\textit{CL}\} ('the shadow') and \{$^{4.0}$\textit{H} $^{4.1}$\textit{NEIM}, $^{4}$\textit{HNEIM}\} ('the pleasant') are word-aligned using the proposed algorithm, and we can obtain 4/5 and 4/7 for precision and recall using the proposed method. 
The CoNLL evaluation script is unable to assess such mismatches because  \textit{the concatenation of tokens in gold file and in system file differ}.\footnote{This is an actual \texttt{UDError} message.} 
Unfortunately, we were unable to find real-world examples of morphological mismatches from GSD treebanks. Since \texttt{stanza} has been trained on UDs, it would produce UD-friendly results without discrepancies. Therefore, we view this area as a potential subject for future investigation. 

Given the absence of consensus evaluation methods for tokenization and sentence boundary detection, as well as the lack of direct approaches to evaluate preprocessing outcomes, a comprehensive comparison with previous work is unfeasible. Instead of utilizing the plain text format, where conventional preprocessing tools are typically applied, we opt to perform comparisons using the CoNLL-U evaluation script in the CoNLL-U format, which offers representational advantages over plain text, such as representing tokenization results like \textit{1 000} as a single token.

We have expanded our alignment-based evaluation approach to include other tasks, such as evaluating constituency parsing results.
The widely used \texttt{evalb} script has traditionally been employed for evaluating the accuracy of constituency parsing results, albeit with the requirement for consistent tokenization and sentence boundaries. 
We align sentences and words when discrepancies arise to to overcome several known issues associated with \texttt{evalb} by utilizing the `jointly preprocessed alignment-based method \citep{jo-etal-2024-novel,park-etal-2024-jp}. 
The proposed approach will also be applicable to various sentence-based evaluation metrics, including POS tagging, machine translation, and grammatical error correction.

\subsection{Conclusion} \label{preprocessing-conclusion}

While most existing tokenization and sentence boundary detection (SBD) implementations are generally considered suitable for direct re-implementation, it is important to note that when they are applied to new use cases, many miscounted true positives are likely to be overlooked. As a result, these inaccuracies remain hidden and not immediately apparent in the intermediate preprocessing results. However, these text segmentation tasks play a fundamental role in sentence processing. Any unnoticed inaccuracies at these early stages can potentially be magnified in downstream NLP tasks, significantly affecting the entire NLP pipeline. This issue of miscounted true positives is a largely unacknowledged aspect of sentence preprocessing.
By introducing sentence and word alignments into the proposed pipeline, we can better identify and reassess such hidden but prevalent inaccuracies in the foundational preprocessing steps. Through the jp-algorithm, we can focus on addressing mismatches that occur during crucial preprocessing procedures and accurately count the true positives during evaluation.\footnote{This chapter is based on "An Untold Story of Preprocessing Task Evaluation: An Alignment-based Joint Evaluation Approach" by Eunkyul Leah Jo, Angela Yoonseo Park, Grace Tianjiao Zhang, Izia Xiaoxiao Wang, Junrui Wang, MingJia Mao, and Jungyeul Park, published in \textit{Proceedings of the 2024 Joint International Conference on Computational Linguistics, Language Resources and Evaluation (LREC-COLING 2024)}, pages 1327–1338, Torino, Italia. ELRA and ICCL \citep{jo-etal-2024-untold-story}.}

\section{\textsc{JP-evalb}} \label{jp-evalb}

\subsection{Introduction}

Evaluation is a systematic method for assessing a design or implementation to measure how well it achieves its goals. In natural language processing (NLP) systems, quality is assessed using evaluation criteria and measures by comparing them to gold standard answer keys. In the context of constituent parsers, we evaluate the fitness of our predicted parse tree against the human-labeled reference parse tree in the test set. For constituent parsing, whether statistical or neural, we rely on the \texttt{EVALB} implementation\footnote{\url{http://nlp.cs.nyu.edu/evalb}. 
There is also an \texttt{EVALB\_SPMRL} implementation, specifically designed for the SPMRL shared task \citep{seddah-EtAl:2013:SPMRL,seddah-kubler-tsarfaty:2014}.
}. 
It uses the PARSEVAL measures \citep{black-etal-1991-procedure} as the standard method for evaluating parser performance. 
A constituent in a hypothesis parse of a sentence is labeled as correct if it matches a constituent in the reference parse with the same non-terminal symbol and span (starting and end indexes). Despite its success in evaluating language technology, \texttt{EVALB} faces an unresolved critical issue in our discipline.
\texttt{EVALB} has constraints, such as requiring the same tokenization results. Its implementation assumes equal-length gold and system files, with one tree per line. Nevertheless, we evaluate parser accuracy using \texttt{EVALB}'s standard F1 metric for constituent parsing.

Furthermore, in today's component-based NLP systems, it is common practice to evaluate parsers individually. This approach helps improve accuracy by preventing errors from propagating through dependent preprocessing steps. 
We propose a new way of constituent parsing evaluation algorithm, which better simulates real-world scenarios and extends beyond controlled and task-specific settings.
Hence, we propose a new way of calculating PARSEVAL measures, which aim to solve some limitations of \texttt{EVALB} for more error-free and accurate evaluation metrics. By rectifying its restrictions, we would be able to present refined precision and recall for the F1 measures in constituent parsing evaluation. 

To emphasize the importance of our new methodology, we will first address the task-specific inherent problems in tokenization and sentence boundary detection before constituent parsing. We will then demonstrate the new implementation of PARSEVAL measures by presenting solutions to each identified mismatch case and their corresponding algorithms. To ensure the reliability and applicability of these algorithms, we will also conduct additional discussion towards the end of the chapter.

\subsection{Known problems}

To illustrate how we present this new approach, consider some known problems of \texttt{EVALB} that dictate why this new solution is needed. Firstly, evaluation cannot be complete if the terminal nodes of the gold and system trees are different, causing a word mismatch error.
An example of this can be found when the gold and system spans differ on the character level with tokens like \textit{This} versus \textit{this}. These tokens are considered identical if we disregard the distinction made by letter case.
Hence, we can resolve this character discrepancy by converting all letters to lowercase. This adjustment allows our evaluation system to treat \textit{This} and \textit{this} as a matching word pair.

Secondly, tokens represented as terminal nodes in gold parse trees can differ from those in parser outputs due to the token and sentence segmentation of the system. During preprocessing, even with the same sentence boundary, tokenization discrepancies may arise when compared to the gold standard tree from the Penn Treebank. This mainly occurs when periods and contractions create ambiguity among words that are abbreviations or acronyms. Such discrepancies can lead to the preprocessing results diverging into several different tokenization schemes. Importantly, \texttt{EVALB} is unable to evaluate constituent parsing when the system's tokenization result differs from the gold standard.

\begin{center}
\begin{tabular}{r rc l}
Example: & gold     & \textit{This ca$\sqcup$n't be right$\sqcup$.} &\\
         & system   & \textit{~this can$\sqcup$not be right$\sqcup$.} & where $\sqcup$ is a token delimiter.\\ 
\end{tabular}
\end{center}

\noindent The discrepancy is evident in such a comparison of \textit{ca$\sqcup$n't} (gold) versus \textit{can$\sqcup$not} (system) for \textit{cannot}. 
In this context, it is readily apparent to human eyes that the gold and system tokens are actually the same. To handle such an instance that \texttt{EVALB} cannot manage, we observe that \textit{ca$\sqcup$n't} and \textit{can$\sqcup$not} are indifferent to each other between all tokens when we create the set of constituents.
This observation plays a pivotal role in shaping our approach to address tokenization challenges, and it is equally significant in resolving issues related to sentence boundaries prior to constituent parsing.

The mission of a sentence boundary detection system is to recognize where each sentence starts and ends. A major hurdle in this task is to detect sentence beginnings and endings given some text that lacks punctuation marks. In the following example, although there is no tokenization discrepancy, a sentence boundary discrepancy exists. In the system, \textit{Click here To view it.} is perceived as two separate sentences: \textit{Click here} and \textit{To view it}. The previous method proposed by \texttt{EVALB} could not assign a score to the unmatched sentences.
However, it is worth noting that there are partial matches between the gold and system trees, even though the current \texttt{EVALB} does not consider them. 

\begin{center}
\begin{tabular}{r rc l}
Example: & gold     & \textit{Click here To view it$\sqcup$.} &\\
         & system   & \textit{Click here $\sqcap$ To view it$\sqcup$.} & where $\sqcap$ is a sentence delimiter.\\ 
\end{tabular}
\end{center}

Consequently, tokens undergoing tokenization and sentences handled through sentence boundary detection share a common quality during evaluation. The gold and system results turn out to be two identical sequences of characters. However, they may still differ in length across tokens and lines due to the various tokenization and sentence boundary detection results.
Therefore, we suggest the next step beyond \texttt{EVALB}, re-indexing system lines through sentence and word alignment. As part of our solution, we propose an evaluation-by-alignment algorithm to avoid mismatches in sentences and words when deriving constituents for eventual evaluation. The algorithms of the new PARSEVAL measures allow us to reassess such edge cases of mismatch.

Finally, the question of how to evaluate constituent parsing results from these end-to-end systems has been a longstanding challenge. Conventionally, \texttt{EVALB} has proven useful in a component-based preprocessing pipeline, with each component evaluated individually under ideal circumstances.
However, conducting end-to-end evaluations with all preprocessing in a single pipeline can offer an alternative perspective in constituent parsing evaluation, and this is the approach adopted for the proposed new PARSEVAL measures.
By addressing the constraints discovered in \texttt{EVALB} that lead to issues in preprocessing, we create an opportunity to compare end-to-end parser results. Even when different preprocessing results are produced due to the use of various models in sentence boundary detection and tokenization, the extension of the evaluation technique with the new way of calculating PARSEVAL measures makes this comparison possible.

\subsection{Implementing new PARSEVAL measures}

\paragraph{Algorithm}

To describe the proposed algorithms, we use the following notations for conciseness and simplicity. $\mathcal{T_{L}}$ and $\mathcal{T_{R}}$ introduce the entire parse trees of gold and system files, respectively. 
$\mathcal{T_{L}}$ is a simplified notation representing $\mathcal{T}_{\mathcal{L}(l)}$, where $l$ is the list of tokens in $\mathcal{L}$. This notation applies in the same manner to $\mathcal{R}$.
$\mathcal{S}_{\mathcal{T}}$ represents a set of constituents of a tree $\mathcal{T}$, and $\mathcal{C}(\mathcal{T})$ is the total number constituents of $\mathcal{T}$.
$\mathcal{C}(\texttt{tp})$ is the number of true positive constituents where $\mathcal{S}_{\mathcal{T_{L}}}\cap\mathcal{S}_{\mathcal{T_{R}}}$, and we count it per aligned sentence. 
The presented Algorithm~\ref{alg:main-algorithm} demonstrates the pseudo-code for the new PARSEVAL measures. 

\begin{algorithm}
\caption{Pseudo-code for new PARSEVAL measures}\label{alg:main-algorithm}
\begin{algorithmic}[1]
\STATE{\textbf{function} {\textsc{PARSEVALmeasures}} ($\mathcal{T_{L}}$ and $\mathcal{T_{R}}$):}
\begin{ALC@g}
\STATE {Extract the list of tokens $\mathcal{L}$ and $\mathcal{R}$ from $\mathcal{T_{L}}$ and $\mathcal{T_{R}}$}
\STATE {$\mathcal{L^{\prime}}$, $\mathcal{R^{\prime}}$ $\gets$ \textsc{SentenceAlignment}($\mathcal{L}$, $\mathcal{R}$) where $\textsc{len}(\mathcal{L^{\prime}}) = \textsc{len}(\mathcal{R^{\prime}})$}
\STATE {Align trees based on $\mathcal{L^{\prime}}$ and $\mathcal{R^{\prime}}$ to obtain $\mathcal{T_{L^{\prime}}}$ and $\mathcal{T_{R^{\prime}}}$}

\WHILE{$\mathcal{T}_{\mathcal{L}^{\prime}}$ and $\mathcal{T}_{\mathcal{R}^{\prime}}$}
\STATE {Extract the list of tokens ${l}$ and ${r}$ from $\mathcal{T}_{\mathcal{L}^{\prime}_{i}}$ and $\mathcal{T}_{\mathcal{R}^{\prime}_{i}}$}
\STATE {$l^{\prime}, r^{\prime} \gets {\textsc{WordAlignment}}(l,r)$}
\STATE {$\mathcal{S}_{\mathcal{T}_{\mathcal{L}}}$ $\gets$ \textsc{GetConstituent}($\mathcal{T}_{\mathcal{L}^{\prime}_{i}(l^{\prime})},0$) where $0< i \leq \textsc{len}(\mathcal{L^{\prime}})$}
\STATE {$\mathcal{S}_{\mathcal{T}_{\mathcal{R}}}$ $\gets$ \textsc{GetConstituent}($\mathcal{T}_{\mathcal{R}^{\prime}_{i}(r^{\prime})},0$) where $0< i \leq \textsc{len}(\mathcal{R^{\prime}})$}
\STATE {$\mathcal{C}(\mathcal{T_{L}}) \gets \mathcal{C}(\mathcal{T_{L}}) +$ \textsc{len}($\mathcal{S}_{\mathcal{T}_{\mathcal{L}}}$)}
\STATE {$\mathcal{C}(\mathcal{T_{R}}) \gets \mathcal{C}(\mathcal{T_{R}}) +$ \textsc{len}($\mathcal{S}_{\mathcal{T}_{\mathcal{R}}}$)}
\WHILE{$\mathcal{S}_{\mathcal{T}_{\mathcal{L}}}$ {and} $\mathcal{S}_{\mathcal{T}_{\mathcal{R}}}$}
\IF{(\textsc{label}, {\textsc{start}$_{\mathcal{L}}$}, {\textsc{end}$_{\mathcal{L}}$},$l^{\prime}_{j}$)
$=$
(\textsc{label}, {\textsc{start}$_{\mathcal{R}}$}, {\textsc{end}$_{\mathcal{R}}$},$r^{\prime}_{j}$)}
\STATE{$\mathcal{C}(\texttt{tp}) \gets \mathcal{C}(\texttt{tp}) + 1 $  
}
\ENDIF
\ENDWHILE
\ENDWHILE
\RETURN {$\mathcal{C}(\mathcal{T_{L}})$, $\mathcal{C}(\mathcal{T_{R}})$, and $\mathcal{C}(\texttt{tp})$} 
\end{ALC@g}
\end{algorithmic} 
\end{algorithm}

\begin{algorithm}
\caption{Pseudo-code for sentence alignment}\label{sentence-algorithm}
\begin{algorithmic}[1]
\STATE{\textbf{function} \textsc{SentenceAlignment} ($\mathcal{L}$, $\mathcal{R}$):}
\begin{ALC@g}
\WHILE{$\mathcal{L}$ and $\mathcal{R}$}
\IF{ $(\mathcal{L}_{i({\not\sqcup})}$ $=$ $\mathcal{R}_{j({\not\sqcup})})$ \COMMENT{\textsc{case 1 $_{(i,j)}$}}\\ 
~~~~ $\lor$ $( \mathcal{L}_{i({\not\sqcup})}$ $\simeq$ $\mathcal{R}_{j({\not\sqcup})} \land 
(
\mathcal{L}_{i+1({\not\sqcup})}$ $=$ $\mathcal{R}_{j+1({\not\sqcup})}
\lor
\mathcal{L}_{i+1({\not\sqcup})}$ $\simeq$ $\mathcal{R}_{j+1({\not\sqcup})} 
) )$ \COMMENT{\textsc{case 2 $_{(i,j)}$}}\\ 
} 
\STATE{$\mathcal{L^{\prime}}$, $\mathcal{R^{\prime}}$ $\gets$ $\mathcal{L^{\prime}}+\mathcal{L}_i, \mathcal{R^{\prime}}+\mathcal{R}_j$ where $0<i\leq {\textsc{len}}(\mathcal{L})$, $0<j\leq {\textsc{len}}(\mathcal{R})$}
\ELSE
\WHILE{$\neg$(\textsc{case 1 $_{(i+1,j+1)}$} $\lor$ \textsc{case 2 $_{(i+1,j+1)}$})}
\IF{$\textsc{len}(\mathcal{L}_{i}) < \textsc{len}(\mathcal{R}_{j})$}
\STATE{${L}^{\prime}$ $\gets$ ${L}^{\prime}+\mathcal{L}_{i}$} 
\STATE{$i \gets i+1 $}
\ELSE
\STATE{${R}^{\prime}$ $\gets$ ${R}^{\prime}+\mathcal{R}_{j}$}
\STATE{$j \gets j+1 $}
\ENDIF

\ENDWHILE
\STATE{
$\mathcal{L^{\prime}}$, $\mathcal{R^{\prime}}$ 
$\gets$ 
$\mathcal{L^{\prime}}+L^{\prime}, \mathcal{R^{\prime}}+R^{\prime}$
}
\ENDIF
\ENDWHILE
\RETURN {$\mathcal{L^{\prime}}$, $\mathcal{R^{\prime}}$}
\end{ALC@g}
\end{algorithmic} 
\end{algorithm}

\begin{algorithm}
\caption{Pseudo-code for word alignment} \label{word-algorithm}
\begin{algorithmic}[1]
\STATE{\textbf{function} \textsc{WordAlignment} ($l$, $r$):}
\begin{ALC@g}
\WHILE{$l$ and $r$}
\IF{ ($l_{i}$ $=$ $r_{j}$) \COMMENT{\textsc{case 1 $_{(i,j)}$}}\\ 
~~~~ $\lor$
(($l_{i}$ $\neq$ $r_{j}$) $\land$ ($l_{i+1}$ $=$ $r_{j+1}$))
\COMMENT{\textsc{case 2 $_{(i,j)}$}}\\
}
\STATE{$l^{\prime}$, $r^{\prime}$ $\gets$ $l_{i}, r_{j}$ where $0<i\leq \textsc{len}(l)$, $0<j\leq {\textsc{len}}(r)$}
\ELSE
\WHILE{$\neg$(\textsc{case 1 $_{(i+1,j+1)}$} $\lor$ \textsc{case 2 $_{(i+1,j+1)}$})}
\IF{$(\textsc{len}(l) - \textsc{len}(l_{0},l_{i}))  > (\textsc{len}(r) - \textsc{len}(r_{0},r_{j}))$}
\STATE{${ll}$ $\gets$ ${ll}+l_{i}$} 
\STATE{$i \gets i+1 $}
\ELSE
\STATE{${rr}$ $\gets$ ${rr}+{r}_{j}$}
\STATE{$j \gets j+1 $}
\ENDIF
\ENDWHILE
\STATE{$l^{\prime}$, $r^{\prime}$ $\gets$ $ll, rr$}
\ENDIF
\ENDWHILE
\RETURN {$l^{\prime}$, $r^{\prime}$}
\end{ALC@g}
\end{algorithmic} 
\end{algorithm}

In the first stage, we extract leaves $\mathcal{L}$ and $\mathcal{R}$ from the parse trees and align sentences to obtain $\mathcal{L^{\prime}}$ and $\mathcal{R^{\prime}}$ using Algorithm~\ref{sentence-algorithm}. 
While the necessity of sentence alignment is rooted in a common phenomenon in cross-language tasks {such as machine translation}, the intralingual alignment between gold and system sentences does not share the same necessity {because $\mathcal{L}$ and $\mathcal{R}$ are} identical sentences that only differ in sentence boundaries and tokenization results.
A notation ${\not\sqcup}$ is introduced to represent spaces that are removed {during sentence alignment when comparing $\mathcal{L}{i}$ and $\mathcal{R}_{j}$, irrespective of their tokenization results.}
{If there is a mismatch due to differences in sentence boundaries, the algorithm accumulates the sentences until the next pair of sentences represented as \textsc{case $n$ ${(i+1,j+1)}$}, is matched.}
In the next stage {of Algorithm~\ref{alg:main-algorithm},} 
we align trees based on $\mathcal{L}^{\prime}$ and $\mathcal{R}^{\prime}$ to obtain $\mathcal{T}_{\mathcal{L}^{\prime}}$ and $\mathcal{T}_{\mathcal{R}^{\prime}}$. 
By iterating through $\mathcal{T}_{\mathcal{L}^{\prime}}$ and $\mathcal{T}_{\mathcal{R}^{\prime}}$, we conduct word alignment and compare pairs of sets of constituents for each corresponding pair of $\mathcal{T}_{\mathcal{L}^{\prime}_{i}}$ and {$\mathcal{T}_{\mathcal{R}^{\prime}_{j}}$}.
The word alignment in Algorithm~\ref{word-algorithm} follows a logic similar to sentence alignment, wherein words are accumulated in $ll$ and $rr$ if the pairs of $l_i$ and $r_j$ do not match due to tokenization mismatches.
{Finally,} we extract a set of constituents using Algorithm~\ref{get-constituents}, a straightforward procedure for obtaining constituents from a given tree, which includes the label name, start index, end index, and a list of tokens.
{The current proposed method utilizes simple pattern matching for sentence and word alignment, operating under the assumption that the gold and system sentences are the same, with minimal potential for morphological mismatches.
This differs from sentence and word alignment in machine translation.
MT usually relies on recursive editing and EM algorithms due to the inherent difference between source and target languages.}\label{rc2}

\begin{algorithm}
\caption{Pseudo-code for getting constituents} 
\label{get-constituents}
\begin{algorithmic}[1]
\STATE{\textbf{function} \textsc{GetConstituent} ($\mathcal{T}$, start):}
\begin{ALC@g}
\IF{\textsc{height}($\mathcal{T}$)>2}
\STATE {\textsc{end} $\gets$ \textsc{start} $+$ \textsc{len}(\textsc{leaves}($\mathcal{T}$))}
\STATE {$\mathcal{S}_{\mathcal{T}} \gets \mathcal{S}_{\mathcal{T}}~ +$ (\textsc{label}($\mathcal{T}$), \textsc{start}, \textsc{end}, \textsc{leaves}($\mathcal{T}$))}
\WHILE{$T$}
\STATE{\textsc{GetConstituent}($\mathcal{T}_{i}$, {start}) where $\mathcal{T}_{i}$ is a child of $\mathcal{T}$}
\STATE{\textsc{start} $\gets$ \textsc{len}(\textsc{leaves}($\mathcal{T}_{i}$))}
\ENDWHILE
\ENDIF
\RETURN {$\mathcal{S}_{\mathcal{T}}$}
\end{ALC@g}
\end{algorithmic} 
\end{algorithm} 


\paragraph{Word mismatch}
We have observed that the expression of contractions varies significantly, resulting in inherent challenges related to word mismatches.
As the number of contractions and symbols to be converted in a language is finite, we composed an exception list for our system to capture such cases for each language to facilitate the word alignment process between gold and system sentences.
In the following example, we achieve perfect precision and recall of 5/5 for both because their constituent trees are exactly matched, regardless of any mismatched words.

\begin{center}
\resizebox{\textwidth}{!}{\footnotesize
\begin{tabular}{p{1cm}p{1cm}p{1cm}p{1cm}p{1cm}p{1cm} cc | cc p{1cm}p{1cm}p{1cm}p{1cm}p{1cm}p{1cm}}
&& \multicolumn{6}{c}{(gold)} & \multicolumn{6}{c}{(system)} &&\\

&&\multicolumn{1}{c}{\cellcolor{blue!15}S$_{(0,4)}$}&  &   &  \multicolumn{1}{c}{\cellcolor{yellow!15}NP$_{(0,1)}$} &DT & \textit{$^{0}$This}  & \textit{$^{0}$this} & DT & \multicolumn{1}{c}{\cellcolor{yellow!15}NP$_{(0,1)}$}  & & & \multicolumn{1}{c}{\cellcolor{blue!15}S$_{(0,4)}$}&&\\

&&\cellcolor{blue!15} &\multicolumn{1}{c}{\cellcolor{red!15}VP$_{(1,4)}$}&   &     &MD & \textit{$^{1.0}$ca}    & \textit{$^{1.0}$can}  & MD & & & \multicolumn{1}{c}{\cellcolor{red!15}VP$_{(1,4)}$} & \cellcolor{blue!15}&&\\

&&\cellcolor{blue!15} &\cellcolor{red!15}  &   &     &RB & \textit{$^{1.1}$n't}   & \textit{$^{1.1}$not}  & RB & & & \cellcolor{red!15}& \cellcolor{blue!15}&&\\

&&\cellcolor{blue!15} &\cellcolor{red!15}  & \multicolumn{1}{c}{\cellcolor{green!15}VP$_{(2,4)}$}&     &VB & \textit{$^{2}$be}    & \textit{$^{2}$be}   & VB & & \multicolumn{1}{c}{\cellcolor{green!15}VP$_{(2,4)}$} & \cellcolor{red!15}& \cellcolor{blue!15}&&\\

&&\cellcolor{blue!15} &\cellcolor{red!15}  & \cellcolor{green!15}  & \multicolumn{1}{c}{\cellcolor{yellow!15}AdjP$_{(3,4)}$}&JJ & \textit{$^{3}$right} & \textit{$^{3}$right}& JJ & \multicolumn{1}{c}{\cellcolor{yellow!15}AdjP$_{(3,4)}$} & \cellcolor{green!15} & \cellcolor{red!15} & \cellcolor{blue!15}&&\\
\end{tabular}
}
\end{center}

\noindent If the word mismatch example is not in the exception list, we perform the word alignment. We can still achieve perfect precision and recall (5/5 for both) without the word mismatch exception list because their constituent trees can be exactly matched based on the word-alignment of \{$^{1.0}$\textit{ca}  $^{1.1}$\textit{n't}\} and \{$^{1.0}$\textit{can}  $^{1.1}$\textit{not}\}.

\begin{center}
\begin{tabular}{r c cc c c }
gold &  $^{0}$\textit{This} & $^{1.0}$\textit{ca} & $^{1.1}$\textit{n't} & $^{2}$\textit{be} & $^{3}$\textit{right}\\
system & $^{0}$\textit{this} & $^{1.0}$\textit{can} & $^{1.1}$\textit{not} & $^{2}$\textit{be} & $^{3}$\textit{right}\\ 
\end{tabular}
\end{center}

The effectiveness of the word alignment approach remains intact even for morphological mismatches where "morphological segmentation is not the inverse of concatenation" \citep{tsarfaty-etal-2012-joint}, such as in morphologically rich languages.  
For example, we trace back to the sentence in Hebrew described in \citet{tsarfaty-etal-2012-joint} as a word mismatch example caused by morphological analyses:
\begin{center}
\begin{tabular}{r ccc cccc}
gold & $^{0}$\textit{B} & $^{1.0}$\textit{H} & $^{1.1}$\textit{CL} & $^{2}$\textit{FL} & $^{3}$\textit{HM} & $^{4.0}$\textit{H} & $^{4.1}$\textit{NEIM} \\
& 'in' & 'the' & 'shadow' & 'of' & 'them' & 'the' & 'pleasant' \\
system & $^{0}$\textit{B} & \multicolumn{2}{c}{$^{1}$\textit{CL}} & $^{2}$\textit{FL} & $^{3}$\textit{HM} & \multicolumn{2}{c}{$^{4}$\textit{HNEIM}} \\ 
& 'in' & \multicolumn{2}{c}{'shadow'} & 'of' & 'them' & \multicolumn{2}{c}{'made-pleasant'} \\
\end{tabular}
\end{center}
Pairs of \{$^{1.0}$\textit{H} $^{1.1}$\textit{CL}, $^{1}$\textit{CL}\} ('the shadow') and \{$^{4.0}$\textit{H} $^{4.1}$\textit{NEIM}, $^{4}$\textit{HNEIM}\} ('the pleasant') are word-aligned using the proposed algorithm, resulting in a precision of 4/4 and recall of 4/6. 

\begin{center}
\resizebox{\textwidth}{!}{\footnotesize
\begin{tabular}{p{1cm} p{1cm} p{1cm}p{1cm}p{1cm} cc | cc p{1cm}p{1cm}p{1cm}p{1cm} p{1cm}} 

{}& \multicolumn{6}{c}{(gold)} & \multicolumn{6}{c}{(system)} &{}\\

{}& \multicolumn{1}{c}{\cellcolor{blue!15}PP$_{(0,5)}$} & {} & {} & {} & 
'in' & \textit{$^{0}$B} &  \textit{$^{0}$B} & 'in' &  
{} & {} & {} & \multicolumn{1}{c}{\cellcolor{blue!15}PP$_{(0,5)}$} &{}\\

&\multicolumn{1}{c}{\cellcolor{blue!15}} & \multicolumn{1}{c}{\cellcolor{red!15}NP$_{(1,5)}$} & \multicolumn{1}{c}{\cellcolor{green!15}NP$_{(1,4)}$} & {} & 
'the' & \textit{$^{1.0}$H} &  
\textit{} & {} &  
{} & {} & {} & \multicolumn{1}{c}{\cellcolor{blue!15}}& \\

&\multicolumn{1}{c}{\cellcolor{blue!15}} & \multicolumn{1}{c}{\cellcolor{red!15}} & \multicolumn{1}{c}{\cellcolor{green!15}} & {} & 
'shadow' & \textit{$^{1.1}$CL} &  
\textit{$^{1}$CL} & {'shadow'} &  
{} & \multicolumn{1}{c}{\cellcolor{green!15}NP$_{(1,4)}$} & \multicolumn{1}{c}{\cellcolor{red!15}NP$_{(1,5)}$}  & \multicolumn{1}{c}{\cellcolor{blue!15}}&\\

&\multicolumn{1}{c}{\cellcolor{blue!15}} & \multicolumn{1}{c}{\cellcolor{red!15}} & \multicolumn{1}{c}{\cellcolor{green!15}} & \multicolumn{1}{c}{\cellcolor{red!15}PP$_{(2,4)}$} & 
'of' & \textit{$^{2}$FL} &  
\textit{$^{2}$FL} & {'of'} &  
\multicolumn{1}{c}{\cellcolor{red!15}PP$_{(2,4)}$} & \multicolumn{1}{c}{\cellcolor{green!15}} & \multicolumn{1}{c}{\cellcolor{red!15}}  & \multicolumn{1}{c}{\cellcolor{blue!15}}&\\

&\multicolumn{1}{c}{\cellcolor{blue!15}} & \multicolumn{1}{c}{\cellcolor{red!15}} & \multicolumn{1}{c}{\cellcolor{green!15}} & \multicolumn{1}{c}{\cellcolor{red!15}} & 
'them' & \textit{$^{3}$HM} &  
\textit{$^{3}$HM} & {'them'} &  
\multicolumn{1}{c}{\cellcolor{red!15}} & \multicolumn{1}{c}{\cellcolor{green!15}} & \multicolumn{1}{c}{\cellcolor{red!15}}  & \multicolumn{1}{c}{\cellcolor{blue!15}}&\\

&\multicolumn{1}{c}{\cellcolor{blue!15}} & \multicolumn{1}{c}{\cellcolor{red!15}} & {} & \multicolumn{1}{c}{\cellcolor{blue!15}AdjP$_{(4,5)}$} & 
'the' & \textit{$^{4.0}$H} &  
\textit{} & {} &  {} & 
{} & \multicolumn{1}{c}{\cellcolor{red!15}}  & \multicolumn{1}{c}{\cellcolor{blue!15}}&\\

&\multicolumn{1}{c}{\cellcolor{blue!15}} & \multicolumn{1}{c}{\cellcolor{red!15}} & {} & \multicolumn{1}{c}{\cellcolor{blue!15}} & 
'pleasant' & \textit{$^{4.1}$NEIM} &  
\textit{$^{4}$HNEIM} & {'made-pleasant'} &  {} & 
{} & \multicolumn{1}{c}{\cellcolor{red!15}}  & \multicolumn{1}{c}{\cellcolor{blue!15}}&\\

\end{tabular}
}
\end{center}

\paragraph{Sentence mismatch}
When there are sentence mismatches, they would be aligned and merged as a single tree using a dummy root node: for example, \textsc{@s} which can be ignored during evaluation.
In the following example, we obtain precision of 5/8 and recall of 5/7.

\begin{center}
\resizebox{\textwidth}{!}{\footnotesize
\begin{tabular}{p{1cm}p{1cm}p{1cm} p{1cm}p{1cm}p{1cm} cc 
| cc p{1cm}p{1cm}p{1cm} p{1cm}p{1cm}p{1cm}  }
\multicolumn{8}{c}{(gold)} & \multicolumn{8}{c}{(system, merged after alignment)} \\

\multicolumn{1}{c}{\cellcolor{blue!15}S$_{(0,6)}$}& \multicolumn{1}{c}{\cellcolor{red!15}S$_{(0,5)}$} & {} &
{} & {} & {} & 
VB & \textit{$^{0}$Click} &  \textit{$^{0}$Click} &  VB & 
{} & \multicolumn{1}{c}{\cellcolor{blue!15}VP$_{(0,2)}$} & \multicolumn{1}{c}{\cellcolor{red!15}S$_{(0,2)}$} & {} & {} & \multicolumn{1}{c}{\cellcolor{blue!15}@S$_{(0,6)}$}\\

\cellcolor{blue!15} & \cellcolor{red!15} & {} & 
{} & {} & \multicolumn{1}{c}{\cellcolor{green!15}AdvP$_{(1,2)}$} & 
RB & \textit{$^{1}$here} &  \textit{$^{1}$here} &  RB & 
\multicolumn{1}{c}{\cellcolor{green!15}AdvP$_{(1,2)}$} & \cellcolor{blue!15} & \cellcolor{red!15} &  & & \cellcolor{blue!15} \\

\cellcolor{blue!15} & \cellcolor{red!15} & \multicolumn{1}{c}{\cellcolor{green!15}S$_{(2,5)}$} & 
\multicolumn{1}{c}{\cellcolor{yellow!15}VP$_{(2,5)}$} & {} & {} & 
TO & \textit{$^{2}$To} &  \textit{$^{2}$To} &  TO & 
&  & \multicolumn{1}{c}{\cellcolor{yellow!15}VP$_{(2,5)}$} & \multicolumn{1}{c}{\cellcolor{green!15}S$_{(2,5)}$} & \multicolumn{1}{c}{\cellcolor{red!15}S$_{(2,6)}$}& \cellcolor{blue!15}\\

\cellcolor{blue!15} & \cellcolor{red!15} & {\cellcolor{green!15}} & 
{\cellcolor{yellow!15}} & \multicolumn{1}{c}{\cellcolor{blue!15}VP$_{(3,5)}$} & {} & 
VB & \textit{$^{3}$view} &  \textit{$^{3}$view} &  VB & 
& \multicolumn{1}{c}{\cellcolor{blue!15}VP$_{(3,5)}$}
& \multicolumn{1}{c}{\cellcolor{yellow!15}} & \multicolumn{1}{c}{\cellcolor{green!15}} & \multicolumn{1}{c}{\cellcolor{red!15}}& \cellcolor{blue!15}\\

\cellcolor{blue!15} & \cellcolor{red!15} & {\cellcolor{green!15}} & 
{\cellcolor{yellow!15}} & \cellcolor{blue!15} & \multicolumn{1}{c}{\cellcolor{green!15}NP$_{(4,5)}$} & 
PRP & \textit{$^{4}$it} &  \textit{$^{4}$it} &  PRP & \multicolumn{1}{c}{\cellcolor{green!15}NP$_{(4,5)}$}
& {\cellcolor{blue!15}}
& {\cellcolor{yellow!15}} & {\cellcolor{green!15}} & {\cellcolor{red!15}}& \cellcolor{blue!15}\\

\cellcolor{blue!15} & & & & &  & .  & $^{5}$. & $^{5}$. & . & 
& & & & \cellcolor{red!15}& \cellcolor{blue!15} \\
\end{tabular}
}
\end{center}

\paragraph{Assumptions}
To address morphological analysis discrepancies in the parse tree during evaluation, we establish the following two assumptions: (i)
The entire tree constituent can be considered a true positive, even if the morphological segmentation or analysis differs from the gold analysis, as long as the two sentences (gold and system) are aligned and their {root} labels are the same. \label{rc4}
(ii) The subtree constituent can be considered a true positive if lexical items align in word alignment, and their phrase labels are the same.

\subsection{Case studies}

It's important to note that the original \texttt{evalb} excludes problematic symbols and punctuation marks in the tree structure. Our results include all tokens in the given sentence, and bracket numbers reflect the actual constituents in the system and gold parse trees. 
Accuracy in the last column of the result is determined by comparing the correct number of POS-tagged words to the gold sentence including punctuation marks, differing from the original \texttt{evalb} which doesn't consider word counts or correct POS tags. 
Figure~\ref{comparison} visually depicts the difference in constituent lists between \texttt{jp-evalb} and \texttt{evalb}. 
The original \texttt{evalb} excludes punctuation marks from its consideration of constituents, resulting in our representation of word index numbers in red for \texttt{evalb}. Consequently, \texttt{evalb} displays constituents without punctuation marks and calculates POS tagging accuracy based on six word tokens. On the other hand, \texttt{jp-evalb} includes punctuation marks in constituents and evaluates POS tagging accuracy using eight tokens, which includes two punctuation marks in the sentence. 
We note that the inclusion of punctuation marks in the constituents does not affect the total count, as punctuation marks do not constitute a constituent by themselves.

\begin{figure}
\centering
\begin{subfigure}[b]{\textwidth}    
\footnotesize{
\centering
\synttree
[TOP [S [INTJ [RB [$^{0,{\color{red}0}}$No] ]] [, [$^{1}$,]] [NP [PRP [$^{2,{\color{red}1}}$it]]] 
[VP [VBD [$^{3,{\color{red}2}}$was]] [RB [$^{4,{\color{red}3}}$n't]] [NP [NNP [$^{5,{\color{red}4}}$Black]] [NNP [$^{6,{\color{red}5}}$Monday]]]] [$\cdot$ [$^{7}$$\cdot$] ]]]
}
\caption{Example of the parse tree}\label{parse-tree}
\end{subfigure}\hfill

\begin{subfigure}[b]{\textwidth}

\footnotesize{
\begin{tabular}{l}
~\\
\texttt{ ('S', 0, 8, "No {,} it was n't Black Monday {.}")}\\
\texttt{ ('INTJ', 0, 1, 'No') }\\
\texttt{ ('NP', 2, 3, 'it') }\\
\texttt{ ('VP', 3, 7, "was n't Black Monday") }\\
\texttt{ ('NP', 5, 7, 'Black Monday') }\\
\end{tabular}
}

\caption{List of constituents by \texttt{jp-evalb}}\label{constituents-jp-evalb}
\end{subfigure}\hfill

\begin{subfigure}[b]{\textwidth}
\footnotesize{
\begin{tabular}{l}
~\\
\texttt{('S', {\color{red}0}, {\color{red}6}, "No it was n't Black Monday") }\\
\texttt{('INTJ', {\color{red}0}, {\color{red}1}, "No") }\\
\texttt{('NP', {\color{red}1}, {\color{red}2}, "it")}\\
\texttt{('VP', {\color{red}2}, {\color{red}6}, "was n't Black Monday") }\\
\texttt{('NP', {\color{red}4}, {\color{red}6}, "Black Monday") }\\
\end{tabular}
}

\caption{List of constituents by \texttt{evalb}}\label{constituents-evalb}
\end{subfigure}

\caption{Difference between \texttt{jp-evalb} and \texttt{evalb}}\label{comparison}
\end{figure}

\paragraph{Section 23 of the English Penn treebank}

Under identical conditions where sentences and words match, the proposed method requires around 4.5 seconds for evaluating the section 23 of the Penn Treebank. 
On the same machine, \texttt{evalb} completes the task less than 0.1 seconds. 
We do not claim that our proposed implementation is fast or faster than \texttt{evalb}, recognizing the well-established differences in performance between compiled languages like C, which \texttt{evalb} used, and interpreted languages such as Python, which our current implementation uses. 
Our proposed method also introduces additional runtime for sentence and word alignment, a process not performed by \texttt{evalb}.
We present excerpts from three result files generated by \texttt{evalb} and our proposed method in Figure~\ref{ptb-section23}. The parsed results were obtained using the PCFG-LA Berkeley parser \citep{petrov-klein:2007:main}. 
It's worth noting that there may be slight variations between the two sets of results because \texttt{evalb} excludes constituents with specific symbols and punctuation marks during evaluation. 
However, as we mentioned earlier, \texttt{jp-evalb} can reproduce the exact same results as \texttt{evalb} for a legacy reason.

\begin{figure}[!ht]
  \centering

\begin{subfigure}[b]{\textwidth}
\centering
\footnotesize{
\texttt{\begin{tabular}{rrr rrr rrr rrr}
\multicolumn{12}{c}{}\\
\multicolumn{2}{c}{Sent}& &&& Mt &  \multicolumn{2}{c}{Br} & Cr&& Co& Tag\\
ID&  L&  St& Re&  Pr&  Br& gd& te& Br& Wd&  Tg&  Acc \\
\hdashline
1& 8& 0&  100.00& 100.00&  5&5& 5&0&8&  7& 87.50\\
2&40& 0&70.97&  73.33& 22&  31&30&7&  40& 40&100.00\\
3&31& 0&95.24&  95.24& 20&  21&21&0&  31& 31&100.00\\
4&35& 0&90.48&  86.36& 19&  21&22&2&  35& 35&100.00\\
5&26& 0&86.96&  86.96& 20&  23&23&2&  26& 25& 96.15\\
\multicolumn{12}{l}{.....}
\end{tabular}
}}
\caption{Example of \texttt{jp-evalb} results considering punctuation marks during evaluation}
\label{jp-evalb-results}
\end{subfigure}

\begin{subfigure}[b]{\textwidth}
\centering
\footnotesize{
\texttt{\begin{tabular}{rrr rrr rrr rrr}
\multicolumn{2}{c}{Sent}& &&& Mt &  \multicolumn{2}{c}{Br} & Cr&& Co& Tag\\
ID&  L&  St& Re&  Pr&  Br& gd& te& Br& Wd&  Tg&  Acc \\
\hdashline
1& 8& 0&  100.00& 100.00&   5& 5& 5&0&6&  5& 83.33\\
2&40& 0&70.97&  73.33& 22&  31&30&7&  37& 37&100.00\\
3&31& 0&95.24&  95.24& 20&  21&21&0&  26& 26&100.00\\
4&35& 0&90.48&  86.36& 19&  21&22&2&  32& 32&100.00\\
5&26& 0&86.96&  86.96& 20&  23&23&2&  24& 23& 95.83\\
\multicolumn{12}{l}{.....}
\end{tabular}
}}
\caption{Example of \texttt{jp-evalb} results with the legacy option, which produces the exact same results as \texttt{evalb}}
 \label{jp-evalb-legacy-results}
\end{subfigure}

\begin{subfigure}[b]{\textwidth}
\centering
\footnotesize{
\texttt{\begin{tabular}{rrr rrr rrr rrr}
\multicolumn{2}{c}{Sent}& &&& Mt &  \multicolumn{2}{c}{Br} & Cr&& Co& Tag\\
ID&  L&  St& Re&  Pr&  Br& gd& te& Br& Wd&  Tg&  Acc \\
\hdashline
1& 8& 0&  100.00& 100.00&   5& 5& 5&0&6&  5& 83.33\\
2&40& 0&70.97&  73.33& 22&  31&30&7&  37& 37&100.00\\
3&31& 0&95.24&  95.24& 20&  21&21&0&  26& 26&100.00\\
4&35& 0&90.48&  86.36& 19&  21&22&2&  32& 32&100.00\\
5&26& 0&86.96&  86.96& 20&  23&23&2&  24& 23& 95.83\\
\multicolumn{12}{l}{.....}
\end{tabular}
}}
 \caption{Example of the original \texttt{evalb} results}
 \label{evalb-results}
\end{subfigure}

\caption{Examples of evaluation results on Section 23 of the English Penn treebank}
\label{ptb-section23}
\end{figure}

\paragraph{Bug cases identified by \texttt{evalb}}

We evaluate bug cases identified by \texttt{evalb}. Figure~\ref{evalb-bug} displays all five identified bug cases, showcasing successful evaluation without any failures. 
In three instances (sentences 1, 2, and 5), a few symbols are treated as words during POS tagging. This leads to discrepancies in sentence length because \texttt{evalb} discards symbols in the gold parse tree during evaluation.
Our proposed solution involves not disregarding any problematic labels and including symbols as words during evaluation. 
This approach implies that POS tagging results are based on the entire token numbers. It is noteworthy that \texttt{evalb}'s POS tagging results are rooted in the number of words, excluding symbols. 
The two remaining cases (sentences 3 and 4) involve actual word mismatches where trace symbols (*-\textit{num}) are inserted into the sentences. 
Naturally, \texttt{evalb} cannot handle these cases due to word mismatches. However, as we explained, our proposed algorithm addresses this issue by performing word alignment after sentence alignment.

\begin{figure}[!ht]
\centering
\footnotesize{
\texttt{\begin{tabular}{rrr rrr rrr rrr}
\multicolumn{12}{c}{}\\
\multicolumn{2}{c}{Sent}& &&& Mt &  \multicolumn{2}{c}{Br} & Cr&& Co& Tag\\
ID&  L&  St& Re&  Pr&  Br& gd& te& Br& Wd&  Tg&  Acc \\
\hdashline
1&   37&    0&   77.27&  62.96&    17&     22&   27&      5&     37&    30 &   81.08 \\
2&   21&    0&   69.23&  60.00&     9&     13&   15&      2&     21&    17&    80.95 \\
   3&   47&    0&   77.78&  80.00&    28&     36&   35&      4&     48&    43&    89.58\\
   4&   26&    0&   33.33&  35.29&     6   &  18&   17&      8&     27&    19&    70.37\\
   5&   44&    0&   42.31&  32.35&    11&     26&   34&     17&     44&    33&    75.00 \\
\end{tabular}
}}
\caption{Evaluation results of bug cases}
\label{evalb-bug}
\end{figure}

\paragraph{Korean end-to-end parsing evaluation}
We conduct a comprehensive parsing evaluation for Korean, using system-segmented sequences as input for constituency parsing. These sequences may deviate from the corresponding gold standard sentences and tokens. 
We utilized the following resources for our parsing evaluation to simulate the end-to-end process:
(i) A set of 148 test sentences with 4538 tokens (morphemes)  from \texttt{BGAA0001} of the Korean Sejong treebank, as detailed in \citet{kim-park:2022}. In the present experiment, all sentences have been merged into a single text block.
(ii) POS tagging performed by \texttt{sjmorph.model} \citep{park-tyers:2019:LAW} for morpheme segmentation.\footnote{\url{https://github.com/jungyeul/sjmorph}}
The model's pipeline includes sentence boundary detection and tokenization through morphological analysis, generating an input format for the parser. 
(iii) A Berkeley parser model for Korean trained on the Korean Sejong treebank \citep{park-hong-cha:2016:PACLIC}.\footnote{\url{https://zenodo.org/records/3995084}}.
Figure~\ref{korean-evalb} presents the showcase results of end-to-end Korean constituency parsing. 
Given our sentence boundary detection and tokenization processes, there is a possibility of encountering sentence and word mismatches during constituency parsing evaluation. 
The system results show 123 sentences and 4367 morphemes because differences in sentence boundaries and tokenization results. 
During the evaluation, \texttt{jp-evalb} successfully aligns even in the presence of sentence and word mismatches, and subsequently, the results of constituency parsing are assessed.

\begin{figure}[!ht]
\centering
\footnotesize{
\texttt{\begin{tabular}{rrr rrr rrr rrr}
\multicolumn{12}{c}{}\\
\multicolumn{2}{c}{Sent}& &&& Mt &  \multicolumn{2}{c}{Br} & Cr&& Co& Tag\\
ID&  L&  St& Re&  Pr&  Br& gd& te& Br& Wd&  Tg&  Acc \\
\hdashline
1&28& 0&85.71 & 85.71& 18&  21&21&3&  29& 26& 89.66\\
2&27& 0&91.30 & 84.00& 21&  23&25&2&  28& 25& 89.29\\
3&33& 0&88.00 & 88.00& 22&  25&25&3&  35& 31& 88.57\\
4&43& 0&72.73 & 72.73& 24&  33&33&7&  43& 40& 93.02\\
5&18& 0&69.57 & 84.21& 16&  23&19&2&  19& 12& 63.16\\
\multicolumn{12}{l}{.....}
\end{tabular}
}}
\caption{Evaluation results of the end-to-end Korean constituency parsing}
\label{korean-evalb}
\end{figure}

\subsection{Discussion}

\paragraph{Complexity}
The proposed algorithm has a linear time complexity. Sentence and word alignments require $O(I+J)$, where $I$ and $J$ represent the lengths of the gold and system sentences or words. 
The process for constituent tree matches uses tree traversal algorithm which requires $O(N+E)$ where $N$ is a number of nodes and $E$ is for branches. 
We retain the same time complexity of the original \texttt{EVALB} by adding alignment-based preprocessing for mismatches of sentences and words.

\paragraph{Comparison}
Table~\ref{metric-comparison} compares previous parsing evaluation metrics with the proposed algorithm. \texttt{tedeval} \citep{tsarfaty-etal-2012-joint} is based on the tree edit distance of \citet{bille-2005}, and numbers of nonterminal nodes in system and gold trees. A similar idea on the tree edit distance has proposed for classifying constituent parsing errors based on subtree movement, node creation, and node deletion \citep{kummerfeld-EtAl:2012:EMNLP-CoNLL}.
\texttt{conllu\_eval} for dependency parsing evaluation within Universal Dependencies \citep{nivre-etal-2020-universal} views tokens and sentences as spans. If there is a mismatch of positions of spans between the system and the gold file on a character level, whichever file has a smaller start value will skip to the next token until there is no start value mismatch. Evaluating sentence boundaries also follows similar processes as tokens. The start and end values of {the sentence span} are compared between the system and the gold file. When they match, it increases the count of correctly matched sentences.
\texttt{sparseval} \citep{roark-etal-2006-sparseval} uses a head percolation table \citep{collins-1999-head} to identify head-child relations between two terminal nodes from constituent parsing trees, and calculate the dependency score.
We also add an \texttt{aligning trees} method \citep{calder-1997-aligning} in our comparison, which performs an alignment of the tree structures from two different treebanks for the same sentence, both of which employ distinct POS labels.

\begin{table}[!th]
\centering
\resizebox{\textwidth}{!}{
\footnotesize{
\begin{tabular} {r l l} \hline
     & evaluation approach & addressing mismatches\\\hline
\texttt{tedeval}&  tree-edit distance based on constituent trees & words\\
\texttt{conllu\_eval} & dependency scoring & words and sentences\\
\texttt{sparseval} & dependency scoring & words and sentences\\ 
\texttt{aligning trees} &    constituent tree matches  & words \\  
\hdashline
\texttt{EVALB} & constituent tree matches & not applicable\\
{proposed method} & constituent tree matches & words and sentences  \\ \hline 
\end{tabular}
}}
\caption{Comparison to previous parsing evaluation metrics}\label{metric-comparison}
\end{table}

\paragraph{A note on constituent parsing}
Syntactic analysis in the current field of language technology has been predominantly reliant on dependencies. 
Semantic parsing in its higher-level analyses often relies heavily on dependency structures as well.
Dependency parsing and its evaluation method have their own advantages, such as a more direct representation of grammatical relations and often simpler parsing algorithms.
However, constituent parsing maintains the hierarchical structure of a sentence, which can still be valuable for understanding the syntactic relationships between words and phrases.
Numerous studies in formal syntax have focused on constituent structures, including combinatory categorial grammar (CCG) parsing \citep{lewis-lee-zettlemoyer:2016:NAACL,lee-lewis-zettlemoyer:2016:EMNLP,stanojevic-steedman:2020:ACL,yamaki-etal-2023-holographic} 
or tree-adjoining grammar (TAG) parsing \citep{kasai-etal-2017-tag,kasai-etal-2018-end}. Notably, CCG and TAG inherently incorporate dependency structures.
In addition to these approaches, new methods for constituent parsing, such as the linearization parsing method \citep{vinyals-etal-2015-grammar,fernandez-gonzalez-gomez-rodriguez-2020-enriched,wei-etal-2020-span}, have been actively explored. 
If a method designed to achieve the goal of creating an end-to-end system utilizes constituent structures, it necessitates more robust evaluation methods for assessing its constituent structure.

\subsection{Conclusion}

Despite the widespread use and acceptance of the previous implementation of PARSEVAL measures as the standard tool for constituent parsing evaluation, it has a significant limitation in that it requires specific task-oriented environments. Consequently, there is still room for a more robust and reliable evaluation approach.
Various metrics have attempted to address issues related to word and sentence mismatches by employing complex tree operations or adopting dependency scoring methods. In contrast, our proposed method aligns sentences and words as a preprocessing step without altering the original PARSEVAL measures. 
This approach allows us to preserve the complexity of the previous implementation of PARSEVAL while introducing a linear time alignment process.
Given the high compatibility of our method with existing PARSEVAL measures, it also ensures the consistency and seamless integration of previous work evaluated using PARSEVAL into our approach.
Ultimately, this new measurement approach offers the opportunity to evaluate constituent parsing within an end-to-end pipeline. It addresses discrepancies that may arise during earlier steps, such as sentence boundary detection and tokenization, thus enabling a more comprehensive evaluation of constituent parsing.\footnote{This chapter is based on "A Novel Alignment-based Approach for PARSEVAL Measures" by Eunkyul Leah Jo, Angela Yoonseo Park, and Jungyeul Park. \textit{Computational Linguistics} 2024; doi: \url{https://doi.org/10.1162/coli_a_00512} \citep{jo-etal-2024-novel}, and "\texttt{jp-evalb}: Robust Alignment-based PARSEVAL Measures" by Jungyeul Park, Junrui Wang, Eunkyul Jo, Angela Park. In \textit{Proceedings of the 2024 Conference of the North American Chapter of the Association for Computational Linguistics: Human Language Technologies (Volume 3: System Demonstrations)}, pages 70–77, Mexico City, Mexico. Association for Computational Linguistics \citep{park-etal-2024-jp}.
}

\section{\textsc{JP-errant}} \label{jp-errant}

\subsection{Introduction} \label{introduction}

{In modern natural language processing (NLP), end-to-end systems have become increasingly popular due to their ability to manage entire tasks from start to finish, offering streamlined and efficient solutions. In this context, evaluation is important as it allows for consistent and objective assessment of these systems, ensuring they meet the intended goals without the need for manual intervention. A good  evaluation system must be flexible, able to adapt to various tasks and data types, and robust, providing reliable results even in the face of diverse or unexpected inputs. It should also align with high-quality standards to accurately measure the effectiveness of the design or implementation being evaluated.}
For instance, the \textit{CoNLL 2017-2018 Shared Task: Multilingual Parsing from Raw Text to Universal Dependencies} \citep{zeman-etal-2017-conll,zeman-etal-2018-conll} demonstrated that a system could take raw text and parse it into a structured format that shows how words relate to each other in many languages. This approach is comprehensive, covering everything from identifying sentence boundaries and breaking the text into words, to labeling parts of speech and analyzing dependency relationships. Most importantly, the evaluation method of Universal Dependencies (UD) is designed to accurately measure the performance of the entire process, even if there are mismatches in sentence boundaries between the system's output and the predefined standard. This makes the metric flexible, robust, and applicable in various settings, accommodating differences that might arise in the preprocessing stages.

{Grammatical error correction (GEC) plays an essential role in facilitating effective communication, supporting language learning, and ensuring the accuracy of written texts. GEC systems provide automated assistance for both instructors and learners, making them invaluable tools in educational and professional settings. Over the years, various NLP systems and methodologies have been developed to enhance the effectiveness of automated GEC. Alongside these advancements, several evaluation metrics, including \texttt{M$^2$} \citep{dahlmeier-ng-2012-better}, \texttt{GLEU} \citep{napoles-etal-2015-ground}, \texttt{errant} \citep{bryant-etal-2017-automatic,bryant:2019}, and \texttt{PT M$^2$} \citep{gong-etal-2022-revisiting}, have been introduced to measure the performance and reliability of these systems, ensuring they meet the high standards required for accurate grammatical correction.}
However, these metrics often share a common limitation: they require predefined, consistent sentence boundaries between the gold standard—an ideal set of corrections—and the outputs generated by the system.

When applied to raw text input—reflecting real-world language learners’ writing scenarios—the current GEC evaluation method suffers due to differing sentence boundaries detected during preprocessing, where the sentences in learners’ writing and the predefined corrections might not align. This challenge is similar to issues faced in other NLP tasks, such as Machine Translation (MT), where sentence alignment between source and target sentences is crucial for creating a parallel corpus.
In MT, sentence alignment involves matching sentences in two or more languages, connecting each sentence in one language to its corresponding sentences in another. Sentence alignment has evolved over several decades, leading to the development of various algorithms. Initially, alignment studies relied on a length-based statistical method \citep{gale-church:1993:CL}, which used bilingual corpora to model differences in sentence length across languages as a basis for alignment. Later advancements included more sophisticated techniques like \texttt{Bleualign}, which uses an iterative bootstrapping method to refine length-based alignment. Other early approaches improved alignment accuracy by incorporating lexical correspondences, exemplified by \texttt{hunalign} \citep{varga-EtAl:2005} and the IBM model's lexicon translation approach \citep{moore:2002}. More recent efforts, like \texttt{vecalign} \citep{thompson-koehn-2019-vecalign}, integrate linguistic knowledge, heuristics, and various scoring methods to enhance alignment efficiency.

Built upon advancements in MT alignment, we propose a refined approach to address GEC-specific challenges, particularly in end-to-end evaluation scenarios. The key contributions of our work are as follows:
We introduce an alignment-based method that significantly improves end-to-end GEC evaluation by addressing sentence boundary discrepancies that often arise during preprocessing, especially when systems process raw, unsegmented text. Our approach employs an advanced \textit{jointly preprocessed} algorithm, overcoming limitations of traditional methods that rely on predefined sentence boundaries.
Moreover, we provide additional enhancements to GEC evaluation {by reimplementing \texttt{errant}}:
(i) We improve error annotation accuracy by replacing \texttt{spaCy} with \texttt{stanza} for language processing, leading to more precise part-of-speech tagging and dependency parsing {($\S$\ref{GEC-evaluation-results})}.
(ii) We extend our approach to multilingual contexts, demonstrating its potential for consistent grammatical error annotation and evaluation across multiple languages {($\S$\ref{multilingual-errant})}.

Our work aims to enhance the robustness, relevance, and real-world applicability of GEC evaluation methodologies. Our approach addresses the complexities of language learners' writing in real-world contexts, ensuring reliable evaluations across diverse text inputs and more precisely reflecting the demands of actual language usage.

\subsection{Previous GEC evaluation measures}

The MaxMatch (\texttt{M$^{2}$}) metric identifies the sequence of edits from the input to the system correction that achieves the maximum overlap with the gold standard edits, based on Levenshtein distance \citep{dahlmeier-ng-2012-better}. The \texttt{GLEU} metric extends the BLEU metric used in machine translation \citep{papineni-etal-2002-bleu}, modifying the precision calculation by giving extra weight to n-grams in the candidate text that align with the reference but not with the source (i.e., the set of n-grams $R \backslash S$). It also introduces a penalty for n-grams present in both the candidate and the source but absent in the reference, referred to as false negatives ($S \backslash R$) \citep{napoles-etal-2015-ground}.

A novel pretraining-based approach to \texttt{M$^{2}$} uses BERTScore and BARTScore to calculate edit scores, allowing assessments based on insights from pretrained metrics. However, directly applying PT-based metrics often yields unsatisfactory correlations with human judgments due to an excessive focus on unchanged sentence parts. To address this, \texttt{PT-M$^{2}$} has been introduced, leveraging PT-based metrics only for scoring corrected parts, significantly improving correlation with human evaluations and achieving a state-of-the-art Pearson correlation of 0.95 on the CoNLL14 evaluation task \citep{gong-etal-2022-revisiting}.

While these different metrics have their strengths and limitations, currently \texttt{errant} (ERRor ANnotation Toolkit) is the de facto standard for evaluating GEC. 
\texttt{errant} compares error annotations between the gold standard and system m2 files\footnote{m2 is a common format for representing grammatical errors and corrections. For each sentence, it includes the original tokenized text (the \textit{S} line), and one or more error annotation lines (\textit{A} lines). These \textit{A} lines contain the position of each error, the error type (or no error), the correction, and other information. See Figure~\ref{jp-errant-procedure} for an example in English.}, calculating precision, recall, and reporting the F$_{0.5}$ score. This score emphasizes precision over recall, reflecting the importance of providing accurate feedback to language learners in GEC systems \citep{bryant-etal-2017-automatic,bryant:2019}. \texttt{errant} addresses an important limitation of the original \texttt{M$^{2}$} metric--the tendency to inflate performance by heavily weighting true positives.
Another advantage of \texttt{errant} over other metrics is that in addition to providing a score, it also offers detailed error annotation, which facilitates a deeper analysis of system performance and specific error patterns.
\texttt{errant} has been adapted for multiple languages, including German, Chinese, and Korean, among others \citep{boyd-2018-using, hinson-etal-2020-heterogeneous,zhang-etal-2022-mucgec,sonawane-etal-2020-generating, belkebir-habash-2021-automatic,naplava-etal-2022-czech, katinskaia-etal-2022-semi,yoon-etal-2023-towards}.

{Given the advantages of \texttt{errant} and its widely accepted status as the de facto standard for GEC evaluation, our work adapted \texttt{errant} by incorporating an alignment-based preprocessing approach. This adaptation addresses challenges in end-to-end GEC scenarios, particularly discrepancies in sentence boundaries between the gold standard and system predictions during preprocessing. Our method ensures accurate evaluations even with differing sentence boundaries, maintaining \texttt{errant}'s reliability in real-world GEC applications.}

\subsection{Alignment-based \texttt{errant}} \label{algo-jp-errant}

We utilize an alignment-based evaluation algorithm to enhance end-to-end GEC evaluation measures. Recognizing that sentence boundaries between the gold standard and system outputs may vary during preprocessing, this algorithm employs sentence alignment to accurately match sentences from the gold and system GEC results, ensuring correct evaluations. Consequently, while the fundamental GEC evaluation measures remain unchanged, they are now applied to sentence-aligned results, improving the accuracy and reliability of the metrics.

We adapt a \textit{jointly preprocessed} algorithm, {where it preprocesses sentence boundary and tokenization between source and target through alignment,} as described in Algorithm~\ref{jp-sentence-algorithm}. This algorithm is specifically designed for environments where gold and system sentences are nearly identical in a monolingual context. {A similar alignment-based joint preprocessing approach has been shown to be effective in improving evaluation of {constituent parsing \citep{jo-etal-2024-novel,park-etal-2024-jp} and preprocessing tasks \citep{jo-etal-2024-untold-story} where they have shown the effectiveness of the algorithm for several languages including several European languages as well as Chinese and Korean}.}
This contrasts with traditional sentence alignment methods in MT that often require recursive editing to accommodate significant differences between source and target languages. In cases of mismatches due to varying sentence boundaries, our  pattern matching-based algorithm accumulates sentences until it finds a matching pair. The computational efficiency of our approach is notable, requiring linear time, $O(m+n)$, where $m$ and $n$ are the lengths of the gold and system sentences, respectively. This is a significant improvement over the traditional cubic complexity, $n^3$, of standard length-based sentence alignment algorithms in MT.
The proposed \texttt{jp-algorithm} introduces sentence alignment to ensure correct GEC evaluations.




\begin{algorithm}[!ht]
\caption{Pseudo-code for sentence alignment}\label{jp-sentence-algorithm}
{\footnotesize
\begin{algorithmic}[1]
\STATE{\textbf{function} \textsc{PatternMatchingSA} ($\mathcal{L}$, $\mathcal{R}$):}
\begin{ALC@g}
\WHILE{$\mathcal{L}$ and $\mathcal{R}$}
\IF{$\mathcal{L}_{i({\not\sqcup})}$ $=$ $\mathcal{R}_{j({\not\sqcup})}$
}
\STATE{$\mathcal{L^{\prime}}$, $\mathcal{R^{\prime}}$ $\gets$ $\mathcal{L^{\prime}}+\mathcal{L}_i, \mathcal{R^{\prime}}+\mathcal{R}_j$ where $0<i\leq {\textsc{len}}(\mathcal{L})$, $0<j\leq {\textsc{len}}(\mathcal{R})$}
\ELSE
\WHILE{$\neg$$(\mathcal{L}_{i({\not\sqcup})}$ $=$ $\mathcal{R}_{j({\not\sqcup})})$
}
\IF{$\textsc{len}(\mathcal{L}_{i}) < \textsc{len}(\mathcal{R}_{j})$}
\STATE{${L}^{\prime}$ $\gets$ ${L}^{\prime}+\mathcal{L}_{i}$} 
\STATE{$i \gets i+1 $}
\ELSE
\STATE{${R}^{\prime}$ $\gets$ ${R}^{\prime}+\mathcal{R}_{j}$}
\STATE{$j \gets j+1 $}
\ENDIF

\ENDWHILE
\STATE{
$\mathcal{L^{\prime}}$, $\mathcal{R^{\prime}}$ 
$\gets$ 
$\mathcal{L^{\prime}}+L^{\prime}, \mathcal{R^{\prime}}+R^{\prime}$
}
\ENDIF
\ENDWHILE
\RETURN {$\mathcal{L^{\prime}}$, $\mathcal{R^{\prime}}$}
\end{ALC@g}
\end{algorithmic} 
}
\end{algorithm}

In Equation~\eqref{sa-case11}, we define that sequences $\mathcal{L}_i$ and $\mathcal{R}_j$ can be aligned if they match when all spaces are removed, denoted as $\mathcal{L}_i\not\sqcup$ and $\mathcal{R}_j\not\sqcup$. This method aims to minimize differences due to tokenization. 
\begin{equation}
\mathcal{L}_{i({\not\sqcup})} == \mathcal{R}_{j({\not\sqcup})}   \label{sa-case11} 
\end{equation}
However, simply removing spaces and concatenating words may not sufficiently identify identical sentence pairs. Variations in tokenization can introduce grammatical morphemes absent in the gold-standard sentences, or vice versa. For example, the contraction \textit{can't} presents tokenization challenges, as it can be tokenized as \textit{can not} or \textit{ca n't}, with each version introducing different characters. Such variations mean that our character-level evaluation may fail to accurately capture these discrepancies. The inconsistency in tokenization standards across different corpora further complicates this issue. For instance, the English UD corpus from EWT tokenizes \textit{can't} as \textit{ca} and \textit{n't}, whereas ParTUT tokenizes it as \textit{can} and \textit{not}.
We therefore propose to align sequences $\mathcal{L}_{i}$ and $\mathcal{R}_{j}$, denoted as $\mathcal{L}_{i\not\sqcup}$ and $\mathcal{R}_{j\not\sqcup}$ respectively, when they demonstrate close character similarities that exceed a predefined threshold, $\alpha$. Moreover, the subsequent sequences, $\mathcal{L}_{i+1}$ and $\mathcal{R}_{j+1}$, must either directly match or exhibit sufficient similarity, as outlined in Equation~\eqref{sa-case22}.
\begin{eqnarray}
(\mathcal{L}_{i({\not\sqcup})} \simeq \mathcal{R}_{j({\not\sqcup})}) \land  \nonumber\\(\mathcal{L}_{i+1({\not\sqcup})} == \mathcal{R}_{j+1({\not\sqcup})} \lor \mathcal{L}_{i+1({\not\sqcup})} \simeq \mathcal{R}_{j+1({\not\sqcup})}) \label{sa-case22}
\end{eqnarray}
We modify the Jaro-Winkler distance, traditionally used to measure the similarity between two strings, by incorporating a suffix scale in addition to the existing prefix scale. The original Jaro similarity, denoted by $sim_j$, calculates matches based on the number of forward-matching characters between two strings, $s_1$ and $s_2$. The Jaro-Winkler distance enhances this similarity by introducing a prefix scale $p$ for a specified prefix length $l$. Our modification extends this method by adding a similar scale for a defined suffix length, thereby improving the algorithm’s ability to recognize suffix similarities as well.
\begin{equation}
\alpha = sim_{j} - \frac{(lp + l'p)(1 - sim_{j})}{2}
\end{equation}
\noindent where $sim_j$ is the Jaro similarity between two strings $s_1$ and $s_2$, $l$ and $l'$ are the lengths of the common prefix and suffix of the strings, respectively, and $p$ is a constant scaling factor (set to 0.1).\footnote{The value of $\alpha$ represents a trade-off between the correctness and precision of the alignments. If $\alpha$ is too small, we risk boundary errors (false positives), while if $\alpha$ is too large, we may miss some boundaries (false negatives).}
If $\mathcal{L}_i$ and $\mathcal{R}_j$ cannot be aligned, we proceed by concatenating sequences based on their lengths. Specifically, if the length of $\mathcal{L}_{i:m}$ exceeds that of $\mathcal{R}_{j:n}$, then $\mathcal{L}_i$ is concatenated with $\mathcal{L}_{i+1}$. Conversely, if $\mathcal{R}_{j:n}$ is longer, then $\mathcal{R}_j$ is concatenated with $\mathcal{R}_{j+1}$. This concatenation process is repeated until the pairs $\mathcal{L}_{i+1}$ and $\mathcal{R}_{j+1}$ meet the established sentence matching criteria.

We have re-implemented \texttt{errant} incorporating a joint preprocessing step, now referred to as \texttt{jp-errant}. Unlike the original \texttt{errant}, which used \texttt{spaCy} for language processing, \texttt{jp-errant} employs the part-of-speech tagging capabilities of \texttt{stanza} \citep{qi-etal-2020-stanza}, chosen for its demonstrably clear performance.\footnote{\url{https://stanfordnlp.github.io/stanza/performance.html}}
We maintain the original error annotation scheme of \texttt{errant}, but we adjust word positions by re-indexing them during the sentence alignment process when sentences are concatenated. This adjustment is crucial to handle discrepancies in sentence boundaries between the gold standard and those processed by \texttt{stanza}.

In the alignment algorithm, concatenating sentences necessitates updates to the positions of corresponding edits. After sentence alignment, the re-indexing process is carried out in two primary steps:
First, we update all non-empty edits. For each concatenated sentence, we accumulate its token count to serve as the offset for subsequent edits. Consider the following example: when concatenating sentences $S_{i} = [w_{1}, w_{2}, ..., w_{m}]$ and $S_{j} = [w_{1}, w_{2}, ..., w_{n}]$, the edits $E_{i} = [e_{(a, b)}]$ and $E_{j} = [e_{(c, d)}]$ are adjusted to become $[e_{(a, b)}, e_{(c+m, d+m)}]$ in the concatenated sequence.
Second, after re-indexing all edits, we proceed to remove any unnecessary empty edits. This step is crucial to ensure that the m2 results are fully aligned, with no unnecessary empty edits.

Details of the re-indexing process by \texttt{jp-errant} are shown in Figure~\ref{jp-errant-procedure}.
For example, the m2 file's state is shown both before and after sentence alignment. The gold m2 file features an empty edit (\texttt{-1 -1|||noop}) and a capitalization edit (\texttt{0 1|||R:ADV}), where the adverb \textit{how} is replaced by \textit{How}. Conversely, the \texttt{stanza} m2 file contains two empty edits, which indicate places where no grammatical errors were corrected by the system. 

\begin{figure}[!ht]
\centering
\footnotesize{
\begin{tabular}{l| l}\hline 
\multicolumn{2}{l}{\cellcolor{gray!10}1. Preparation} \\ \hline 
\multicolumn{1}{r|}{gold m2} 
& \texttt{S Kate Ashby ,}\\
& \texttt{A -1 -1|||noop|||-NONE-|||REQUIRED|||-NONE-|||0} \\
& \texttt{S how are you ? I hope you are well .}\\
& \texttt{A 0 1|||R:ADV|||How|||REQUIRED|||-NONE-|||0}\\ 
\hdashline

\multicolumn{1}{r|}{\texttt{stanza} m2} 
& \texttt{S Kate Ashby , how are you ?}\\
& \texttt{A -1 -1|||noop|||-NONE-|||REQUIRED|||-NONE-|||0} \\
& \texttt{S I hope you are well .}\\
& \texttt{A -1 -1|||noop|||-NONE-|||REQUIRED|||-NONE-|||0}\\ 
\hline 

\multicolumn{2}{l}{\cellcolor{gray!10}2. Sentence alignment} \\ \hline 
\multicolumn{1}{r|}{gold m2} & \texttt{S Kate Ashby , how are you ? I hope you are well .}\\
& \texttt{A -1 -1|||noop|||-NONE-|||REQUIRED|||-NONE-|||0} \\
& \texttt{A 0 1|||R:ADV|||How|||REQUIRED|||-NONE-|||0}\\ 
\hdashline

\multicolumn{1}{r|}{\texttt{stanza} m2} & \texttt{S Kate Ashby , how are you ? I hope you are well .}\\
& \texttt{A -1 -1|||noop|||-NONE-|||REQUIRED|||-NONE-|||0} \\
& \texttt{A -1 -1|||noop|||-NONE-|||REQUIRED|||-NONE-|||0}\\ \hline

\multicolumn{2}{l}{\cellcolor{gray!10}3. Re-indexing} \\ \hline 
\multicolumn{1}{r|}{gold m2} & \texttt{S Kate Ashby , how are you ? I hope you are well .}\\
& \texttt{A  3 4|||R:ADV|||How|||REQUIRED|||-NONE-|||0} \\ \hdashline

\multicolumn{1}{r|}{\texttt{stanza} m2} & \texttt{S Kate Ashby , how are you ? I hope you are well .}\\
& \texttt{A -1 -1|||noop|||-NONE-|||REQUIRED|||-NONE-|||0}    \\ \hline 
\end{tabular}
}
\caption{Procedure example of \texttt{jp-errant}: \texttt{stanza} m2 indicates that sentence boundaries are detected by \texttt{stanza} from raw text.} \label{jp-errant-procedure}
\end{figure}

\subsection{Experiments and results} \label{errant-experiments-results}

\paragraph{GEC dataset}\label{gec-dataset}
For our experiments, we utilized the development dataset from the Cambridge English Write \& Improve (W\&I) corpus, which was introduced during the \textit{Building Educational Applications 2019 Shared Task: Grammatical Error Correction (BEA2019)} \citep{bryant-etal-2019-bea}. This dataset is manually annotated with Common European Framework of Reference (CEFR) proficiency levels—beginner (A), intermediate (B), and advanced (C) \citep{yannakoudakis-etal-2018-wi}. The texts, written by L2 English learners, show a trend where sentences from higher proficiency levels tend to be longer than those from lower levels. Specifically, the average token counts per sentence for levels A, B, and C are 17.538, 18.304, and 19.212, respectively. We analyzed the distribution of errors across different language proficiency levels. The error types in levels B and C are similar, including missing punctuation marks (\texttt{M:PUNCT}), incorrect prepositions (\texttt{R:PREP}), and missing determiners (\texttt{M:DET}). Additionally, level A frequently exhibits orthographic errors (\texttt{R:ORTH}), such as case or whitespace issues. 
Table~\ref{top-error-type} presents the ratios of the most frequent error types within the W\&I training data, where the ratios represent the distribution of grammatical errors \citep{zeng-etal-2024-evaluating-prompting}.

\begin{table}[!ht]
\centering
\begin{center}
\footnotesize{
\begin{tabular}{cc | cc | cc} \hline
\multicolumn{2}{c|}{Proficiency A} & \multicolumn{2}{c|}{Proficiency B} & \multicolumn{2}{c}{Proficiency C}\\ \hline 
M:PUNCT & 0.0933 & M:PUNCT & 0.1134 & M:PUNCT & 0.1183 \\  
R:ORTH & 0.0602 & R:PREP & 0.0589 & R:PREP & 0.0517 \\  
R:PREP & 0.0506 & M:DET & 0.0442 & M:DET & 0.0345 \\  
R:VERB:TENSE & 0.0455 & R:VERB & 0.0414 & R:VERB & 0.0323 \\  
R:VERB & 0.0419 & R:VERB:TENSE & 0.0393 & R:VERB:TENSE & 0.0273 \\   \hline
\end{tabular}
}
\end{center}  
\caption{The most frequent errors and their ratios in the W\&I dataset}
\label{top-error-type}

\end{table}

\paragraph{GEC evaluation results} \label{GEC-evaluation-results}
We utilized two off-the-shelf state-of-the-art GEC systems: \textsc{gector} \citep{omelianchuk-etal-2020-gector} and \textsc{t5} \citep{rothe-etal-2021-simple}. Briefly, \textsc{gector} employs a sequence tagging approach instead of sequence generation. This system uses a Transformer encoder to predict token-level edit operations, making it significantly faster and more efficient than traditional seq2seq models. In our experiments, we used the \texttt{RoBERTa} pre-trained model as the encoder, which showed the best performance among various transformer models tested \citep{omelianchuk-etal-2020-gector}.\footnote{\url{https://github.com/grammarly/gector}}
\textsc{t5}, or Text-To-Text Transfer Transformer, is a unified framework for NLP tasks that converts all tasks into a text-to-text format. We used the \texttt{T5-small} model, a smaller variant with approximately 60 million parameters. This model was fine-tuned on the cleaned English LANG-8 corpus and achieved reported $F_{0.5}$ scores of $60.54$ and $65.01$ on the CoNLL2014 and BEA2019 test sets, respectively \citep{rothe-etal-2021-simple}.\footnote{\url{https://huggingface.co/Unbabel/gec-t5_small}}

The GEC results using the original \texttt{errant} and \texttt{jp-errant} with the gold-standard sentence boundaries, as well as \texttt{jp-errant} with the system-generated sentence boundaries, are presented in Table~\ref{experiment-results}. {It's important to note that the original \texttt{errant} does not allow the use of different sentence boundaries, which precludes a ``\textsc{sys}  + \texttt{errant}'' setup.} {The results of GEC may also differ between texts with gold-standard sentence boundaries and those with system-generated boundaries due to the nature of sequence-to-sequence GEC.}
The former shows the reproducibility of \texttt{jp-errant}, while the latter presents how the proposed method can handle real-world scenarios.
If there are mismatches in sentence boundaries between the gold-standard and system-generated results, we initiate a sentence alignment process.

\begin{table}[!ht]
\centering
\resizebox{\textwidth}{!}{
\footnotesize{
\begin{tabular}{ r | cccccc | cccccc} \hline 

\multicolumn{1}{c|}{\textsc{gector}} &  \multicolumn{6}{c|}{A} & \multicolumn{6}{c}{B} \\ 
& TP & FP & FN &  Prec & Rec & F0.5 & TP & FP & FN & Prec & Rec & F0.5\\ \hline 
\textsc{gold} + \texttt{errant} & 1299&798&1680&0.6195&0.4361&0.5714& 1049&621&1470&0.6281&0.4164&0.5702\\ \hdashline
\textsc{gold} + \texttt{jp-errant} & 1295&794&1678&0.6199&0.4356&0.5715&1047&623&1456&0.6269&0.4183&0.5701\\ \hdashline
\textsc{sys}  + \texttt{jp-errant}& 1220&842&1753&0.5917&0.4104&0.5436&1039&626&1464&0.624&0.4151&0.5670 \\ \hline
&  \multicolumn{6}{c|}{C}  & \multicolumn{6}{c}{all}\\ 
& TP & FP & FN & Prec & Rec & F0.5  & TP & FP & FN & Prec & Rec & F0.5\\ \hline  
\textsc{gold} + \texttt{errant} & 415&350&706&0.5425&0.3702&0.4963& 2763&1769&3856&0.6097&0.4174&0.5582\\ \hdashline
\textsc{gold} + \texttt{jp-errant} &414&350&703&0.5419&0.3706&0.496&2756&1767&3837&0.6093&0.4180&0.5582\\ \hdashline
\textsc{sys}  + \texttt{jp-errant} &413&347&704&0.5434&0.3697&0.4968 & 2672&1815&3921&0.5955&0.4053&0.5444\\ \hline \hline 

\multicolumn{1}{c|}{\textsc{t5}} &  \multicolumn{6}{c|}{A} & \multicolumn{6}{c}{B} \\ 
& TP & FP & FN &  Prec & Rec & F0.5 & TP & FP & FN & Prec & Rec & F0.5\\ \hline 
\textsc{gold} + \texttt{errant} & 1271&696&1708&0.6462&0.4267&0.5859&960&593&1559&0.6182&0.3811&0.5498\\ \hdashline
\textsc{gold} + \texttt{jp-errant} & 1265&689&1708&0.6474&0.4255&0.5862 & 945&592&1558&0.6148&0.3775&0.5462\\ \hdashline
\textsc{sys}  + \texttt{jp-errant} & 1173&771&1800&0.6034&0.3946&0.5456&928&614&1575&0.6018&0.3708&0.5351\textsc{} 
\\ \hline
&  \multicolumn{6}{c|}{C}  & \multicolumn{6}{c}{all}\\ 
& TP & FP & FN & Prec & Rec & F0.5  & TP & FP & FN & Prec & Rec & F0.5\\ \hline  
\textsc{gold} + \texttt{errant} & 358&350&763&0.5056&0.3194&0.4528& 2589&1639&4030&0.6123&0.3911&0.5501\\ \hdashline
\textsc{gold} + \texttt{jp-errant} & 355&332&762&0.5167&0.3178&0.4592&2565&1613&4028&0.6139&0.3890&0.5503 \\ \hdashline
\textsc{sys}  + \texttt{jp-errant} &  351&332&766&0.5139&0.3142&0.456&2452&1717&4141&0.5882&0.3719&0.5269\\ \hline 
\end{tabular}
}
}
\caption{SOTA GEC results achieved by \textsc{gector}  and \textsc{t5} with the English-specific error classification module with gold and system sentence boundaries} \label{experiment-results}
\end{table}

We have integrated the English-specific classification module from the original \texttt{errant}, which identifies types of grammatical errors for English. This module categorizes detailed grammatical error types, such as \texttt{NOUN:POSS} for errors related to possessive noun suffixes. It utilizes universal part-of-speech tags \citep{petrov-das-mcdonald:2012:LREC} and dependency relation tags to determine error types. For instance, if the first token in an edit is tagged as \texttt{PART}—typically indicating particles or function words—and its dependency relation is \texttt{case:poss} (indicative of a possessive case), the classifier labels it as a \texttt{NOUN:POSS} error based on this information.
Without the classification module, \texttt{jp-errant} can still produce generic error annotations with corresponding POS labels and evaluate the results based on error edits regardless of languages. However, with this language-specific classification module, it can generate language-specific error annotations for other languages.

While we successfully reproduced \texttt{errant}, discrepancies persist, as shown in Figure~\ref{errant-diff}. One notable distinction lies in preposition naming: all POS labels adhere to the Universal POS label names, yet the original \texttt{errant} continues to use \texttt{PREP} instead of \texttt{ADP} for prepositions. Another difference arises from an error in POS tagging by \texttt{spaCy}, which was employed by the original \texttt{errant}. In the second sentence, \textit{your} at positions 3 and 4 is identified as a pronoun (\texttt{PRON}) by \texttt{stanza}, whereas \texttt{spaCy} labels it as a determiner (\texttt{DET}).

\begin{figure}[!ht]
\centering
{
\footnotesize{
\begin{tabular}{l}\hline 
\cellcolor{gray!10}\texttt{errant} \\
\texttt{S It 's difficult answer at the question " ... } \\
\texttt{A 3 3|||M:VERB:FORM|||to|||REQUIRED|||-NONE-|||0} \\
\texttt{A 4 5|||{\color{red}U:ADP}||||||REQUIRED|||-NONE-|||0}\\ 
\hdashline
\texttt{S Thank you for your e - mail , it was wonderful to hear from you .}\\
\texttt{A 3 4|||{\color{red}R:DET}|||your|||REQUIRED|||-NONE-|||0} \\
\texttt{A 7 9|||R:PUNCT|||. It|||REQUIRED|||-NONE-|||0} \\\hline 

\cellcolor{gray!10}\texttt{jp-errant} \\
\texttt{S It 's difficult answer at the question " ...}  \\
\texttt{A 3 3|||M:VERB:FORM|||to|||REQUIRED|||-NONE-|||0} \\
\texttt{A 4 5|||{\color{green}U:PREP}||||||REQUIRED|||-NONE-|||0} \\ \hdashline
\texttt{S Thank you for your e - mail , it was wonderful to hear from you .}\\
\texttt{A 3 4|||{\color{green}R:PRON}|||your|||REQUIRED|||-NONE-|||0} \\
\texttt{A 7 9|||R:PUNCT|||. It|||REQUIRED|||-NONE-|||0}\\ \hline 
\end{tabular}
}
}
    \caption{Differences between \texttt{errant} and \texttt{jp-errant}}
    \label{errant-diff}
\end{figure}

\subsection{Multilingual alignment-based \texttt{errant}} \label{multilingual-errant}

{We present multilingual \texttt{errant} results, focusing on Chinese and Korean L2 GEC. By aligning and evaluating these languages, we demonstrate the challenges and potential solutions in applying the proposed GEC evaluation methodology across different languages.}

\paragraph{Chinese L2 GEC dataset}
Multi-Reference Multi-Source Evaluation Dataset for Chinese Grammatical Error Correction (CGEC) is a multi-reference multi-source dataset comprising sentences from the NLPCC18 test set \citep{zhao-et-al-2018-nlpcc}, CGED-2018 and CGED-2020 test datasets \citep{rao-etal-2018-overview,rao-etal-2020-overview}, and randomly selected Lang-8 dataset \citep{zhang-etal-2022-mucgec}.\footnote{\url{https://github.com/HillZhang1999/MuCGEC}} 
The MuCGEC dataset exhibits an average number of target references per sentence exceeding 2. 
They discovered that augmenting the average number of references per sentence enhances the reliability of evaluations, attributed to its multi-reference characteristics.

\paragraph{Korean L2 GEC dataset}
The GEC dataset for Korean consists of two distinct types: L1 writing from the Center for Teaching and Learning for Korean, and L2 writing from the National Institute of Korean Language (NIKL) corpus \citep{yoon-etal-2023-towards}.\footnote{\url{https://github.com/soyoung97/Standard_Korean_GEC}}
This dataset includes the original text, the corrected text, and its corresponding error-annotated errant-style m2 file, automatically generated for Korean. Utilizing the proposed split (70/15/15) of the GEC dataset for Korean.
The original L2 dataset is collected by NIKL with 3613 files. It provides the original L2 sentences, and their morphological segmentation with part-of-speech tags. 
The correction is annotated at the morpheme level by adding, deleting and replacing the grammatical error morpheme.

\paragraph{Error annotation for Chinese and Korean}
{Figure~\ref{multilingual-m2} shows examples of differences in error annotation (i.e., m2 files) for Chinese and Korean.}
{\texttt{ChERRANT} (Chinese \texttt{errant}), an adaptation of the original English \texttt{errant}, is the most recent development in annotating Chinese grammatical errors and evaluating Chinese GEC systems \citep{zhang-etal-2022-mucgec}. \texttt{ChERRANT} operates at two levels of granularity: character-based and word-based, and primarily categorizes errors into three operational types: redundant (\texttt{R}, equivalent to \texttt{U} for unnecessary in the original \texttt{errant}), missing (\texttt{M}), and substitution (\texttt{S}, equivalent to \texttt{R} for replacement in the original \texttt{errant}) errors. For word-based error annotation, \texttt{ChERRANT} utilizes LTP-based word segmentation \citep{che-etal-2010-ltp}\footnote{\url{https://github.com/HIT-SCIR/ltp}} and converts its part-of-speech (POS) tags to Universal POS labels \citep{petrov-das-mcdonald:2012:LREC}. While word-based annotation allows for more detailed error categorization (e.g., \textsc{s:verb} would indicate a verb substitution error), due to potential word segmentation inaccuracies, \texttt{ChERRANT} applies character-level annotation by default and has not fully implemented all word-level annotations. \texttt{jp-errant}, as proposed in the current work, attempts to address the limitations of \texttt{ChERRANT} by: 1) using \texttt{stanza} for improved Chinese word segmentation and POS tagging, and 2) adhering to the original \texttt{errant} conventions (\texttt{M}issing, \texttt{U}necessary, and \texttt{R}eplacement) for consistent multilingual grammatical error annotation.}

{While the original Koraen L2 dataset collected by the National Institute of Korean Language (NIKL), provided error annotations with three different levels based on the POS of the morpheme (e.g. noun, verb, case marker, and other grammatical categories), transformation of the morpheme (omission, addition, replacement, and misformation\footnote{The term \textit{misformation} by NIKL is not commonly used to refer to a spelling error.}), and its linguistic dimensions (pronunciation, syntax, and discourse), the previous work proposed fourteen error types  such as \textsc{insertion}, \textsc{deletion}, and \textsc{ws} (word space) at the word level by converting sequences of morphemes into words \citep{yoon-etal-2023-towards}.
Given that grammatical errors manifest at the morpheme level, while the current error annotation operates at the word level, the previous work established two priority rules for categorizing error types to assign a single error type for each word, as follows: (i) \textsc{insertion} $>$ \textsc{deletion} $>$ others, and (ii) \textsc{ws} $>$ \textsc{wo} $>$ \textsc{spell} $>$ \textsc{shorten} $>$ \textsc{punctuation} $>$ others, where \textsc{wo} stands for 'word order'. However, the \textsc{punctuation} and \textsc{wo} error types do not appear in the m2 files, indicating that this type of error might not have been explicitly annotated or utilized in the previous work.}

\begin{figure}[!ht]
\begin{subfigure}[b]{\textwidth}    
\centering
\footnotesize{
\begin{tabular}{l}
{m2 file generated by \texttt{ChERRANT}}:\\
\texttt{S \zh{对 一个 生名 来说 空气 污染 是 很 危害 的 问题 ， 对 身体 不好 。}}\\
\texttt{T0-A0 \zh{对 一个 生命 来说 空气 污染 是 有 很 大 危害 的 问题 ， 对 身体 不好 。}}\\
\texttt{A 2 3|||S:SPELL|||生命|||REQUIRED|||-NONE-|||0}\\
\texttt{A 7 7|||M:VERB|||有|||REQUIRED|||-NONE-|||0}\\
\texttt{A 8 8|||M:ADJ|||大|||REQUIRED|||-NONE-|||0}\\
~\\
{m2 file generated by \texttt{jp-errant}}:\\
\texttt{S \zh{对 一个 生名 来说 空气 污染 是 很 危害 的 问题 ， 对 身体 不好 。}}\\
\texttt{A 2 3|||R:NOUN|||生命|||REQUIRED|||-NONE-|||0}\\
\texttt{A 7 7|||M:VERB|||有|||REQUIRED|||-NONE-|||0}\\
\texttt{A 8 8|||M:ADJ|||大|||REQUIRED|||-NONE-|||0}\\
\end{tabular}
}
\caption{Example of word based m2 files for Chinese (`Air pollution is a very harmful problem to a life and is bad for the body.')}
\label{chinese-word-based-m2}
\end{subfigure}

\hfill

\begin{subfigure}[b]{\textwidth}    
\centering
\footnotesize{
\begin{CJK}{UTF8}{mj}
\begin{tabular}{l}
{m2 file generated by \texttt{KAGAS}}:\\
\texttt{S 한국어수업할때 너무 자고 싶었다}\\
\texttt{A 0 1|||WS|||한국어 수업할 때|||REQUIRED|||-NONE-|||0}\\
\texttt{A 4 4|||INSERTION|||.|||REQUIRED|||-NONE-|||0}\\
~\\
{m2 file generated by \texttt{jp-errant}}:\\
\texttt{S 한국어수업할때 너무 자고 싶었다}\\
\texttt{A 0 1|||R:NOUN VERB NOUN|||한국어 수업할 때|||REQUIRED|||-NONE-|||0}\\
\texttt{A 4 4|||M:PUNCT|||.|||REQUIRED|||-NONE-|||0}\\
\end{tabular}
\end{CJK}
}
\caption{Example of m2 files for Korean (`I really wanted to sleep during Korean class.')}
\label{korean-m2}
\end{subfigure}
\caption{Difference example of m2 files for Chinese and Korean}\label{multilingual-m2}
\end{figure}


\paragraph{Multilingual experiments and results}
For Chinese GEC, we employed a recent system based on the Chinese \texttt{BART-large} model \citep{zhang-etal-2023-nasgec}. 
The model was trained using a combination of the HSK and Lang8 datasets, totaling approximately 1.3 million sentence pairs.
This model is designed to effectively handle grammatical errors in texts written by Chinese L2 learners.\footnote{\url{https://huggingface.co/HillZhang/real_learner_bart_CGEC}}
Similarly, for Korean GEC, we utilized a system based on the Korean \texttt{BART} model \citep{yoon-etal-2023-towards}. 
The model was fine-tuned using both Korean L1 and L2 datasets.
Compared to Hanspell, a widely regarded top rule-based Korean GEC system, this model demonstrates both efficiency and superior performance in correcting grammatical errors in Korean texts.\footnote{\url{https://huggingface.co/Soyoung97/gec_kr}}
Table~\ref{multilingual-results} presents the multilingual GEC results using \texttt{jp-errant}, where all input text is concatenated into a single text block, and sentence boundaries are detected using \texttt{stanza}.
{In previous work, language-specific tools were used for error annotation and evaluation: \texttt{ChERRANT} for Chinese \citep{zhang-etal-2022-mucgec} and KAGAS for Korean \citep[Korean Automatic Grammatical error Annotation System;][]{yoon-etal-2023-towards}. In the present work, we apply the generic \texttt{jp-errant} to generate m2 files for Chinese and Korean without any language-specific error classification.} The m2 files generated by \texttt{jp-errant} adhere to the original m2 file conventions (\texttt{M}issing, \texttt{U}necessary, and \texttt{R}eplacement) to support future multilingual GEC evaluations. Discrepancies between previous work and \texttt{jp-errant} would be reduced if language-specific error classifications were provided. 
{A possible reason that \texttt{ChERRANT} and \texttt{jp-errant} have similar results on Chinese L2 data is there are few language-specific grammatical error annotations. According to results from \texttt{ChERRANT}, about 4.56\% of the errors in the validation set belong to language-specific error types that were not part of the the original \texttt{errant}: spelling errors (\texttt{SPELL}), missing components (\texttt{MC}), and quantifier errors (\texttt{QUANT}). Spelling errors in Chinese are defined based on the strings similarity in pinyin pronunciation and character shape. Missing components are defined as placeholders for important words that were somehow omitted in the source sentence. Quantifiers are were not part of the original \texttt{errant} labels. Among these, spelling errors are the most common, accounting for 3.96\% of all errors.}
{In contrast, Table~\ref{multilingual-results} shows large discrepancies in the Korean results. It has been recognized that the previous KAGAS system has limitations in error detection and annotation. Our current implementation using \texttt{stanza} demonstrates improved language processing capabilities for Korean. A refined Korean-specific error annotation system is currently in development, aiming to address existing issues and enhance the accuracy of Korean GEC evaluation.}

\begin{table}[!ht]
\centering
\resizebox{\textwidth}{!}{
\footnotesize{
\begin{tabular}{ r | ccc ccc | ccc ccc} \hline 
& \multicolumn{6}{c|}{Chinese} &  \multicolumn{6}{c}{Korean}\\
& TP & FP &  FN & Prec & Rec & F0.5 & TP & FP &  FN & Prec & Rec & F0.5 \\ \hline 
\textsc{gold} + previous work & 699	&1854&	3534&	0.2738&	0.1651&	0.242 & 3037&	7323&	5822&	0.2931&	0.3428&	0.3019 \\
\textsc{gold} + jp-errant & 788&	2215&	3765&	0.2624&	0.1731&	0.2379 & 3187&	2212&	5090&	0.5903&	0.385&	0.5334\\ \hline 
\textsc{sys} + jp-errant & 499& 1781& 4054& 0.2189& 0.1096& 0.1825& 3418& 2307& 4859& 0.597& 0.413& 0.5482\\ \hline 
\end{tabular}
}
}
\caption{Multilingual GEC results by previous work and \texttt{jp-errant} with gold and system sentence boundaries} \label{multilingual-results}
\end{table}

\subsection{Conclusion} \label{errant-conclusion}

In this study, we addressed various challenges imatical error correction by implementing and refining a methodology for end-to-end processing. We demonstrated the effectiveness of our methods by reproducing previous methodologies and highlighting our contributions.
Our primary contribution is the refined alignment process, which addresses discrepancies in sentence boundaries between gold-standard and system-generated results. Despite successfully reproducing \texttt{errant} using \texttt{stanza}, we identified persistent discrepancies, such as differences in POS label naming conventions and tagging errors, which was used in the original \texttt{errant}.
We also took an additional step to facilitate multilingual evaluation by generalizing \texttt{errant}, and we presented case studies for Chinese and Korean, which enabled potential multilingual evaluation for GEC.
Such universality will provide a framework for the consistent annotation of grammatical errors across different languages.
In conclusion, we lay a robust foundation for advancing grammatical error correction, particularly its evaluation, and enhancing its applicability in real-world contexts.\footnote{This chapter is based on "Refined Evaluation for End-to-End Grammatical Error Correction Using an Alignment-Based Approach" by Junrui Wang, Mengyang Qiu, Yang Gu, Zihao Huang, and Jungyeul Park. 
In \textit{Proceedings of the 31st International Conference on Computational Linguistics}, pages 774–785, Abu Dhabi, UAE. Association for Computational Linguistics \citep{wang-etal-2025-refined}.}

\section{Discussion on Evaluation by Alignment}

This chapter argued that many standard evaluation protocols in natural language processing implicitly assume that system output and reference output share the same segmentation decisions, including sentence boundaries and token boundaries. In practice, end to end systems operating on raw text routinely violate this assumption, so boundary mismatches can dominate the score and obscure the behavior of the model that the evaluation is intended to measure.

To address this failure mode, we introduced the jp algorithm (jointly preprocessed evaluation algorithm), which replaces boundary dependent comparison with alignment based comparison. The central move is to evaluate after constructing an explicit alignment between system output and reference output, so that evaluation remains well defined even when tokenization or sentence boundary detection decisions differ. Under this view, preprocessing is not treated as a hidden precondition of evaluation but as a first class source of variation that must be controlled for during scoring.

We also showed how the same alignment perspective carries to downstream evaluation settings in which segmentation is part of the interface between components, including constituent parsing and grammatical error correction. Rather than forcing all outputs into a single canonical segmentation, jp based evaluation compares outputs in a way that isolates task relevant differences from artifacts introduced by inconsistent preprocessing.

The contributions of this chapter are threefold: (1) a concrete alignment based protocol for evaluation under inconsistent preprocessing, (2) a unified recipe that can be instantiated across tasks with different output structures, and (3) an evaluation stance oriented toward raw text conditions in which segmentation is variable. More broadly, evaluation by alignment reframes robustness not as a property of the model alone, but as a property of the entire measurement pipeline.

\singlespacing
\chapter{Conclusion}
\doublespacing

In this memoir, I addressed two key aspects of natural language processing that have structured my work over the past decade: the creation of linguistic resources and the design of evaluation methodologies for system performance. The common thread is methodological. In both strands, annotation and scoring are treated not as auxiliary engineering steps but as the formal interface between linguistic structure and computation, where seemingly minor representational choices can determine what a model can learn and what an evaluation can legitimately claim.

This perspective also clarifies the most immediate extensions of the present work. A first direction is to continue the morpheme based line of annotation for Korean by expanding the UniDive framework on top of Korean Universal Dependencies. UniDive foregrounds the empirical tension between universality, language diversity, and language specific idiosyncrasy in language technology, and it provides a principled setting in which Korean morphology can be represented without being compressed into word level conventions that erase linguistically relevant boundaries. I have already implemented UniDive for Korean based on Korean Universal Dependencies, and the current work focuses on stabilizing the morpheme level inventory and the conversion and validation pipeline.\footnote{\url{https://github.com/jungyeul/k-unidive}}

In parallel to resource creation, this memoir emphasized a second, equally central thread of my work: evaluation. Evaluation is often treated as a final reporting step, yet in practice it functions as an implicit definition of the task itself. When the definition is misspecified, progress becomes difficult to interpret and comparisons across systems become brittle. This is especially visible for end to end systems operating on raw text, where evaluation must remain meaningful even when system output diverges from the reference in surface form while preserving the intended content.

A first direction concerns fluency based evaluation, where metrics such as BLEU \citep{papineni-etal-2002-bleu}, ROUGE \citep{lin-2004-rouge}, and GLEU \citep{napoles-etal-2015-ground} are used as proxies for surface well formedness and n-gram level similarity. These metrics remain attractive because they are simple to compute, stable, and broadly comparable across settings. At the same time, their reliability depends crucially on the relationship between references and acceptable outputs. A single reference can under represent legitimate variation, while differences in preprocessing conventions can introduce artificial penalties that do not reflect system quality. My future work will therefore treat fluency based evaluation not as a fixed recipe, but as an object of study: when these metrics track human judgments, when they fail, and how their assumptions interact with text normalization, segmentation conventions, and the degree of permissible paraphrase.

A second direction is multi reference evaluation. When a task admits multiple acceptable realizations, evaluation should reflect that multiplicity rather than requiring systems to match a single surface form. Multi reference settings reduce false negatives, improve robustness under naturally variable outputs, and better align automatic scores with the space of valid solutions. My ongoing and planned work will further develop evaluation protocols that explicitly accommodate multiple targets, including for rewriting style tasks and grammatical error correction. This includes refining how reference sets are constructed, how coverage and diversity of references are characterized, and how multi reference scoring should be reported to remain comparable across datasets and studies.

These two directions converge on a shared point: evaluation should preserve interpretability under realistic variation. Fluency based metrics provide continuity with prior work and a pragmatic baseline, while multi reference evaluation addresses the core limitation of single reference scoring in open ended generation and correction. By integrating these components into a coherent evaluation framework, my goal is to make reported improvements more trustworthy and to enable system comparisons that remain valid when outputs differ in form but not in substance.

\singlespacing
\bibliographystyle{plainnat}
\bibliography{park}

\end{document}